\documentclass[twoside,11pt]{article}

\usepackage{hyperref}
\usepackage{apacite}

\usepackage{jair}

\jairheading{72}{2021}{329-376}{06/2021}{10/2021}
\ShortHeadings{Enhancing SUNNY for Algorithm Selection}
{Liu, Amadini, Gabbrielli \&  Mauro}
\firstpageno{1}

\usepackage{subcaption}
\usepackage{
beramono 
} 

\usepackage{algorithm}

\usepackage{lineno}
\usepackage[T1]{fontenc}
\usepackage[utf8]{inputenc}
\usepackage{algpseudocode}
\usepackage{listings}
\usepackage{color}
\usepackage{xspace}
\usepackage{url}
\usepackage{multicol}
\usepackage{booktabs}
\usepackage{graphicx}
\usepackage{enumerate}
\usepackage{xcolor}
\usepackage{makecell}
\usepackage{soul}
\usepackage{todonotes}
\usepackage{amsthm}

\usepackage{pgfplotstable}
\usepackage{booktabs}
\usepackage{colortbl}
\usepackage{pgfplots}
\usepackage{pgfplotstable}
\usepackage{tikz}
\usepackage{collcell}
\usepackage{graphicx}
\usepackage{booktabs}
\usepackage{xcolor}

\newcommand\tableColorR{61}
\newcommand\tableColorG{120}
\newcommand\tableColorB{179}

\pgfplotstableset{
    /color cells/min/.initial=0,
    /color cells/max/.initial=1000,
    /color cells/textcolor/.initial=,
    color cells/.code={%
        \pgfqkeys{/color cells}{#1}%
        \pgfkeysalso{%
            postproc cell content/.code={%
                \begingroup
                \pgfkeysgetvalue{/pgfplots/table/@preprocessed cell content}\value
                \ifx\value\empty
                    \endgroup
                \else
                \pgfmathfloatparsenumber{\value}%
                \pgfmathfloattofixed{\pgfmathresult}%
                \let\value=\pgfmathresult
                \pgfplotscolormapaccess
                    [\pgfkeysvalueof{/color cells/min}:\pgfkeysvalueof{/color cells/max}]
                    {\value}
                    {\pgfkeysvalueof{/pgfplots/colormap name}}%
                \pgfkeysgetvalue{/pgfplots/table/@cell content}\typesetvalue
                \pgfkeysgetvalue{/color cells/textcolor}\textcolorvalue
                \toks0=\expandafter{\typesetvalue}%
                \xdef\temp{%
                    \noexpand\pgfkeysalso{%
                        @cell content={%
                            \noexpand\cellcolor[rgb]{\pgfmathresult}%
                            \noexpand\definecolor{mapped color}{rgb}{\pgfmathresult}%
                            \ifx\textcolorvalue\empty
                            \else
                                \noexpand\color{\textcolorvalue}%
                            \fi
                            \the\toks0 %
                        }%
                    }%
                }%
                \endgroup
                \temp
                \fi
            }%
        }%
    }
}

\newtheorem{definition}{Definition}

\usepackage{amsmath,amssymb} 

\usepackage{multirow}

\usepackage{xspace}
\usepackage{booktabs}
\usepackage{tabularx}

\usepackage{csvsimple}
\makeatletter
\csvset{
  autotabularcenter/.style={
    file=#1,
    after head=\csv@pretable\begin{tabular}{|*{\csv@columncount}{c|}}\csv@tablehead,
    table head=\hline\csvlinetotablerow\\\hline,
    late after line=\\,
    table foot=\\\hline,
    late after last line=\csv@tablefoot\end{tabular}\csv@posttable,
    command=\csvlinetotablerow},
  autobooktabularcenter/.style={
    file=#1,
    after head=\csv@pretable\begin{tabular}{*{\csv@columncount}{c}}\csv@tablehead,
    table head=\toprule\csvlinetotablerow\\\midrule,
    late after line=\\,
    table foot=\\\bottomrule,
    late after last line=\csv@tablefoot\end{tabular}\csv@posttable,
    command=\csvlinetotablerow},
}
\makeatother

\newcommand{\aslib}{\textnormal{ASlib}\xspace}
\newcommand{\sunny}{SUNNY\xspace}
\newcommand{\tsunny}{\textsf{TSUNNY}\xspace}
\newcommand{\sunnycp}{\textsf{sunny-cp}\xspace}
\newcommand{\sunnyas}{\textsf{sunny-as}\xspace}
\newcommand{\tool}{\textsf{sunny-as2}\xspace}

\newcommand{\toolk}{\textsf{sunny-as2-k}\xspace}
\newcommand{\toolf}{\textsf{sunny-as2-f}\xspace}
\newcommand{\toolfk}{\textsf{sunny-as2-fk}\xspace}
\newcommand{\gsunny}{\textsf{greedy-SUNNY}\xspace}
\newcommand{\all}{H_{all}\xspace}
\newcommand{\sch}{H_{sch}\xspace}
\newcommand{\sel}{H_{sel}\xspace}
\newcommand{\borda}{\textsf{Borda}\xspace}

\newcommand{\parx}[1]{\mathsf{PAR}_{#1}}

\newcommand{\parten}{\mathsf{\parx{10}}}

\newcommand{\I}{\mathcal{I}}
\newcommand{\A}{\mathcal{A}}
\newcommand{\F}{\mathcal{F}}
\newcommand{\cmp}{\textsf{cmp}\xspace}
\newtheorem{ex}{Example}

\usepackage[figuresright]{rotating}

\begin{document}

\title{\tool: Enhancing SUNNY for Algorithm Selection}

\author{\name Tong Liu \email lteu@icloud.com  \\
	   \addr Faculty of Computer Science, \\
       Free University of Bozen-Bolzano,
       Italy 
       \AND
	   \name Roberto Amadini \email roberto.amadini@unibo.it\\
       \name Maurizio Gabbrielli \email maurizio.gabbrielli@unibo.it \\
       \addr Department of Computer Science and Engineering, \\
       University of Bologna, 
       Italy
       \AND
       \name Jacopo Mauro \email mauro@imada.sdu.dk \\
       \addr Department of Mathematics and Computer Science,\\
        University of Southern Denmark, Denmark}


\maketitle



\begin{abstract}
SUNNY is an Algorithm Selection (AS) technique originally tailored for 
Constraint Programming (CP). SUNNY is based on the $k$-nearest neighbors algorithm and 
enables one to schedule, from a portfolio of solvers, a subset of 
solvers to be run on a given CP problem. 
This approach has proved to be effective for CP problems.

In 2015, the ASlib benchmarks were released 
for comparing AS systems coming 
from disparate fields (e.g., ASP, QBF, and SAT) and SUNNY was extended to deal 
with 
generic AS problems. This led to the development of \sunnyas, a prototypical  
algorithm selector based on SUNNY for ASlib scenarios. A major improvement 
of \sunnyas, called \tool, was then submitted to the Open Algorithm Selection Challenge (OASC) 
in 2017, where it turned out to be the best approach 
for the runtime minimization of decision problems.

In this work we present the technical advancements of \tool, by 
detailing through several empirical evaluations and by providing new insights. 
Its current version, built on the top of the preliminary version submitted to OASC, 
is able to outperform \sunnyas and other
state-of-the-art AS methods, including
those who did not attend the challenge.
\end{abstract}



\section{Introduction}
\label{sec:intro}
Solving combinatorial problems is hard and, especially for NP-hard problems,
there is not a 
dominant algorithm for each class of problems. 
A natural way to face the disparate nature of combinatorial problems and 
obtain a globally better solver is to use a \textit{portfolio} of 
different algorithms (or solvers) to be selected on different problem instances.
The task of identifying suitable algorithm(s) for specific instances of a problem 
is known as per-instance \textit{Algorithm Selection} (AS). 
By using AS, portfolio solvers are able to outperform state-of-the-art 
single solvers in many fields such as Propositional Satisfiability 
(SAT), Constraint Programming (CP), Answer Set Programming (ASP), Quantified 
Boolean Formula (QBF). 

A significant number of domain-specific AS strategies have been studied. However, it is hard if not impossible to judge 
which of them is the best strategy in general. To address this problem, the 
\textit{Algorithm Selection library} (ASlib)~\cite{bischl2016aslib} has been proposed. 
ASlib consists of scenarios collected from a broad range of domains, 
aiming to give a  cross-the-board performance comparison of different AS 
techniques, with the scope of comparing various AS techniques on the same 
ground.
Based on the ASlib benchmarks, rigorous validations and AS competitions 
have been recently held.


In this paper, we focus on the SUNNY portfolio approach 
\cite{sunnycp2,sunny}, 
originally developed to solve Constraint 
Satisfaction Problems (CSPs). SUNNY is based on the 
$k$-nearest neighbors ($k$-NN) algorithm.
%
%
Given a previously unseen problem instance $P$, it first extracts its 
\textit{feature vector} $F_P$, i.e., a collection of numerical attributes 
characterizing $P$, and then finds the $k$ training instances 
``most similar'' to $F_P$ according to the Euclidean distance. 
Afterwards, SUNNY selects the best solvers for 
these $k$ instances, and assign a time slot proportional to the number of 
solved instances to the selected solvers.
Finally, the selected solvers are
sorted by average solving time and then executed on $P$.

Initially designed for CSPs, SUNNY has then been customized to 
solve Constraint Optimization Problems (COPs) and 
to enable the parallel execution of its solvers.
The resulting portfolio solver, called \sunnycp~\cite{sunnycp2},
won the gold medal in the Open Track of the Minizinc Challenge~\cite{mzn-challenge}---the yearly international 
competition for CP solvers---in 2015, 2016, and 2017~\cite{DBLP:journals/tplp/AmadiniGM18}.

In 2015, SUNNY was extended to deal with general AS scenarios---for which CP problems are a particular case~\cite{paper_cilc}.
The resulting tool, called \sunnyas, natively handled ASlib scenarios 
and was submitted to the 2015 
ICON Challenge on Algorithm Selection~\cite{kotthoff2015icon} to be compared 
with other AS systems.
Unfortunately, the outcome was not satisfactory: only a few
competitive results were achieved by \sunnyas, that turned out to be 
particularly weak on SAT scenarios. We therefore decided to improve the performance 
of \sunnyas 
by following two main paths: \textit{(i)} \textit{feature selection}, and 
\textit{(ii)} \textit{neighborhood size configuration}.

Feature selection (FS) is a well-known process that consists of 
removing redundant and potentially harmful features 
from feature vectors.  
It is well established that a good feature selection can lead to significant performance gains.
In the 2015 ICON challenge, one version of \sunnyas used a simple \emph{filter} method 
based on information gain that, however, did not bring any benefit.
This is not surprising, because filter methods are efficient but agnostic of the specific 
predictive task to be performed---they work as a pre-processing step 
regardless of the chosen predictor. 
Hence we decided to move to \emph{wrapper} methods, which are more computationally 
expensive---they use the prediction system of interest to assess the selected features---but typically more accurate.



The neighborhood size configuration (shortly, \textit{$k$-configuration}) consists in 
choosing an optimal $k$-value for the $k$-nearest neighbors algorithm on which SUNNY 
relies. \sunnyas did not use any $k$-configuration in the 2015 ICON challenge, and this 
definitely penalized its performance. For example, 
\citeA{DBLP:conf/lion/LindauerBH16} 
pointed up that
SUNNY can be significantly 
boosted by training and tuning its $k$-value.

The new insights on feature selection and $k$-configuration 
led to development of \tool, an extension of \sunnyas that enables 
the SUNNY algorithm to learn \emph{both} the supposed best features and the 
neighborhood size.
We are 
not aware of other AS approaches selecting features and neighborhood size in the way 
\tool does it.
Moreover, \tool exploits a polynomial-time \emph{greedy} version of \sunny making the training phase more efficient---the worst-case time complexity of the original \sunny is indeed
exponential in the size of the portfolio.

In 2017, a preliminary version of \tool was submitted to the \textit{Open 
Algorithm Selection Challenge} (OASC), a revised edition of the 2015 ICON challenge. 
Thanks to the new enhancements, \tool achieved much better results~\cite{lindauer2019algorithm}:
it reached the overall third position and, in particular, it was the approach 
achieving the best runtime minimization for satisfaction problems 
(i.e., the goal for which SUNNY was originally designed).
Later on, as we shall see, the OASC version of \tool was further improved.
In particular, we tuned the configuration of its parameters 
(e.g., cross-validation mode, size of the training set, etc.) after conducting a 
comprehensive set of experiments over the OASC scenarios.

In this work, we detail the technical improvements of \tool by 
showing their impact on different scenarios of the ASlib.
The original contributions of this paper include:
\begin{itemize}
  \item the description of \tool and its variants, i.e., \toolf, \toolk and 
\toolfk---performing respectively feature selection, $k$-configuration and both;\footnote{Despite the good results in the OASC, a paper describing \tool was 
never  published before.}
  \item extensive and detailed empirical evaluations showing how the performance of \tool can vary across different scenarios, and motivating the default settings of \tool parameters;
  \item an original and in-depth study of the SUNNY algorithm, including insights on the 
  instances \emph{unsolved} by \tool and the use of a \emph{greedy} approach as a surrogate 
  of the original SUNNY approach;
  \item an empirical comparison of \tool against different state-of-the-art algorithm selectors, showing a promising and robust performance of \tool across different scenarios and performance metrics.
\end{itemize}

We performed a considerable number of experiments to understand the 
impact of the new technical improvements. 
Among the lessons we learned, we mention that:
\begin{itemize}
 \item feature selection and $k$-configuration are quite effective for SUNNY, and perform 
 better when \emph{integrated};
 \item the greedy approach enables a training 
methodology which is faster and \textit{more effective} 
w.r.t.~a training performed with the original SUNNY approach;
 \item the ``similarity assumption'' on which the $k$-NN algorithm used by SUNNY relies, 
 stating that similar instances have similar performance, is weak if not wrong;
 \item the effectiveness of an algorithm selector is strongly coupled to the evaluation metric used 
 to measure its performance. Nonetheless, \tool appears to more robust than other approaches when changing the performance metric.
\end{itemize}

The performance of \tool naturally varies according to the peculiarities of 
the given scenario and the chosen performance metric. We noticed that \tool performs 
consistently well on scenarios having a reasonable amount of instances 
and where the theoretical \emph{speedup} of 
a portfolio approach, w.r.t~the best solver of the scenario, is not minimal.

We also noticed that a limited amount of training instances is enough to reach a good 
prediction performance and that the nested cross-validation leads to more robust results.
In addition,
the results of our experiments 
corroborate some previous findings, e.g., 
that it is possible to reach the best performance by considering only a small 
neighborhood size and a small number of features.

\emph{Paper structure. }
In Sect.~\ref{sec:related} 
we review the literature on Algorithm Selection.
In Sect.~\ref{sec:preliminaries} we give background notions before 
describing \tool in Sect.~\ref{sec:sunny-opt}.
Sect.~\ref{sec:experiments} describes the experiments 
over different configurations of \tool, 
while Sect.~\ref{sec:sunny-more} provides 
more insights on the SUNNY algorithm, including a comparison with other AS approaches. 
We draw concluding remarks in 
Sect.~\ref{sec:conclusions}, 
while the Appendix contains additional experiments and information for 
the interested reader.

\section{Related Work}
\label{sec:related}

Algorithm Selection (AS) aims at identifying on a per-instance basis the relevant 
algorithm, or set of algorithms, to run in order to enhance the problem-solving 
performance.
This concept finds wide application in decision problems as well
as in optimization problems, although most of the AS systems have been developed for decision problems --- in particular for SAT/CSP problems. However, given the generality and flexibility of the AS framework, AS approaches have also been used in other domains such as combinatorial optimization, planning, scheduling, and so on. 
In the following, we provide an overview of the most known and successful 
AS approaches we are aware of. For further insights about AS 
and related problems, we refer the interested reader to the comprehensive 
surveys in 
\citeA{kerschke2019automated,kotthoff2016algorithm,DBLP:conf/lopstr/AmadiniGM15,
DBLP:journals/csur/Smith-Miles08}.

About a decade ago, AS began to attract the attention of the SAT community and portfolio-based techniques started their spread. In particular, 
suitable tracks were added to the SAT competition to evaluate the 
performance of portfolio solvers. 
%
%
SATzilla~\cite{xu2008satzilla,xu2012evaluating} was one of the first SAT portfolio solvers. Its first version~\cite{xu2008satzilla} used a ridge regression method to predict the effectiveness (i.e., the runtime or a performance score) of a SAT solver on unforeseen SAT instances.
This version won several gold medals in the 2007 and 2009 SAT competitions.
 
In 2012, a new version of SATzilla was introduced~\cite{xu2012evaluating}. 
This implementation improved the previous version with 
a weighted random forest approach provided 
with a cost-sensitive loss function for punishing misclassification in direct 
proportion to their performance impact.
These improvements allowed SATzilla to outperform the previous version and to win the SAT Challenge in 2012.

Another well-known AS approach for SAT problems is 3S~\cite{kadioglu2011algorithm}.
Like SUNNY, the 3S selector relies on $k$-NN under the assumption that 
performances of different solvers are similar for instances with similar 
features.  
3S combines AS and algorithm scheduling, in static and dynamic ways. In particular, it first executes in the first 10\% of its time budget short runs of solvers according to a fixed schedule computed offline. Then, at run time, a designated solver is selected via $k$-NN and executed for the remaining time.
3S was the best-performing dynamic portfolio at the International SAT Competition 2011.

3S was inspired by ISAC~\cite{isac}, a method for instance-specific algorithm configuration based on $g$-means clustering~\cite{k_means} and $k$-NN. 
The goal of ISAC is to produce a suitable parameter setting for a new input instance, given a set of training samples.
ISAC can also be used as 
an algorithm selector: in the work by \citeA{isac-as}, three different ways of 
using  ISAC to generate SAT portfolio solvers are presented (pure solver 
portfolio, optimized solver portfolio, and instance-specific meta-solver 
configuration).

Another approach based on $k$-NN is SNNAP~\cite{collautti2013snnap}, which first predicts the performance of each algorithm with regression models and
then uses this information for a $k$-NN approach in the predicted performance space. As for 3S, SNNAP was inspired by the ISAC approach. In particular, it augmented ISAC by taking into account the past performance of solvers as part of the feature vector.

CSHC~\cite{malitsky2013algorithm} is a clustering-based approach also  inspired by ISAC and 3S.  
In particular, CHSC combines 3S’s static scheduler with an algorithm selector based on cost sensitive hierarchical clustering which creates a multi-class classification model. 
CSHC won the gold medal in the 2013 SAT competition.

From the algorithm selection point of view, the only difference between selecting CSP solvers or SAT solvers is the nature of the underlying problems, which is in turn reflected into different (yet similar) types of features.
Besides this, the goal is the same: minimizing the (penalized) average runtime. Hence, SAT portfolio approaches can be quite straightforwardly adapted to CSP portfolio approaches (and vice versa). An empirical 
evaluation of different AS approaches for solving CSPs (including adapted SAT portfolios and off-the-shelf machine learning approaches) is provided 
in \citeA{cpaior_paper}.

As mentioned in Sect.~\ref{sec:intro}, SUNNY was originally designed to solve CSPs. 
Empirical comparisons between SUNNY and ISAC-like, SATzilla-like and 3S-like 
approaches for solving CSPs are reported in 
\citeA{sunny,paper_amai,amadini2016extensive}.\footnote{
	The notation ISAC-like, SATzilla-like and 3S-like
	indicates that the original SAT-based approach was adapted or re-implemented to be evaluated on CP instances.}
Apart from SUNNY, other well-known CSP portfolio approaches are CPHydra and Proteus.

CPHydra~\cite{cphydra} was probably the first CSP solver using a portfolio approach. Similarly to SUNNY, CPHydra employs $k$-NN to compute a  schedule of solvers which maximizes the
chances of solving an instance within a given timeout. 
CPHydra, however, computes the schedule of solvers differently, 
and does not define any heuristic for scheduling the selected solvers.
CPHydra won the 2008 International CSP Solver Competition, but subsequent investigations~\cite{cpaior_paper,sunny,amadini2016extensive} showed
some weaknesses in scalability and runtime minimization.

Proteus~\cite{proteus} is a hierarchical portfolio-based approach to CSP solving that does not rely purely
on CSP solvers, but may convert a CSP to SAT by choosing a conversion
technique and an accommodating SAT solver. A number of machine learning techniques are employed at each level in the hierarchy, e.g., decision trees, regression, $k$-NN, and support vector machines.

Besides the SAT and CSP settings, the flexibility of the AS framework 
led to the construction of effective algorithm portfolios in related settings.
For example, portfolio solvers as Aspeed and 
claspfolio have been proposed for solving Answer-Set Programming (ASP) problems.
Aspeed~\cite{hoos2015aspeed} is a variant of 3S where the per-instance 
long-running solver selection 
has been replaced by a solver schedule.
\citeA{DBLP:conf/lion/LindauerBH16} 
released
ISA which further improved Aspeed by introducing an optimization
objective ``timeout-minimal'' in the schedule generation.
Claspfolio~\cite{claspfolio2} supports different AS mechanisms (e.g. ISAC-like, 3S-like, SATzilla-like) and was a gold medallist in different tracks of the ASP Competition 2009 and 2011.
The contribution of ME-ASP~\cite{maratea2013multi}
is also worth mentioning. ME-ASP identifies one solver per ASP instance. To make its prediction robust, 
it exploits the strength of several independent classifiers (six in total, including $k$-NN, SVM, Random forests, etc) and chooses the best one according to their cross-validation performances on training instances.
An improvement of ME-ASP is described in \citeA{maratea2013policy}, where the 
authors 
added the capability of updating the learned policies when the original approach fails to give good predictions. The idea of coupling classification with policy adaptation methods comes from AQME~\cite{pulina2009self}, a multi-engine solver for quantified Boolean formulas (QBF).

SATZilla~\cite{xu2012evaluating} has been rather influential also outside the SAT domain.
For example, in AI planning,
Planzilla~\cite{rizzini2017static} and its improved variants 
(model-based approaches)
were all inspired by
the random forests and regression techniques proposed by SATZilla/Zilla.
Similarly, for Satisfiability Modulo Theories (SMT) problems, MachSMT~\cite{scott2020machsmt} 
was recently introduced and its essential parts 
also rely on random forests.
The main difference between the model-based Planzilla selector
and MachSMT is that the first one chooses solvers 
minimizing the ratio between solved instances and solving time, 
while the latter only considers the solving time of candidate solvers.

%

A number of AS approaches have been developed to tackle optimization 
problems. In this case, mapping SAT/CSP algorithm selection techniques to the more 
general Max-SAT~\cite{DBLP:journals/ai/AnsoteguiGMS16} and Constraint Optimization Problem (COP)~\cite{paper_amai} settings are not so straightforward. The main issue here is how to evaluate sub-optimal solutions, and optimal solutions for which optimality has not been proved by a solver. 
A reasonable performance metric for optimization problems computes 
a (normalized) score reflecting the quality of the best solution found by  a solver in a given time window. However, one 
can also think to other metrics taking into account the \emph{anytime} 
performance of a solver, i.e., the sub-optimal solutions it finds during the 
search (see, e.g., the area score of \citeA{paper_amai}).

We reiterate here the importance of tracking the sub-optimal solutions for AS scenarios, especially for those AS approaches that, like SUNNY, schedule more than one solver. The importance of a good anytime performance has been also acknowledged by the MiniZinc Challenge~\cite{DBLP:journals/constraints/StuckeyBF10}, the yearly international competition for CP solvers, that starting from 2017 introduced the area score which measures the area under the curve 
defined by $f_s(t) = v$ where $v$ is the best value found by solver $s$ at time $t$.
To our knowledge, SUNNY is the only general-purpose AS approach taking into account the area score to select a solver: the other approaches only consider the best value $f(\tau)$ at the stroke of the timeout $\tau$.

A number of \emph{ad hoc} AS approaches have been developed instead for some specific optimisation problems like Knapsack, Most Probable Explanation, Set Partitioning, Travel Salesman Problem~\cite{DBLP:journals/anor/GuoH07,DBLP:journals/corr/abs-1211-0906,DBLP:journals/ec/KerschkeKBHT18,DBLP:conf/lion/KotthoffKHT15}.

Considering the AS approaches that attended the 2015 
ICON challenge~\cite{lindauer2019algorithm},
 apart from \sunnyas, other five AS systems were submitted: ASAP, 
AutoFolio, FlexFolio, Zilla, ZillaFolio. It is worth noticing that, unlike SUNNY, all of them 
are \emph{hybrid} systems combining different AS approaches.

ASAP (Algorithm Selector And Prescheduler system)~\cite{gonard17a,gonard2019algorithm} relies on random forests and $k$-NN. It combines pre-solving scheduling and per-instance algorithm selection by training them jointly.
 
AutoFolio~\cite{lindauer2015autofolio} combines several algorithm selection approaches (e.g., SATZilla, 3S, SNNAP, ISAC, LLAMA~\cite{llama}) in a single
system and uses algorithm configuration~\cite{smac} to search
for the best approach and its hyperparameter settings for the scenario
at hand.
Along with the scenarios in \aslib, 
AutoFolio also demonstrated its effectiveness in
dealing with Circuit QBFs \cite{hoos2018portfolio}.
Unsurprisingly, this paper also shows that the quality of the selected features
can substantially impact the selection accuracy of AutoFolio.
 
FlexFolio~\cite{flexfolio} is a claspfolio-based AS system~\cite{claspfolio2} 
integrating various 
feature generators, solver selection approaches, solver portfolios, as well as 
solver-schedule-based pre-solving techniques into a single, unified framework.
 
Zilla is an evolution of SATZilla~\cite{xu2008satzilla,xu2012evaluating} 
using pair-wise, cost-sensitive random forests combined with pre-solving
schedules. ZillaFolio combines Zilla and AutoFolio by first evaluating both approaches on the training set. Then, it chooses the best one for generating the predictions for the test set.

The OASC 
2017 challenge~\cite{lindauer17a} included a preliminary version of \tool,  
improved versions of ASAP (i.e., ASAP.v2 and ASAP.v3), an improved version of 
Zilla (i.e., {}*Zilla) and a new contestant which came in two 
flavors: AS-ASL and AS-RF. 
Both AS-ASL and AS-RF~\cite{malone2017asl} used a greedy wrapper-based 
feature selection approach with the AS selector as evaluator to locate relevant features. The system was trained differently for the 
two versions: 
AS-ASL uses ensemble learning model while AS-RF uses the random 
forest. A final schedule is built on the trained model. 


One common thing between ASAP.v2/3, {}*Zilla and AS-RF/ASL is 
that all of them attempt to solve an unseen problem instance by statically 
scheduling a number of solver(s) 
\emph{before} the AS process. The solver AS-ASL selects a single solver while ASAP and 
*Zilla define a static solver schedule.
A comprehensive summary of the above approaches as well as several challenge 
insights are discussed in \citeA{lindauer2019algorithm}. 

For the sake of completeness we also mention parallel AS approaches, although they do not fall within the scope of this paper.
The parallel version of \sunny~\cite{sunnycp2} won several gold medals in the MiniZinc Challenges 
by selecting relevant solvers to run in parallel
on a per-instance basis. In contrast, 
the work by \citeA{lindauer2017automatic} studied
the methods for static parallel portfolio construction.
In addition to selecting relevant solvers, they
also identifies performing parameter values for the selected 
solvers. Given a limited time budget for training,
a large amount of candidate solvers and their 
wide configuration space, the task of making
parallel portfolio is not trivial. 
Therefore, they examined greedy techniques to
speed up their procedures, and clause sharing for algorithm configuration
to improve prediction performance.
Likewise, in the domain of AI planning, 
portfolio parallelization has also been investigated.
An example is the 
static parallel portfolio proposed by \citeA{vallati2018configuration}
where planners are scheduled to each available CPU core.

We conclude by mentioning some interesting AS approaches that, however, did not 
attend the 2015 and 2017 challenges.
The work by~\citeA{ansotegui2018self} is
 built upon CSHC. They first estimate the confidence of 
the predicted solutions and then use the estimations to decide whether it is 
appropriate to substitute the solution with a static schedule.
By using the OASC dataset, the authors demonstrated a 
significant improvement over the original CSHC approach reaching 
the state-of-the-art performance in several scenarios.
%
In \citeA{misir2017alors}, the AS problem is seen as a recommendation problem 
solved with the well-known 
technique of collaborative filtering~\cite{DBLP:journals/fthci/EkstrandRK11}. This approach has a 
performance similar to the 
initial version of \sunnyas. 
In \citeA{loreggia2016deep}, the authors 
introduce an original approach that transforms  
the text-encoded instances for the AS into a 2-D image. These images are later 
processed by a Deep Neural Network system to predict the best solver to use 
for each of them. This approach enables to find out (and also generate) relevant features for the 
Algorithm Selection. Preliminary experiments are quite encouraging, even though 
this approach still lags behind w.r.t. state-of-the-art approaches who are 
using and exploiting crafted instance features.

\section{Preliminaries}\label{sec:preliminaries}

In this section we formalize the Algorithm Selection problem 
\cite{bischl2016aslib} and the metrics 
used to evaluate algorithm selectors. We then briefly introduce 
the feature 
selection process and the 
SUNNY algorithm on which \sunnyas and \tool rely. We conclude by providing 
more details about the OASC and its scenarios.

\subsection{Algorithm Selection Problem and Evaluation Metrics}
\label{sec:eval}
To create an algorithm selector we need a scenario with more than 
one algorithm to choose, some instances on which to apply the selector, and a 
performance metric to optimize. This information can be formally defined as 
follows.

\begin{definition}[AS scenario]
An AS scenario is a triple $(\I, \A, m)$ where:
\begin{itemize}
 \item 
$\I$ is a set of \textit{instances},
\item
$\A$ is a set (or portfolio) of \textit{algorithms} (or solvers) with $|\mathcal{A}| > 1$,
\item
$m : \I \times \A \to \mathbb{R}$ is a \textit{performance metric}.
\end{itemize}
\end{definition}

Without loss of generality, from now on we assume that lower values for the performance metric $m$ are better, i.e., the goal is to \emph{minimize} $m$.

An algorithm selector, or shortly a selector, is a function that for each instance of the scenario aims to return the best algorithm, according to the performance metric, for that instance. Formally:
\begin{definition}[Selector]
Given an AS scenario $(\I, \A, m)$ a selector $s$ is a total mapping from $\I$ 
to $\A$.
\end{definition}

The algorithm selection problem~\cite{rice1976algorithm} consists in creating 
the best possible selector. Formally:
\begin{definition}[AS Problem]
Given an AS scenario $(\I, \A, m)$ the AS Problem is the problem of 
finding the selector $s$ such that the overall performance 
$\sum\limits_{i \in \I} m(i, s(i))$ is minimized.
\end{definition}

If the performance metric $m$ is fully defined, the AS Problem can be easily solved by 
assigning to every instance the algorithm with lower value of $m$. 
Unfortunately, 
in the real world, the performance metric $m$ on
$\I$ is only \emph{partially} known. 
In this case, the goal is to define a selector able to \emph{estimate} 
the value of $m$ for the instances $i \in I$ where $m(i,A)$ is unknown.
A selector can be 
validated by partitioning $\I$ into a 
training set $\I_{tr}$  
and a test set 
$\I_{ts}$. The training instances of 
$\I_{tr}$ are used to build the selector $s$, while the test instances of 
$\I_{ts}$ are used to evaluate the performance of $s$: 
$\sum\limits_{i \in \I_{ts}} m(i, s(i))$.
As we shall see, the training set $\I_{tr}$ can be further split to tune and validate the 
parameters of the selector.

Different approaches have been proposed to build and evaluate an algorithm 
selector. First of all, since the instances of $\I$ are often too hard to solve in a reasonable time, typically a solving \emph{timeout} $\tau$ is set.
For this reason, often the performance metric is extended with other 
criteria to penalize an algorithm selector that does not 
find any solution within the timeout. One of the most used is the \emph{Penalized Average Runtime} (PAR) score with penalty $\lambda > 1$ that penalizes instances 
not solved within the timeout with $\lambda$ times the timeout.
Formally, if $m$ denotes the runtime, it is defined as
 $\parx{\lambda} = \dfrac{1}{|\I|}\sum\limits_{i \in \I} m'(i, s(i))$
where
$$m'(i, A) = \begin{cases}
                  m(i, A) & \text{if $m(i, A) < \tau$}\\
                  \lambda \times \tau & \text{otherwise.}
                  \end{cases}
$$
For example, in both the 2015 ICON challenge and the OASC, the $\parten$ score 
was used for
measuring the selectors' performance on every single scenario.

Unfortunately, the PAR value can greatly change across different scenarios 
according to the timeout, making it difficult to assess the global performance
across all the scenarios.
Hence, when dealing with heterogeneous scenarios, it is often 
better to consider \emph{normalized metrics}. 
As baselines, one can consider the performance of the \emph{single best solver} (SBS, the best individual solver according to the performance metric) of the scenario as upper bound  and the performance of the \emph{virtual best solver} (VBS, the oracle selector always able to  
 pick the best solver for all the instances in the test set) as lower bound.
Ideally, 
the performance of a selector should be in between the performance of the SBS and that of the VBS. 
However, while an algorithm selector can never outperform the VBS, it might happen that it performs worse than the SBS. This is more likely to happen when the gap between SBS and VBS is exiguous.

Two metrics are often used in the literature to compare algorithm selectors: 
the \emph{speedup} or \emph{improvement factor}~\cite{lindauer2019algorithm} 
and the \emph{closed  
gap}~\cite{lindauer17a}.
The speedup is a number that measures the relative 
performance of two systems. If $m_s$ and $m_{\text{VBS}}$ are respectively the cumulative performances of 
a selector $s$ and the virtual best solver across all the instances of a scenario,
the speedup of the VBS w.r.t. the 
selector is defined as the ratio 
between $m_s$ and $m_{\text{VBS}}$. Since the selector can not be faster than the VBS, 
this value is always greater than 1, and values closer to 1 are better. To 
normalize this metric in a bounded interval (the upper bound varies across different scenarios) the fraction can be reversed by considering the ratio between $m_{\text{VBS}}$ and $m_s$. In this case the value always falls in $(0, 1]$, and the greater the value the better the selector.

Unlike the speedup, the closed gap score measures
how good a selector is in \emph{improving} the performance of the SBS w.r.t.~the VBS in the AS scenario. Assuming that $m_{\text{SBS}}$ is the cumulative performance of the SBS across all 
the instances of the scenario, the closed gap is defined as:
\[
   \frac{m_{\text{SBS}} - m_s }{m_{\text{SBS}} - m_{\text{VBS}}}
\]
A good selector will have a performance $m_s$ close to the virtual best 
solver, 
which makes the closed gap score close to $1$. On the contrary, 
a poor performance 
consists of having $m_s$ close to the single best solver $m_{\text{SBS}}$,
thus making the closed gap close to $0$ if not even lower.

An alternative way to evaluate the performance of algorithm selectors is to use \textit{comparative} scores without considering the SBS and VBS baselines.
For example, in the MiniZinc Challenge~\cite{mzn-challenge} a \emph{Borda} count is used to 
measure the performance of CP solvers. The Borda count 
allows voters (instances) to order the candidates (solvers) according to their preferences and 
giving to them points corresponding to the number of candidates 
ranked lower. Once all votes have been counted, the candidate with 
the 
most points is the winner.
This scoring system can be 
applied to algorithm selectors in a straightforward way.

Formally, let $(\I, \A, m)$ be a scenario, $\mathcal{S}$ a set of 
selectors, $\tau$ the timeout. Let us denote with $m(i,s)$ the 
performance of selector $s$ on problem $i$. The Borda score of selector $s \in 
\mathcal{S}$ on instance $i \in \I$ is $\borda(i,s) = \sum_{s' \in \mathcal{S} 
- \{s\}} \cmp(m(i,s), m(i, s'))$ where the comparative function $\cmp$ is 
defined as:
\[
\cmp(t, t') = \begin{cases}
0   & \text{ if }t = \tau\\
1   & \text{ if }t < \tau \wedge t' = \tau\\
0.5 & \text{ if }t = t ' = 0\\
\dfrac{t'}{t+t'} & \text{ otherwise.}
\end{cases} 
\]

Since $\cmp$ is always in $[0,1]$, the score $\borda(i,s)$ is always in $[0, 
|\mathcal{S}|-1]$: the higher its value, the more selectors it can beat.
When considering multiple instances, the winner is the selector $s$ that 
maximizes the sum of the scores over all instances, i.e., $\sum_{i \in 
\mathcal{I}}{\borda(i,s)}$.

%

\subsection{Feature Selection}
Typically, AS scenarios 
characterize each instance $i \in \I$ with a corresponding feature vector $\F(i) \in \mathbb{R}^n$, and 
the selection of the best algorithm $A$ for $i$ is actually performed according to $\F(i)$, i.e., $A = s(\F(i))$. The \textit{feature selection} (FS) process 
allows one to consider smaller feature vectors $\F'(i) \in \mathbb{R}^m$, derived from $\F(i)$ by projecting a number $m \leq n$ of its features.
The purpose of feature selection is simplifying the prediction model, lowering the training and feature 
extraction costs, and hopefully improving the prediction accuracy.

FS techniques \cite{intro_feature_selection} basically consist of a combination of two components: 
a \emph{search technique} for finding good subsets of features, and an \emph{evaluation function} to score these subsets. 
Since exploring all the possible subsets of features is computationally intractable for non-trivial 
feature spaces, heuristics are employed to guide the search of the best subsets. 
Greedy search strategies usually come in two flavors: \emph{forward selection} and \emph{backward elimination}. In forward selection, features are progressively incorporated into larger and larger subsets. Conversely, in backward elimination features are progressively removed starting from all the available features. A combination of these two techniques, genetic algorithms, or local search algorithms such as simulated annealing are also used.

There are different ways of classifying FS approaches. 
A well established distinction is between \emph{filters} and \emph{wrappers}. 
Filter methods select the features regardless of the model, trying to suppress the least interesting ones. These methods are particularly efficient and robust to overfitting. In contrast, wrappers evaluate subsets of features possibly detecting the interactions 
between them. Wrapper methods can be more accurate than filters, but have two main disadvantages: they are more exposed to overfitting, and they have a much higher computational cost.
More recently, also \emph{hybrid} and \emph{embedded} FS methods have been proposed~\cite{reviewFS}.
Hybrid methods combine wrappers and filters to get the best 
of these two worlds.
Embedded methods  
are instead \textit{integrated} into the learning algorithm, i.e., 
they perform feature selection during the model training.


In this work we do not consider filter methods. We refer the interested readers 
to \citeA{ictai_paper} to know more about SUNNY with filter-based FS.

\subsection{\sunny and \sunnyas}\label{sec:sunny}
\label{sec:sunnyas}
The SUNNY portfolio approach was firstly introduced in \citeA{sunny}.
SUNNY relies on a number of assumptions:
	\emph{(i)} a small portfolio is usually enough to achieve a good performance;
	\emph{(ii)} solvers either 
	solve a problem quite quickly, or cannot solve it in reasonable time;
	\emph{(iii)} solvers perform similarly on similar instances;
	\emph{(iv)} a too heavy training phase is often an unnecessary burden.
In this section we briefly recap how SUNNY works, 
while in Sect.~\ref{sec:sunny-more} we shall address in more detail these assumptions---especially in light of the experiments reported in Sect.~\ref{sec:experiments}.

SUNNY is
based on the $k$-nearest neighbors ($k$-NN) algorithm and embeds built-in heuristics for schedule generation.
Despite the original version of SUNNY handled CSPs only, here we describe its generalised version---the one 
we used to tackle general ASlib scenarios.

Let us fix the set of instances $\I = \I_{tr} \cup \I_{ts}$, the set of algorithms $\mathcal{A}$, 
the performance metric $m$, and the runtime timeout $\tau$.
Given a test instance $x \in \I_{ts}$, \sunny 
produces a sequential schedule 
$\sigma = [(A_1, t_1), \dots, (A_h, t_h)]$ 
where algorithm $A_i \in \mathcal{A}$  runs for $t_i$ seconds on $x$ and $\sum_{i = 1}^{h} t_i = \tau$.
The schedule is obtained as follows.
First, \sunny employs $k$-NN to select from $\I_{tr}$ the subset 
$I_k$ of the $k$ instances closest to $x$  
according to the Euclidean distance computed on the feature vector $\F(x)$. 
Then, it uses three heuristics to compute $\sigma$:
\textit{(i)} $\sel$, for \emph{selecting} 
the most effective  algorithms $\{A_1, \dots, A_h\} \subseteq \mathcal{A}$ in $I_k$;
\textit{(ii)} $\all$, for \emph{allocating} 
to each $A_i \in \mathcal{A}$ a certain runtime 
$t_i \in [0, \tau]$ for $i = 1, \dots, h$;
\textit{(iii)} $\sch$, for \emph{scheduling} the 
sequential execution of the algorithms according to their performance in $I_k$.

The heuristics $\sel$, $\all$, and $\sch$ are based on the performance metric $m$, and depend on the application domain.
For CSPs,
$\sel$ selects the smallest sets of solvers $S \subseteq \mathcal{A}$ 
that solve the most instances in $I_k$, by using the 
runtime for breaking ties; $\all$ allocates to each $A_i \in S$ a time 
$t_i$ proportional to the instances that $S$ can solve in $I_k$, by using a 
special \emph{backup solver} for covering the instances of 
$I_k$ that are not solvable by any solver;
Finally, $\sch$ sorts the solvers by increasing solving time in $I_k$. For 
Constraint Optimization Problems the approach is similar, but different evaluation metrics are used to also consider the objective value and 
sub-optimal solutions~\cite{paper_amai}.
For more details about \sunny we refer the interested reader to 
\citeA{sunny,paper_amai}. Below 
we show Example \ref{ex:sunny} illustrating how \sunny works on a given CSP. 

\begin{table}[t]
\caption{Runtime (in seconds). $\tau$ means the solver timeout.}
\label{table:runtimes}
\centering
\scalebox{0.9}{
\begin{tabularx}{\linewidth}{XXXXXXX}
 \cline{2-6}\cline{2-6}
   & $x_1$ & $x_2$ & $x_3$ & $x_4$ & $x_5$ \\
\cline{1-6}
$A_1$ & $\tau$ & $\tau$ & \textbf{3} & $\tau$ & \textbf{278} \\
\cline{1-6}
$A_2$ & $\tau$ &  \textbf{593} & $\tau$ & $\tau$ & $\tau$ \\
\cline{1-6}
$A_3$ & $\tau$ & $\tau$ &  \textbf{36} & \textbf{1452} & $\tau$ \\
\cline{1-6}
$A_4$ & $\tau$ & $\tau$ & $\tau$ & \textbf{122} & \textbf{60} \\
\cline{1-6}\cline{1-6}

\multicolumn{1}{c}{}
\end{tabularx}
}
\vspace{-8pt}
\end{table}

\begin{ex}
	\label{ex:sunny}
	Let $x$ be a CSP, $\mathcal{A} = \{A_1, A_2, A_3, A_4 \}$ a portfolio, $A_3$ the 
	backup solver, $\tau = 1800$ seconds the solving timeout, 
	$I_k = \{x_1, ..., x_5\}$ the $k = 5$ neighbors of $x$, and the runtime 
	of solver $A_i$ on problem $x_j$ defined as in Tab.~\ref{table:runtimes}.
	In this case, the smallest set of solvers that solve most instances in the neighborhood are 
	$\{A_1, A_2, A_3\}$, $\{A_1, A_2, A_4\}$, and $\{A_2, A_3, A_4\}$. 
	The heuristic $\sel$ selects $S = \{A_1, A_2, A_4\}$ 
	because these solvers are faster in solving the instances in $I_k$.
	Since $A_1$ and $A_4$ solve 2 instances, $A_2$ solves 1 instance and $x_1$ 
	is not solved by any solver, the time window $[0, \tau]$ is partitioned in 
	$2 + 2 + 1 + 1 = 6$ slots: 2 assigned to $A_1$ and $A_4$, 1 slot to 
	$A_2$, and 1 to the backup solver $A_3$. Finally, $\sch$ sorts in ascending order the solvers 
	by average solving time in $I_k$. The final schedule produced by SUNNY is, therefore,
	$\sigma = [(A_4, 600), (A_1, 600), (A_3, 300), (A_2, 300)]$.
\end{ex}
One of the goals of SUNNY is to avoid the \emph{overfitting} w.r.t. the performance of 
the solvers in the selected neighbors. For this reason, their runtime is only 
marginally used to allocate time to the solvers. A similar but more 
runtime-dependent approach like, e.g., CPHydra~\cite{cphydra} would instead 
compute a runtime-optimal allocation $(A_1, 3), (A_2, 593), (A_4, 122)$, able to cover all 
the neighborhood instances, and then it would distribute this allocation in the solving time window 
$[0, \tau]$.
SUNNY does not follow this logic
to not be too 
tied to the strong assumption that 
the runtime in the neighborhood faithfully reflect the runtime on the instance to be solved.
To understand the rationale behind this choice, let us see the CPHydra-like schedule above: $A_1$ is the solver with the best average runtime in the neighborhood, but its time slot is about 200 times less than 
the one of $A_2$, and about 40 times less than the one of $A_4$. This schedule is clearly skewed towards $A_2$, which after all is the solver having the worst average runtime in the neighborhood.

As one can expect, the design choices of SUNNY have pros and cons. For example, unlike the CPHydra-like schedule, the schedule produced by SUNNY in Example \ref{ex:sunny} cannot solve the instance $x_2$ although $x_2$ 
is actually part of the neighborhood. More insights on SUNNY are provided in Sect.~\ref{sec:sunny-more}.

By default, SUNNY does not perform any feature selection: it simply removes 
all the features that are constant over each $\F(x)$, and scales the remaining 
features into the range $[-1, 1]$ (scaling features is important for algorithms based on $k$-NN).
The default neighborhood size is $\sqrt{\I_{tr}}$, possibly rounded to the nearest integer. 
The backup solver is the 
solver $A^* \in \mathcal{A}$ minimising the sum $\sum\limits_{i \in \I_{tr}} m(i, A^*)$, 
which is usually
the SBS of the scenario.

The \sunnyas~\cite{paper_cilc} tool implements the SUNNY algorithm to handle 
generic AS scenarios of the \aslib. 
In its optional pre-processing phase, performed offline, 
\sunnyas can perform a feature selection based on different filter
methods and select a pre-solver to be run for a limited 
amount of time. At runtime, it produces the schedule of solvers 
by following the approach explained above.

\subsection{2017 OASC Challenge}
\label{subsec:OASC}

In 2017, the COnfiguration and SElection of ALgorithms (COSEAL) group~\cite{cosealhome} organized the first Open Algorithm Selection Challenge (OASC) to compare
different algorithm selectors.

The challenge is built upon the Algorithm 
Selection library (ASlib)~\cite{bischl2016aslib} which includes a collection of different algorithm 
selection scenarios. ASlib distinguishes between two types of scenarios: 
runtime scenarios and quality scenarios.
In runtime scenarios the goal is to select an algorithm that minimizes the runtime (e.g., for decision problems). The goal in quality scenarios is 
instead to find the algorithm
that obtains the highest score according to some metric (e.g., for optimization 
problems).
ASlib does not consider the \emph{anytime} performance:
the \emph{sub-optimal} solutions computed by an algorithm are not tracked.
This makes it impossible to reconstruct \emph{ex-post} the score 
of interleaved executions. For this reason, in the OASC
the scheduling was allowed only for runtime scenarios.

The 2017 OASC consisted of 11 scenarios: 8 runtime and 3 quality scenarios. 
Differently from
the previous ICON challenge for Algorithm Selection held in 2015,
the OASC used scenarios from a broader domain
which come from the recent international competitions
on CSP, MAXSAT, MIP, QBF, and SAT. In the OASC,
each scenario is evaluated by one pair of training and test set
replacing the 10-fold cross-validation of the ICON challenge. The participants 
had access to performance and feature data on
training instances (2/3 of the total), and only the instance features for the 
test instances (1/3 of the total).

In this paper, since  \sunny produces a schedule of solvers 
not usable for quality scenarios, we focus 
only on runtime scenarios. 
An overview of them with their number of 
instances, 
algorithms, features, and the timeouts is shown in 
Tab.~\ref{table:oascscenarios}.\footnote{Note that Bado and Svea have a different timeout value from the source scenarios.
 To avoid confusion,
 the alias names are used when we intend the dataset of the OASC challenge.}


\begin{table}[t]
\caption{OASC Scenarios.}
\label{table:oascscenarios}
\centering
\resizebox{\columnwidth}{!}{%
\begin{tabular}{|l|l||l|l|l|l|l|}
\hline
Scenario & Source           & Algorithms  & Problems  & Features 
& Timeout (OASC) & Timeout (\aslib) \\ \hline
Caren    & CSP-MZN-2016     & 8              & 66      & 95       & 1200 s & 1200 s\\ 
Mira     & MIP-2016         & 5              & 145      & 143      & 7200 s & 7200 s\\ 
Magnus   & MAXSAT-PMS-2016  & 19             & 400      & 37       & 1800 s & 1800 s\\ 
Monty    & MAXSAT-WPMS-2016 & 18             & 420      & 37       & 1800 s & 1800 s\\ 
Quill    & QBF-2016         & 24             & 550      & 46       & 1800 s & 1800 s\\ 
Bado     & BNSL-2016        & 8              & 786     & 86       & 28800 s & 7200 s\\ 
Svea     & SAT12-ALL        & 31             & 1076     & 115      & 4800  s & 1200 s\\ 
Sora     & SAT03-16 INDU    & 10             & 1333     & 483      & 5000 s & 5000 s\\ 
\hline
\end{tabular}
}
\end{table}

\section{\tool}
\label{sec:sunny-opt}


\tool is the evolution of \sunnyas and the selector that attended the 2017 OASC competition.
The most significant innovation of \tool is arguably the introduction 
of an \emph{integrated} approach where the features and the $k$-value 
are possibly \emph{co-learned} during the training step.
This makes \tool ``less lazy'' than the original SUNNY approach, which only scaled the features in $[-1,1]$ without performing any actual training.\footnote{The ``Y'' of the SUNNY acronym actually stands for laz\textit{y}~\cite{sunny}.} The integrated approach we developed is similar to what 
has been done in \citeA{zyout2011embedded,DBLP:journals/eswa/ParkK15} in the context of biology and medicine. However, to the best of our knowledge, no similar approach has been developed for algorithm selection.

Based on training data, \tool automatically 
selects the most relevant features and/or the most promising value of the 
neighborhood parameter $k$
to be used for online prediction.
We recall that, differently 
from \tool, \sunnyas had only a limited support for filter-based feature 
selection, it only allowed the manual configuration 
of SUNNY parameters, and did not support all the evaluation modalities of the current 
selector.

The importance of feature selection and parameters configuration for SUNNY 
were independently discussed with empirical experiments conducted by
\citeA{DBLP:conf/lion/LindauerBH16,ictai_paper}. In particular, 
\citeA{ictai_paper}
demonstrated the benefits of a filter-based feature selection, while 
\citeA{DBLP:conf/lion/LindauerBH16} highlighted that 
parameters like the schedule size $|\sigma|$ and the neighborhood size $k$
can have a substantial impact on the performance of \sunny. In this regard, 
the authors introduced \tsunny, a version of SUNNY that---by allowing the 
configuration 
of both $|\sigma|$ and $k$ parameters---yielded a remarkable improvement 
over the original \sunny.
Our work is however different because: 
first, we introduce a greedy variant of SUNNY for selecting subset of solvers;
second, we combine wrapper-based feature selection and $k$-configuration, while 
their system does not deal with feature selection.

To improve the configuration accuracy and robustness, 
and to assess the quality of a parameters setting, 
\tool relies on \textit{cross-validation} (CV)~\cite{kohavi1995study}. 
%
Cross-validation is useful to mitigate the well-known problem of \textit{overfitting}.
In this regard it is fundamental to split the dataset properly.
For example, in the OASC only one split between test and training instances 
was used to evaluate the performance of algorithm selectors.
As also noticed  by the OASC organizers \cite{lindauer2019algorithm}, \emph{randomness} 
played an important role in the competition. 
In particular, they stated that \emph{``this result 
demonstrates the 
importance of evaluating algorithm selection
systems across multiple random seeds, or multiple test sets''.}
%
%

To evaluate the performance of our 
algorithm selector by overcoming the overfitting problem and to obtain more robust 
and rigorous results, in this work we adopted a 
\textit{repeated nested cross-validation 
approach}~\cite{loughrey2005overfitting}. \label{nestedrepeat} 
A nested cross-validation consists of two 
CVs, an \emph{outer} CV which 
forms 
test-training
pairs, and an \emph{inner} CV applied on the 
training sets used to learn a model that is later assessed on the 
outer test 
sets. 

The original dataset is split into five folds thus obtaining 
five pairs $(T_1, S_1) \ldots, (T_5, S_5)$ where the 
$T_i$ are the \textit{outer} training sets and the $S_i$
are the (outer) \textit{test sets},
for $i=1,\dots,5$. 
For each $T_i$
we then perform an inner 10-fold CV to get a suitable parameter setting.
We split each $T_i$ into further ten sub-folds 
$T'_{i,1}, \dots, T'_{i,10}$, and in turn for $j=1,\dots,10$ we use a sub-fold $T'_{i,j}$ 
as \textit{validation set} to assess the parameter setting computed with the inner \textit{training set}, which is the union of the other nine sub-folds $\bigcup_{k\neq j}T'_{i,k}$.
We then select, among the 10 configurations obtained, the one for which \sunny achieves the best PAR10 score on the corresponding validation set. The selected configuration is used to run \sunny 
on the paired test set $S_i$. 
%
Finally, to reduce the variability and increase the robustness of our approach, we 
repeated the whole process for five times by using different random partitions. The performance of \tool on each scenario was then 
assessed by considering the average closed gap scores over all the $5 \times 5 = 25$ test sets.

Before explaining how \tool learns features and $k$-value, we first describe \gsunny, the ``greedy variant'' of SUNNY.

\subsection{\gsunny}
\label{sec:gsunny}
The selection of solvers performed by \sunny might be too 
computationally expensive, i.e., exponential in the size of the portfolio in the worst case. 
Therefore, to perform a quicker estimation of the quality of a parameter setting, we 
introduced a simpler variant of \sunny that we called \gsunny.

As for \sunny, the mechanism of schedule generation in \gsunny is driven by the concept of \emph{marginal contribution}~\cite{xu2012evaluating}, i.e., how much a new solver can improve the overall 
portfolio. However, 
\gsunny differs from \sunny in the way the schedule of solvers is 
computed. Given the set $\mathcal{N}$ of the instances of the neighborhood, the original \sunny 
approach
computes the smallest set of solvers in the portfolio that maximizes the 
number of solved instances in $\mathcal{N}$. The worst-case time complexity of 
this procedure is exponential in the number of available solvers.

To overcome this limitation,  
\gsunny starts from an empty set of solvers $S$ and adds to it 
one solver at a time by selecting the one that is able to solve the 
largest number of instances in $\mathcal{N}$. 
The instances solved by the selected 
solver are then removed from $\mathcal{N}$ and the process is repeated until a given number 
$\lambda$ of solvers is added to $S$ 
or there are no more instances to solve
(i.e., $\mathcal{N} = \emptyset$).\footnote{As one can expect, \gsunny does not guarantee that $S$ is the \emph{minimal} subset of solvers solving the 
most instances of $\mathcal{N}$.}
Based on some 
empirical experiments, the default value of $\lambda$ was set to a small value 
(i.e., 3) as also suggested by the experiments in 
\citeA{DBLP:conf/lion/LindauerBH16}. If $\lambda$ is a constant, the 
time-complexity of \gsunny is $O(nk)$ where $k=|\mathcal{N}|$ and $n$ is the 
number of available solvers.

\subsection{Learning the Parameters}
\label{subsec:runmodes}

\tool provides different procedures for learning features and/or the $k$ value.
The configuration procedure 
is performed in two phases:
%
%
%
\emph{(i) data preparation}, and \textit{(ii) parameters configuration}.

\subsubsection{Data Preparation}
\label{sec:data_prep}
The dataset is first split into 5 folds $(T_1, S_1) \ldots, (T_5,S_5)$
for the outer CV, and each $T_i$ is in turn split in $T'_{i,1}, \dots,T'_{i,10}$ 
for the inner CV by performing the following four steps: 1) each 
training instance is associated to the solver that solves it in the shortest time; 
2) for each solver, 
the list of its associated instances is ordered from the hardest to the easiest in terms of runtime; 3) we select one instance at a 
time from each set associated to each solver until a \emph{global limit} on the number of  
instances is reached; 4) the selected instances are finally divided into 10 
folds for cross-validation.\footnote{In the first split, if an instance cannot be solved 
by any of the available solvers it will be discarded as commonly done 
in these cases.}
\label{sec:splitmodes}

At step 4),
\tool offers three choices: \textit{random} split, \textit{stratified} split \cite{kohavi1995study} 
and \textit{rank} split (a.k.a. systematic split in~\citeA{reitermanova2010data}).
The random split simply partitions the instances randomly.
The stratified split guarantees that for each label (in our 
context, the best solver for that instance) all the folds contains 
roughly the same percentage of instances. 
The rank split ranks the instances by their hardness, represented by 
the sum of the runtime,
then each fold takes one instance in turn from the ranked instances. 

While the 
stratified approach distributes the instances based on the best solver able to 
solve them, the rank method tries to distribute the instances based on their hardness.
In the first case, every fold will likely have a witness for every label, 
while in the latter every fold will be a mixture of easy and hard instances.

\subsubsection{Parameters Configuration}
\label{sec:param_config}  
\tool enables the automated configuration of the features and/or the $k$-value
by means of the \gsunny approach introduced in Sect.~\ref{sec:gsunny}.
The user can choose between three different \emph{learning modes}, namely:
\begin{enumerate}
\item {\bf \toolk}. In this case, 
\textit{all} the features are used and only the 
$k$-configuration is performed by varying $k$ in the range 
$[1,\mathit{maxK}]$ where $\mathit{maxK}$ is an external parameter set by the user.
The best value of $k$ is then chosen.
\item {\bf \toolf}. In this case, the neighborhood
size $k$ is set to its default value (i.e., the square root of the number of 
instances, rounded to the nearest integer) and a \textit{wrapper-based} feature 
selection 
is performed. 
Iteratively, starting from the empty set, \toolf adds to the set 
of already selected features
the one which better decreases the $\parten$. 
The iteration stops when the $\parten$ increases or reaches a time cap. 

 \item {\bf \toolfk}. This approach combines both
 \toolf and \toolk: the neighborhood size parameter \textit{and} the set of selected features are configured together by running 
{\bf \toolf} with different values 
of $k$ in the range $[1,\mathit{maxK}]$. The $k$ with the lowest $\parten$ is 
then identified. The entire procedure is repeated until the addition of a 
feature with $k$ varying in $[1,\mathit{maxK}]$ does not improve the $\parten$ score or a 
given time cap is reached.
The resulting feature set and $k$ value are chosen for the online 
prediction.\footnote{
Since \toolfk integrates 
the feature selection into the $k$-configuration process, it may be 
considered as an \emph{embedded} FS method.}
\end{enumerate}

\begin{algorithm}[t]
\begin{algorithmic}[1]
	\Function{learnFK}{$\A$, $\lambda$, $\I$, $\textit{maxK}$, $\F$, $\textit{maxF}$}
	    \State $bestF \gets \emptyset$
	    \State $bestK \gets  1$
	    \State $bestScore \gets -\infty$
		\While{$|bestF| < \textit{maxF}$}
		  \State $currScore \gets -\infty$
	      \For{$f \in \F$} \label{line:for1}
	        \State $currFeatures \gets bestF \cup \{f\}$
	        \For{$k \gets 1, \dots, \textit{maxK}$}
	          \State $tmpScore \gets \textsc{getScore}(\A, \lambda, \I, k, currFeatures)$
	          \If{$tmpScore > currScore$}
	            \State $currScore \gets tmpScore$
	            \State $currFeat \gets f$
	            \State $currK \gets k$	            
	          \EndIf
	        \EndFor
	      \EndFor \label{line:for2}
	      \If{$currScore \leq bestScore$} \Comment{Cannot improve the best score}
	        \State\textbf{break}
	      \EndIf
	      \State $bestScore \gets currScore$	      
	      \State $bestF \gets bestF \cup \{currFeat\}$
	      \State $bestK \gets currK$
	      \State $\F = \F - \{currFeat\}$  
		\EndWhile
	\State\Return{$bestF$, $bestK$}
	\EndFunction
\end{algorithmic}
\caption{Configuration procedure of \toolfk.}\label{alg:sunnyfk}
\end{algorithm}

Algorithm~\ref{alg:sunnyfk} shows through pseudocode how \toolfk selects the features and the $k$-value.
The \textsc{learnFK} algorithm takes as input the portfolio of algorithms $\A$, 
the maximum schedule size $\lambda$ for \gsunny, the set of training instances 
$\I$, the maximum neighborhood size $\textit{maxK}$,
the original set of features $\F$, and the upper bound $\textit{maxF}$ on the maximum number 
of features to be selected.
\textsc{learnFK} returns the learned value $bestK \in [1,maxK]$ for the neighborhood size and the learned set 
 of features $bestF \subseteq \F$ having $|bestF| \leq maxF$.
 
After the $i$-th iteration of the outer \emph{for} loop 
(Lines~\ref{line:for1}--\ref{line:for2}) 
the current set of features $currFeatures$ consists of exactly $i$ features. Each time $currFeatures$ is 
set, the inner \emph{for} loop is executed $n$ times to also evaluate different values of $k$ 
on the dataset $\I$. The evaluation is performed by the function 
\textsc{getScore}, returning the score of a particular setting obtained with 
\gsunny (cf. Sect.~\ref{sec:gsunny}). However, \textsc{getScore} can be easily generalized to assess 
the score of a setting obtained with an input algorithm different from \gsunny 
(e.g., the original \sunny approach).


At the end of the outer \emph{for} loop, if adding a new feature could not improve the score obtained in the previous iteration (i.e., with $|currFeatures|-1$ features) the learning process terminates.
Otherwise, both the features and the $k$-value are updated and a new iteration begins, until the score 
cannot be further improved or the maximum number of features $\textit{maxF}$ is reached.

If $d=\min(\textit{maxF}, |\F|)$, 
$n = \min(\textit{maxK}, |\I|)$ and the worst-case time 
complexity of \textsc{getScore} is $\gamma$, then the overall worst-case time 
complexity of \textsc{learnFK} is $O(d^2 n\gamma)$.
This cost is still polynomial w.r.t. $|\A|$, $d$, and $n$ because 
\textsc{getScore} is polynomial thanks to the fact that $\lambda$ is 
a constant.


From \textsc{learnFK} one can easily deduct the algorithm for learning either the $k$-value (for \toolk) or the selected features (for \toolf): in the first case, the outer \emph{for} loop is omitted because features do not vary; in 
the second case, the inner loop is skipped because the value of $k$ is constant.

We conclude this section by summarizing the input parameters that, 
unlike features and $k$-value, are \emph{not} learned automatically 
by \tool:
\begin{enumerate}
\item {\bf split mode:} the way of creating validation folds for the inner CV, including:
random, rank, and stratified split. Default: rank.
\item {\bf training instances limit:} the maximum number of instances
used for training. Default: 700.
\item {\bf feature limit:} the maximum number of features for feature 
selection, 
used by \toolf and \toolfk. Default: 5. 

\item {\bf k range:} the range of neighborhood sizes used by both  
\toolk and \toolfk. Default: [1,30].
\item {\bf schedule limit for training ($\lambda$):} the limit of the schedule 
size 
for \gsunny. Default: 3.
\item {\bf seed:} the seed used to split the training set into folds. Default: 
100.
\item {\bf time cap:} the time cap used  by \toolf and 
\toolfk to perform the training. Default: 24 h.
\end{enumerate}

The default values of these parameters were decided by conducting an 
extensive set of manual experiments over ASlib scenarios, with the goal of 
reaching a good trade-off between the performance and the time needed for the 
training phase (i.e., at most one day). In Sect.~\ref{sec:tuning} we shall report some of these 
experiments.

\section{Experiments}
\label{sec:experiments}

In this section we present (part of) the experiments we conducted over several different 
configurations of \tool.

We first present the benchmarks and the methodology used (Sect. 
\ref{sec:setting}).
Then, in Sect. \ref{sec:assess} we assess the impact of the new components of \tool 
to quantify 
what we can gain by learning the neighborhood size, by using a smaller number of features, and 
by using \gsunny instead of, or together with, the original \sunny approach.
Finally, in Sect.~\ref{sec:tuning} we use as baseline \toolfk, i.e., \tool's most comprehensive approach 
that exploits both
the learning of the neighborhood size and the feature selection, to 
understand how its performance can vary by tuning one parameter at a time 
and by leaving the other parameters to their default values.

In the following, unless otherwise specified, \tool always denotes the \toolfk 
variant.


\subsection{Experimental Setting}
\label{sec:setting}

\begin{table}[t]
\caption{Additional \aslib scenarios.}
\label{table:aslibscenarios}
\centering
\resizebox{\columnwidth}{!}{%
\begin{tabular}{|l|l||l|l|l|l|}
\hline
Scenario & Source           & Algorithms ($m$)  & Problems ($n$) & Features 
($d$) & Timeout ($\tau$)\\ \hline
ASP  & ASP-POTASSCO        & 11     & 1212      & 138     & 600 s\\ 
CPMP     & CPMP-2015    		& 4      & 555       & 22      & 3600 s\\ 
GRAPHS   & GRAPHS-2015         & 7      & 5725     & 35      & 100000000 s\\ 
TSP  & TSP-LION2015        & 4      & 3106     & 122       & 3600 s\\ 

\hline
\end{tabular}
}
\end{table}

We evaluated \tool on the runtime scenarios of the ASlib.
In particular, we selected the 8 runtime scenarios of the OASC challenge described 
in Sect.~\ref{subsec:OASC} (see Tab.~\ref{table:oascscenarios}). These  
scenarios contain problem instances belonging to the following domains: Constraint Satisfaction, Mixed-Integer Programming, SAT solving, Max-SAT solving, Quantified Boolean Formulas, and learning in Bayesian networks.
To avoid biases towards a specific domain,\footnote{
ASlib scenarios are skewed towards SAT problems: almost half of them are based on SAT or Max-SAT.} we added four more  
ASlib scenarios representing all those domains that were not considered in the OASC, namely: Answer Set Programming, Pre-marshalling problem, 
Subgraph Isomorphisms, and Traveling Salesman Problem (see Tab.~\ref{table:aslibscenarios}).

We used the 
repeated nested cross-validation with 5 repetitions, 5 folds in the outer loop and 10 folds in the inner loop, explained in Sect. \ref{sec:sunny-opt}.
For the OASC scenarios, we used only the instances belonging to 
the training set of the OASC 
competition since we later on wanted to check the performance of the last version of \tool on 
the OASC test sets. For the four additional scenarios, since they did 
not come with a separation between training and test sets, we instead applied 
the 
repeated cross-validation
on all their instances.
For each 
scenario, the performance of \tool was evaluated with 
the average closed gap score over all the 
repetitions. In each repetition, the closed gap score was calculated 
as explained in Sect.~\ref{sec:eval} by using the $\parten$ as performance metric $m$.

All the experiments were conducted on Linux machines equipped 
with Intel Core$i5$ 3.30GHz processors and 8 GB of RAM. We used a time cap of 24 hours for learning the 
parameters. All the ASlib 
scenarios are publicly available at \url{https://github.com/coseal/aslib_data}.

\subsection{Assessment of New Components}
\label{sec:assess}
In this section we measure the impact of the new components we introduced in this paper.
We assess what we can gain by learning the neighborhood size and/or the number of features, and 
how \gsunny can improve the original \sunny algorithm.

\subsubsection{Learning Modes}

We compared the \toolf, \toolk, and \toolfk variants of \tool against the original version of \sunnyas
that does not exploit any parameter configuration.

\begin{table}[ht]
\caption{Comparisons of \tool learning modes in terms of closed gap.}
\label{fig:modalities}
    \resizebox{\linewidth}{!}{%
    \pgfplotstabletypeset[
        precision=4,
        fixed zerofill, 
        col sep=comma,
        row sep=crcr,    
         color cells={min=0,max=1},    
        /pgfplots/colormap={whiteblue}{rgb255(0cm)=(255,255,255); rgb255(1cm)=(\tableColorR,\tableColorG,\tableColorB)},
        every head row/.style={%
            before row=\hline,%
            after row=\hline%
        },%
        columns/Approach/.style={reset styles, string type},
        every last column/.style={column type/.add={}{|}},
        every first column/.style={
        		column type={|l|}
            },
        every last column/.style={
                column type/.add={}{|},
                column type/.add={|}{}
        },
        every last row/.style={after row=\hline},
    ]{
			Approach,Caren,Mira,Magnus,Monty,Quill,Bado,Svea,Sora,ASP,CPMP,GRAPHS,TSP,Average\\
			sunny,-0.0517,-0.1289,0.6343,0.4291,0.6976,0.7854,0.6458,0.1781,0.6674,0.7488,0.6968,-0.4457,0.4047\\
			sunny-as2-f,-0.0603,-0.1649,0.4425,0.4489,0.6854,0.7695,0.5783,0.2459,0.7717,0.7771,0.5663,-0.1058,0.4129\\
			sunny-as2-k,0.1611,0.0276,0.6352,0.583,0.7361,0.7976,0.6915,0.3591,0.7193,0.7273,0.6504,-0.8822,0.4338\\
			sunny-as2-fk,0.0845,-0.1891,0.4458,0.5846,0.7139,0.759,0.6643,0.3428,0.7454,0.7885,0.5614,-0.0343,0.4556\\
}
}

\end{table}

\begin{figure}[t]
    \centering
        \includegraphics[width=1\linewidth]{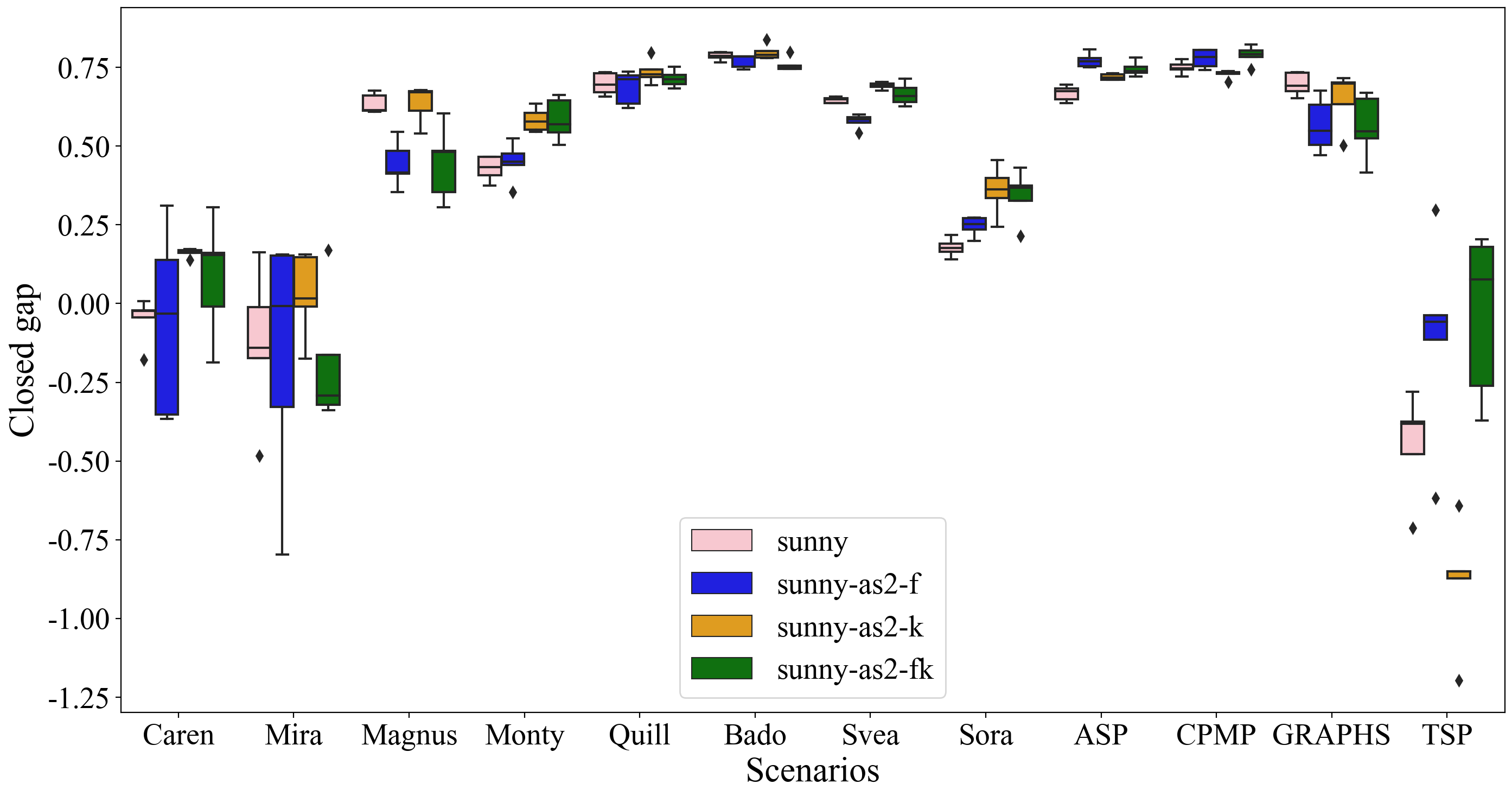}
    \caption{Close gap distribution of various learning modes.\label{boxfig:modalities}}
\end{figure}

We run the different \tool learning modes with their default parameters for all the scenarios.
Tab.~\ref{fig:modalities} shows the average closed gap of each approach across all the repetitions 
performed.
Interestingly, there is not a dominant 
learning mode. As also shown in 
\citeA{DBLP:conf/lion/LindauerBH16}, a proper $k$-configuration 
leads to a good performance improvement for SUNNY---indeed, \toolk is able 
to reach the peak performance in 7 scenarios out of 12.
However, \toolfk has the best average closed gap. One reason for this is the poor 
performance of \toolk in the TSP scenario.

The original \sunnyas is clearly less promising than any other variant of \tool, even 
though for the GRAPHS scenario it achieves the best performance.
What we can conclude from Tab.~\ref{fig:modalities} is that most of the 
performance improvement is due to the selection of the right neighborhood size 
$k$. However, feature selection can also give a positive contribution.

Fig.~\ref{boxfig:modalities} depicts with boxplots the closed gap scores reported in 
Tab.~\ref{boxfig:modalities}. Specifically, for each scenario we collected the corresponding 
25 closed gap scores, one for each test set.  
Each box of Fig. \ref{boxfig:modalities} delimits the first and third quartile of 
the closed gap distribution, while the horizontal line inside each box is the median.
The vertical whiskers  
indicate the rest of the distribution excluding diamonds, which are considered as outliers
since they are outside the inter-quartile ranges.
The larger the box, the less stable a learning system is.
For example, 
\toolf is quite unstable in Caren, Mira and Graphs scenarios.
\toolfk looks more robust than \toolf,
while \toolk and \sunnyas seems to be slightly more stable in most cases.


Tab.~\ref{fig:modalities_time} shows the average time (in minutes) spent for 
training each fold. As we can see, 
\toolk is the fastest approach, followed by \toolf and 
\toolfk. This is not surprising because learning features is a computationally expensive 
task, especially when wrapper methods are used.
\begin{table}[ht]
    \caption{Average training time in minutes of \tool learning modes.}
    \label{fig:modalities_time}
	\resizebox{\columnwidth}{!}{%
		\begin{tabular}{|l|llllllllllll|}
			\hline
			Approach     & Caren & Mira  & Magnus & Monty & Quill & Bado   & Svea   & Sora    & ASP    & CPMP & GRAPHS & TSP    \\
			\hline
			sunny-as2-f  & 0.1   & 0.47  & 0.73   & 0.66  & 2.51  & 7.41   & 29.29  & 105.69  & 32.64  & 0.49 & 160.12 & 161.03 \\
			sunny-as2-k  & 0.02  & 0.12  & 0.25   & 0.26  & 0.7   & 2.39   & 5.84   & 25.15   & 8.81   & 0.38 & 58.67  & 50.64  \\
			sunny-as2-fk & 3.34  & 10.53 & 14.54  & 16.73 & 45.56 & 156.83 & 321.15 & 1174.54 & 329.77 & 9.82 & 73.92  & 222.05 \\ 
			\hline
		\end{tabular}
	}
\end{table}

%

\subsubsection{\gsunny vs \sunny}
\label{sec:gsunny-sunny}

As mentioned in Sect. \ref{sec:gsunny}, \gsunny was introduced to speed up the training 
process. Here we empirically show that 
\gsunny not only speeds up the training of \tool, but it also 
\textit{outperforms} the performance 
achieved by using the original SUNNY approach for training.

In the following experiments we use \tool with its default parameters, by only 
varying the approach adopted for generating 
the schedule of solvers for both training and 
testing.
In the latter case, 
we 
use a time limit of a week due to the long computation time required by the 
original SUNNY approach to create schedules.

Note that SUNNY and \gsunny are interoperable because they share the same underlying AS approach. The only difference is the way they select the subsets of solvers: \gsunny uses a possibly not optimal greedy approach, while SUNNY relies on an exhaustive 
search---possibly exponential in the portfolio size.
The output of the \textsc{learnFK} algorithm (Algorithm \ref{alg:sunnyfk}) is always a set of features $F$ and a neighborhood size $k$, regardless of whether SUNNY or \gsunny is used on the validation set. These parameters are then used to compute the schedules on the unforeseen test instances, regardless of whether SUNNY or \gsunny is used to select the solvers.

The results are reported in Tab.~\ref{fig:sunny_gsunny}. Column 
names denote the pairs of functions used for the 
training and 
testing
respectively. For brevity, we write GSUNNY instead of \gsunny. 
For instance, the second column 
``SUNNY-GSUNNY'' 
means
that \sunny has been used for training, and \gsunny for 
testing.



\begin{table}[ht]
	\caption{Closed gap of different combinations of \sunny/\gsunny 
		for training/testing.\label{fig:sunny_gsunny}}
	\center
	\resizebox{0.9\linewidth}{!}{%
		\pgfplotstabletypeset[
		precision=4,
		fixed zerofill, 
		col sep=comma,
		row sep=crcr,    
		every first column/.style={
			column type={|l|}
		},
		color cells={min=0,max=1},    
		/pgfplots/colormap={whiteblue}{rgb255(0cm)=(255,255,255); rgb255(1cm)=(\tableColorR,\tableColorG,\tableColorB)},
		every head row/.style={%
			before row=\hline,%
			after row=\hline%
		},%
		columns/Scenario/.style={reset styles, string type},
		every last column/.style={
			column type/.add={}{|}
		},
		every last row/.style={
			before row=\hline,%
			after row=\hline
		},
		every row 12 column 8/.style={
			postproc cell content/.append style={
				@cell content/.add={$\bf}{$}
			}
		},
		every row 6 column 1/.style={reset styles, string type},
		every row 6 column 2/.style={reset styles, string type},
		]{%
			Scenario,SUNNY-SUNNY,SUNNY-GSUNNY,GSUNNY-SUNNY,GSUNNY-GSUNNY\\
			Caren,0.0156,-0.0182,0.0845,0.0845\\
			Mira,-0.1907,-0.2519,-0.1891,-0.2502\\
			Magnus,0.4701,0.4692,0.4458,0.4449\\
			Monty,0.5652,0.5655,0.5846,0.5784\\
			Quill,0.6966,0.6823,0.7139,0.6989\\
			Bado,0.7381,0.7324,0.759,0.7496\\
			Svea,Timeout,Timeout,0.6643,0.6587\\
			Sora,0.305,0.2763,0.3428,0.3212\\
			ASP,0.7488,0.7386,0.7454,0.7402\\
			CPMP,0.7624,0.7674,0.7885,0.7809\\
			GRAPHS,0.5948,0.5942,0.5614,0.5613\\
			TSP,0.0297,-0.014,-0.0343,-0.0576\\
			All,0.3946,0.3785,0.4556,0.4426\\
		}
	}
	
\end{table}

Tab.~\ref{fig:sunny_gsunny} shows that the peak performance in each scenario is always reached when SUNNY 
is used for testing. This makes sense: using \gsunny on an unforeseen instance 
might be useful 
in a time-sensitive context where an exponential-time solvers' selection is not 
acceptable but, in general, SUNNY provides a more precise scheduling.
However, the GSUNNY-GSUNNY column shows that on average the score of \tool using \gsunny only is not 
far from the best performance achieved by GSUNNY-SUNNY.

The most interesting thing of Tab.~\ref{fig:sunny_gsunny} is probably that, on average, using \gsunny for learning the features and the $k$ value not only speeds up the training but also improves the prediction 
accuracy. Indeed, the score achieved by GSUNNY-SUNNY and GSUNNY-GSUNNY is consistently better than 
the one of SUNNY-SUNNY and SUNNY-GSUNNY.
We 
conjecture that, in the training phase, it 
might be
more important to prioritise 
the first $\lambda$ solvers solving the most instances in the neighborhood rather than selecting 
a sub-portfolio from all the available solvers as done by SUNNY (we recall that the maximum schedule size $\lambda$ for the default 
\gsunny is 3).


\gsunny can be particularly useful on scenarios 
with a large number of solvers. This is evident in Tab.~\ref{fig:speed_sunnys} 
describing the hours spent 
for training using the different approaches.\footnote{Based on our estimation,  
	the 
	Svea scenario would have taken about 17000 hours to be 
	completed.}
As expected, \gsunny is quicker than the original \sunny approach for any considered
scenarios.

\begin{table}[ht]
	\caption{Hours spent for training by using \gsunny and SUNNY.}
	\label{fig:speed_sunnys}
	\resizebox{\columnwidth}{!}{%
		\begin{tabular}{|l|llllllllllll|}
			\hline
			scenario & Caren & Mira & Magnus & Monty & Quill & Bado & Svea & Sora  & ASP & 
			CPMP & GRAPHS & TSP \\ \hline
			
			GSUNNY   & 0.06  & 0.18 & 0.24   & 0.28  & 0.76  & 2.61 & 5.35   & 19.58 & 5.5  & 0.16 & 1.23   & 3.7  \\
			SUNNY   & 0.27  & 0.2  & 0.55   & 1.09  & 21.57 & 3.54 & Timeout & 35.23 & 8.95 & 0.2  & 1.49   & 4.53
			
			\\ \hline \hline
			\# solvers & 20    & 5     & 19    & 18    & 24    & 8    & 31    & 10    &  11 
			& 4 & 7 & 4          \\ \hline 
			\# insts & 54    & 145     & 370    & 357    & 512    & 700    & 700    & 700    
			&   700 & 555 & 700 & 700        \\ \hline 
			\# features & 95    & 143     & 37    & 37    & 46    & 86    & 115    & 483    
			&   138 & 22 & 35 & 122        \\ \hline 
			
		\end{tabular}
	}
	
\end{table}

\subsection{Tuning the Parameters}
\label{sec:tuning}
In this section we study the sensitivity of the parameters that \tool cannot learn, namely:
the split modes for cross-validation, the limit on the numbers of features to select, 
the limit on the number of training instances, and finally the schedule limit $\lambda$.
We conclude the section by reporting an analysis on the performance variability of \tool.

For all the experiments, we set the 
parameters of \tool to their default values and we varied one 
parameter at a time. 
We mark with 
`Timeout' the cases where the training phase
for at least one fold 
did not finish within a day.
When a training timeout occurs for a specific scenario, we assign to it a closed gap score of 0, i.e., the score of the single best solver. In other terms, if we cannot train a scenario within 24 hours we simply assume that the single best solver is used for that scenario.

\subsubsection{Cross-Validation}

We study the effects of different cross-validations when training
the model. Tab.~\ref{fig:split_mode} compares different cross-validation 
approaches for all the scenarios in our benchmark. For these experiments we set 
the internal parameters of \tool to their 
default values (cf. Sect. \ref{sec:param_config})
except the split mode one.



\begin{table}[ht]
\caption{Random/Stratified/Rank Cross-Validation comparison in terms of closed gap.}
\label{fig:split_mode}
    \resizebox{\linewidth}{!}{%
    \pgfplotstabletypeset[
        precision=4,
        fixed zerofill, 
        col sep=comma,
        row sep=crcr,    
         color cells={min=0,max=1},    
        /pgfplots/colormap={whiteblue}{rgb255(0cm)=(255,255,255); rgb255(1cm)=(\tableColorR,\tableColorG,\tableColorB)},
        every head row/.style={%
            before row=\hline,%
            after row=\hline%
        },%
        columns/Mode/.style={reset styles, string type},
        every last column/.style={column type/.add={}{|}},
         every first column/.style={
        		column type={|l|}
            },
        every last row/.style={after row=\hline},
    ]{%
		Mode,Caren,Mira,Magnus,Monty,Quill,Bado,Svea,Sora,ASP,CPMP,GRAPHS,TSP,Average\\
		random,-0.1092,-0.0121,0.5326,0.5796,0.7077,0.7588,0.6409,0.2992,0.7655,0.7832,0.4678,-0.1417,0.4394\\
		stratified,-0.2196,-0.2269,0.6097,0.5508,0.7253,0.7786,0.6446,0.3317,0.7653,0.7994,0.5406,-0.1489,0.4292\\
		rank,0.0845,-0.1891,0.4458,0.5846,0.7139,0.759,0.6643,0.3428,0.7454,0.7885,0.5614,-0.0343,0.4556\\
}
}
\end{table}

\begin{figure}[t]
    \centering
        \includegraphics[width=1\linewidth]{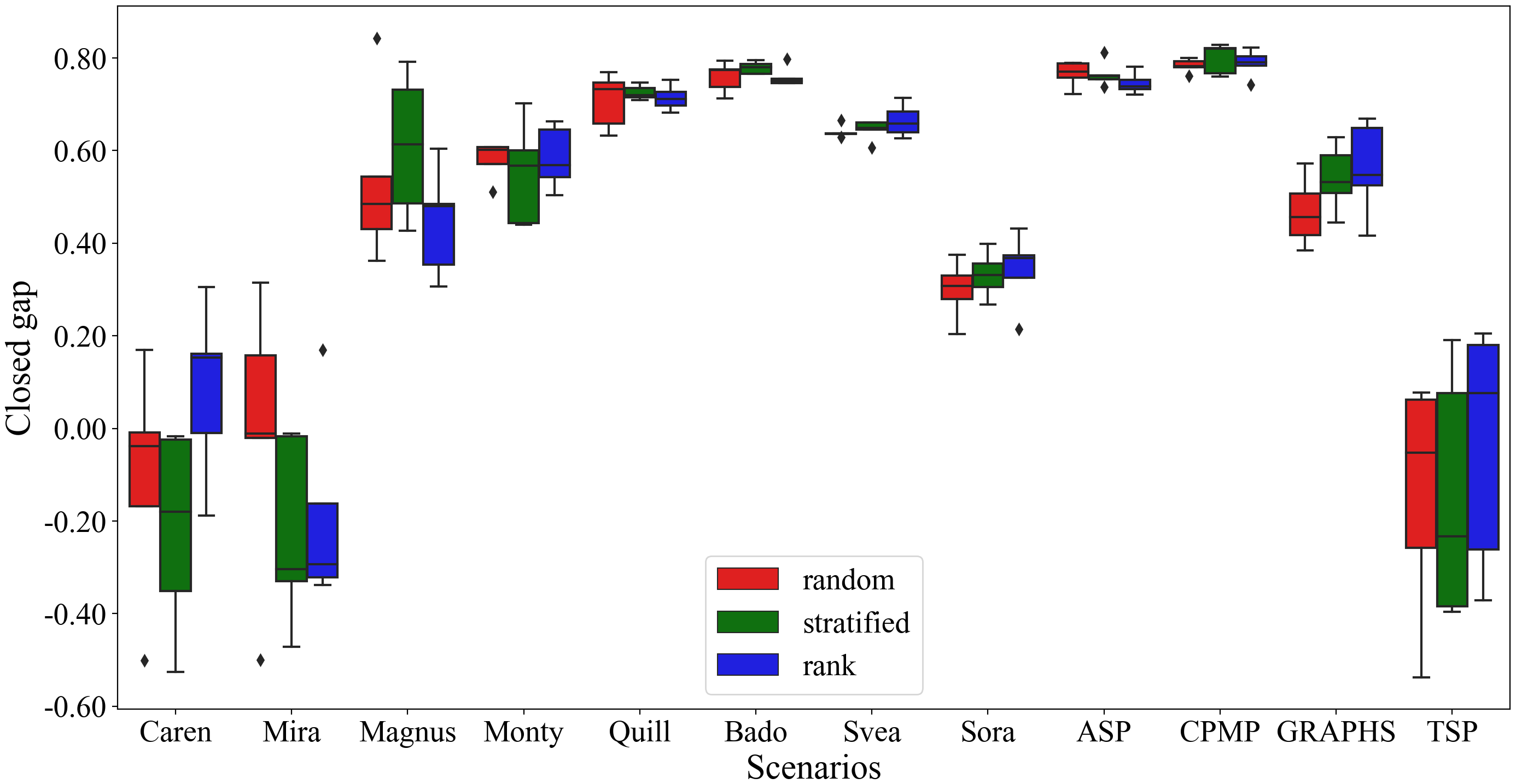}
    \caption{Close gap distribution of various split modes.\label{boxfig:split_mode}}
\end{figure}

The three split modes we examined are:
 \emph{random}, \emph{stratified} and 
\emph{rank}. 
The \emph{random} mode generates folds in a random way;
the \emph{stratified} mode 
generates folds based on class label (fastest algorithm);
the \emph{rank} first orders the instances according to their hardness 
(cf.~\ref{sec:splitmodes}), then systematically partitions them into each fold.

As shown in Tab.~\ref{fig:split_mode}, 
the closed gap of rank CV is on average better than both random and stratified CV.
It appears that distributing instances according to their hardness
leads to more balanced folds, and this in turn implies a better training.
However, there is not a single dominant approach: stratified is the best in four scenarios, rank 
in six, and random in only two scenarios. 
It appears that stratified CV performs better than rank in scenarios
with a higher number of instances.

Fig.~\ref{boxfig:split_mode}
shows the boxplots of Tab.~\ref{fig:split_mode}.
We can see that the rank mode looks more stable than 
random and stratified modes in most scenarios except Svea. 

\subsubsection{Number of Training Instances}

Here we study the impact of the number of training instances.
As above, we fixed the default parameter values listed in Sect. 
\ref{sec:param_config}, and we just varied the limit of training instances.

It is worth noting that, as detailed in the procedure of data preparation (cf. 
Sect.~\ref{subsec:runmodes}), when the limit 
is below the total amount of instances of a scenario, the instances are not selected 
randomly but chosen according to their best solvers and their hardness in order 
to have a more representative training set.

\begin{table}[ht]
\caption{Closed gap by varying the number of training instances.}
\label{fig:training_varying_inst}
    \resizebox{\linewidth}{!}{%
    \pgfplotstabletypeset[
        precision=4,
        fixed zerofill, 
        col sep=comma,
        row sep=crcr,    
         color cells={min=0,max=1},    
        /pgfplots/colormap={whiteblue}{rgb255(0cm)=(255,255,255); rgb255(1cm)=(\tableColorR,\tableColorG,\tableColorB)},
        every head row/.style={%
            before row=\hline,%
            after row=\hline%
        },%
        columns/Scenario/.style={reset styles, string type},
        every last column/.style={
                column type/.add={}{|},
                column type/.add={|}{}
        },
        every first column/.style={
        		column type={|l|}
        },
        every last row/.style={
            before row=\hline,%
            after row=\hline
        },
        every row 7 column 9/.style={reset styles, string type},
        every row 7 column 10/.style={reset styles, string type},
        every row 7 column 11/.style={reset styles, string type},
        every row 7 column 12/.style={reset styles, string type},
        every row 10 column 12/.style={reset styles, string type},
        every row 11 column 12/.style={reset styles, string type},
    ]{%
            Scenario,100,150,200,300,400,500,600,700,800,900,1000,All\\
            Caren,,,,,,,,,,,,0.0845\\
            Mira,-0.1562,,,,,,,,,,,-0.1891\\
            Magnus,0.4935,0.3856,0.5216,0.4458,,,,,,,,0.4458\\
            Monty,0.4381,0.4947,0.5235,0.5846,,,,,,,,0.5846\\
            Quill,0.7599,0.7781,0.7812,0.7641,0.7046,0.7139,,,,,,0.7139\\
            Bado,0.7082,0.7574,0.7815,0.7673,0.779,0.7177,0.7361,0.759,,,,0.759\\
            Svea,0.2845,0.3862,0.4643,0.5671,0.6097,0.6222,0.6354,0.6643,0.6676,0.6563,,0.6563\\
            Sora,0.0451,0.1701,0.2104,0.2139,0.2395,0.2672,0.2998,0.3428,Timeout,Timeout,Timeout,Timeout\\
            ASP,0.6741,0.7135,0.7233,0.744,0.7178,0.7436,0.7494,0.7454,0.77,0.7873,,0.7873\\
            CPMP,0.796,0.7929,0.797,0.8073,0.7748,,,,,,,0.7885\\
            GRAPHS,0.3423,0.4318,0.5416,0.5335,0.4271,0.3999,0.3998,0.5614,0.5948,0.5391,0.5609,Timeout\\
            TSP,-0.4957,-0.3539,-0.1172,-0.0808,0.0612,0.0792,-0.0389,-0.0343,0.0103,-0.2041,-0.1143,Timeout\\
            Average,0.3322  ,0.371  ,0.4269 , 0.4369  ,0.4366 ,0.4382 ,0.4342,0.4556,0.4358,0.4138,0.4231,0.3859\\
}
}

\end{table}

We run \tool with different instance limits starting from 100 (the 
smallest scenario has less than 100 instances) with increments of 100, with a time cap 
of 1 day of computation per fold.

Tab.~\ref{fig:training_varying_inst} presents the average closed gap scores for 
experiments with up to 1000 instances. 
The last column reports the results achieved by considering \emph{all} the 
training instances, while the other columns contain the results 
achieved by considering a fixed number of training instances 
(i.e., $100, 200, \dots, 1000$).
The last
row reports the average closed gap score across all scenarios. In case a 
scenario has less instances than required, 
we simply consider 
all of them.

We omit the results for GRAPHS and TSP scenarios with more than 1000 instances, 
since their closed gaps are below the maximal value reached  
with 800 instances for GRAPHS and with 500 instances for the TSP 
scenario.
The GRAPHS scenario 
timeouts with 2500 instances while TSP 
timeouts with 1500 instances.

We can note that by reducing the number of training instances 
the closed gap of \tool 
does not worsen significantly.
After 200 instances, increasing the number of training instances does 
have a limited impact:
the score oscillates around $0.41$ and $0.46$.
The best average score of $0.4556$ is obtained with $700$ instances.
We conjecture that this is partially due to the procedure for data preparation 
(cf. Sect.~\ref{sec:data_prep}) that 
picks the instances after 
ordering them by hardness, thus reducing the folds skewness. 
The resulting set of instances is large enough to form a homogeneous set reflecting 
the instance class distribution of the entire scenario even after a random or 
stratified split. Adding more instances is not always beneficial. First, 
a large number of training instances deteriorates the 
running time performance of the $k$-NN algorithm on which SUNNY relies 
producing a slowdown of the 
solver selection process for both SUNNY and \gsunny. Second, it can 
also cause a degradation of performances. Probably this is due to the fact that 
more 
instances can introduce additional noise that impacts the selection of solvers 
by \tool.

\subsubsection{Number of Features}

It is well established that a small number of features is often enough to provide a
competitive 
performance for an AS system---and a machine learning system in general.
For example,
according to the reduced set analysis performed by 
\citeA{bischl2016aslib}, no AS scenario required more than 9 features.
To better understand the impact of the number of features we run \tool with a feature limit 
from one to ten (i.e., $\textit{maxF} \in [1,10]$ when calling the \textsc{learnFK} function shown in 
Algorithm \ref{alg:sunnyfk}), and by
leaving the other parameters set to their 
default values as specified in Sect.~\ref{sec:param_config}.

\begin{table}[ht]
\caption{Closed gap of \tool by varying the feature limit.}
    \label{fig:nfeats}
    \resizebox{\linewidth}{!}{%
    \pgfplotstabletypeset[
        precision=4,
        fixed,
        fixed zerofill, 
        col sep=comma,
        row sep=crcr,    
         color cells={min=0,max=1},    
        /pgfplots/colormap={whiteblue}{rgb255(0cm)=(255,255,255); rgb255(1cm)=(\tableColorR,\tableColorG,\tableColorB)},
        every head row/.style={%
            before row=\hline,%
            after row=\hline%
        },%
        columns/Scenario/.style={reset styles, string type},
        every last column/.style={
                column type/.add={}{|},
                column type/.add={|}{}
        },
        every first column/.style={
        		column type={|l|}
            },
        every last row/.style={
            before row=\hline,%
            after row=\hline
        },
        every row 7 column 9/.style={reset styles, string type},
        every row 7 column 10/.style={reset styles, string type},
        every row 7 column 11/.style={reset styles, string type},
        every row 7 column 8/.style={reset styles, string type},
        every row 7 column 7/.style={reset styles, string type},
        every row 12 column 5/.append style={
	        postproc cell content/.append style={
	          @cell content/.add={$\bf}{$} 
	        }
      	},
    ]{%
            Scenario,1,2,3,4,5,6,7,8,9,10,All\\
			Caren,0.1811,0.0579,-0.0124,0.083,0.0845,0.1157,0.1156,0.115,0.0483,0.0803,0.0803\\
			Mira,-0.2221,-0.2292,-0.2541,-0.2261,-0.1891,-0.1919,-0.2278,-0.2248,-0.2242,-0.2226,-0.2295\\
			Magnus,0.4145,0.4753,0.5225,0.4962,0.4458,0.4702,0.4698,0.4707,0.4707,0.4824,0.4824\\
			Monty,0.4373,0.4791,0.5637,0.553,0.5846,0.5643,0.5771,0.5723,0.566,0.5726,0.5785\\
			Quill,0.6726,0.6898,0.7097,0.7208,0.7139,0.7186,0.7159,0.7159,0.7159,0.7159,0.7159\\
			Bado,0.6841,0.7561,0.762,0.7482,0.759,0.7413,0.7559,0.7576,0.7599,0.7578,0.7578\\
			Svea,0.3946,0.5526,0.6044,0.6425,0.6643,0.6601,0.6604,0.668,0.668,0.6623,0.6639\\
			Sora,0.1873,0.2583,0.317,0.3251,0.3428,0.3492,Timeout,Timeout,Timeout,Timeout,Timeout\\
			ASP,0.704,0.7295,0.7531,0.754,0.7454,0.7441,0.748,0.7532,0.7533,0.7533,0.7534\\
			CPMP,0.7909,0.7816,0.7905,0.7905,0.7885,0.7885,0.7885,0.7885,0.7885,0.7885,0.7885\\
			GRAPHS,0.3611,0.5135,0.5517,0.5281,0.5614,0.5491,0.5408,0.5364,0.5321,0.5321,0.5321\\
			TSP,-0.009,-0.0364,-0.0706,-0.0389,-0.0343,-0.1047,-0.0845,-0.1287,-0.1518,-0.1045,-0.1272\\
			Average,0.383,0.419,0.4365,0.448,0.4556,0.4504,0.4216,0.4187,0.4106,0.4182,0.4163\\
}
}
\end{table}

Tab.~\ref{fig:nfeats} shows the closed gap results.
As we can see, often the highest performance
was reached with a limited amount of features, 
and in no scenario the best performance was exclusively achieved with the original feature set.
Although there is not a dominant value for all the scenarios, the overall 
average score is achieved when the 
limit of features is set to five. 

\subsubsection{Schedule Size for \gsunny}
\label{sec:sched_size}
In the training process, \gsunny uses a parameter $\lambda$ to limit the
size of the generated schedule and to be faster than the original \sunny approach when 
computing the schedule of solvers. 
We then
investigated the performance of \tool by varying the $\lambda$ 
parameter of \gsunny (see Algorithm \ref{alg:sunnyfk} in Sect.~\ref{sec:sunny-opt}).


One thing to note before introducing the impact of varying $\lambda$ for \gsunny 
is that, in general, the original SUNNY approach does not produce 
large schedules. This can be seen, e.g., in Tab.~\ref{fig:original_inspection},
reporting the average size of the schedules found by the original \sunny approach and its 
standard deviation when using a 5-fold cross-validation to train \sunny. 
Despite no limit on the schedule size is given, \sunny tends to produce schedules 
with an average size that varies from one to three,
generally around two. This happens because \sunny aim to selects the \emph{smallest}
subset of solvers solving the most instances in the neighborhood.

\begin{table}[ht]
\caption{\sunny's average schedule size with standard 
deviation.}
\label{fig:original_inspection}
\centering
\resizebox{\columnwidth}{!}{%
\begin{tabular}{|l|llllllllllll|}
\hline
              			  & Caren  & Mira & Magnus  & Monty    & Quill   & Bado  & 
Svea   & Sora   & ASP & CPMP &  GRAPHS & TSP    \\
\hline
Schedule size ($\lambda$) & 1.9$\pm$0.4 & 1.5$\pm$0.5 & 1.9$\pm$0.4 & 
1.8$\pm$0.6 & 1.9$\pm$1.1 & 2.0$\pm$0.7 & 3.0$\pm$1.0 & 2.4$\pm$0.7 & 
2.1$\pm$0.7 & 2.4$\pm$0.8 & 1.1$\pm$0.2 & 1.3$\pm$0.5 \\
\hline

\end{tabular}
} 
\end{table}


\begin{table}[ht]
\caption{Closed gap by varying the schedule size of \gsunny.}
    \label{fig:training_schedule_size}
    \resizebox{\linewidth}{!}{%
    \pgfplotstabletypeset[
        precision=4,
        fixed zerofill, 
        col sep=comma,
        row sep=crcr,    
         color cells={min=0,max=1},    
        /pgfplots/colormap={whiteblue}{rgb255(0cm)=(255,255,255); rgb255(1cm)=(\tableColorR,\tableColorG,\tableColorB)},
        every head row/.style={%
            before row=\hline,%
            after row=\hline%
        },%
        columns/Schedule/.style={reset styles,column name={$\lambda$}, string type},
        every last column/.style={
                column type/.add={}{|},
                column type/.add={|}{}
        },
        every first column/.style={
                column type/.add={}{|},
                column type/.add={|}{}
            },
        every last row/.style={
            after row=\hline
        },
    ]{%
           Schedule,Caren,Mira,Magnus,Monty,Quill,Bado,Svea,Sora,ASP,CPMP,GRAPHS,TSP,All\\
			1,-0.0165,-0.3208,0.5431,0.5515,0.6118,0.7258,0.6248,0.2448,0.7649,0.6805,0.4689,-0.1777,0.3918\\
			2,-0.2234,-0.1891,0.4211,0.5508,0.6686,0.7322,0.6396,0.2789,0.7413,0.7218,0.5419,-0.0113,0.406\\
			3,0.0845,-0.1891,0.4458,0.5846,0.7139,0.759,0.6643,0.3428,0.7454,0.7885,0.5614,-0.0343,0.4556\\
			4,0.0812,-0.1891,0.4458,0.5451,0.7087,0.7434,0.6601,0.3263,0.7454,0.7698,0.5614,-0.0343,0.447\\
			5,0.0812,-0.1891,0.4458,0.5451,0.7059,0.7434,0.6584,0.325,0.7454,,0.5614,,0.4465\\
			6,0.0812,,0.4458,0.5451,0.7059,0.7434,0.6584,0.325,0.7454,,0.5614,,0.4465\\
}
}

\end{table}

This witnesses that, in order to understand how \tool performance is impacted by the 
$\lambda$ parameter, there is no need to consider high values for $\lambda$.
For this reason in Tab.~\ref{fig:training_schedule_size} we report  the 
closed gap score achieved when running \tool with its default values except $\lambda$,
which is varied from one to six. 
obtained by running SUNNY).\footnote{A white space in 
Tab.~\ref{fig:training_schedule_size} denotes the impossibility to run 
an experiment for a given $\lambda$ due to the limited amount of solvers 
available in the scenario.}  

By observing the average results for each $\lambda$ value, we can see that 
the overall best performance was reached
with $\lambda$ set to three. If $\lambda$ is less
than three, for most scenarios, the results are worse, and when $\lambda$ 
is greater than three the performances are the same, if not slightly worse.


\subsubsection{Variability}
\label{sec:variab}
One of the major concerns when dealing with predictions is the 
potentially huge impact of \emph{randomness} (e.g., how the folds are split) on the 
results.
To cope with the 
variability of our experiments we have adopted the repeated nested 
cross-validation 
approach that produces more robust results since the randomness of the 
inner cross-validation is weighted out in the outer cross-validation. 
Additionally, the 
repetition of the results 
takes into account the variability induced by 
creating randomly the folds for the outer cross-validation.

\begin{table}[ht]
\caption{Closed gap of training and test set using the best configuration 
found in each of the five folds.}

\label{fig:trainValidation}
    \center
    \resizebox{0.9\linewidth}{!}{%
    \pgfplotstabletypeset[
        precision=4,
        fixed,
        fixed zerofill, 
        col sep=comma,
        row sep=crcr,    
         color cells={min=0,max=1},    
        /pgfplots/colormap={whiteblue}{rgb255(0cm)=(255,255,255); rgb255(1cm)=(\tableColorR,\tableColorG,\tableColorB)},
        header=false,
        every head row/.style={%
            before row=\hline,%
            output empty row,
        },%
        every first column/.style={
                column type/.add={|}{}
        },
        every col no 2/.style={
                 column type/.add={|}{}
        },
        every col no 4/.style={
                 column type/.add={|}{}
        },
        every col no 6/.style={
                 column type/.add={|}{}
        },
        every last column/.style={
                column type/.add={}{|}
        },
        every last row/.style={%
            after row=\hline,
        },%
        every row no 4/.style={%
            after row=\hline,
        },%
        every row no 9/.style={%
            after row=\hline,
        },%
        every row no 0/.append style={
            before row={
                \multicolumn{2}{|c|}{Caren} & \multicolumn{2}{c|}{Mira} & \multicolumn{2}{c|}{Magnus} & \multicolumn{2}{c|}{Monty} \\\hline
                \multicolumn{1}{|c|}{train}  & \multicolumn{1}{c|}{test} & \multicolumn{1}{c|}{train}  & \multicolumn{1}{c|}{test} & \multicolumn{1}{c|}{train}  & \multicolumn{1}{c|}{test} & \multicolumn{1}{c|}{train}  & \multicolumn{1}{c|}{test}\\\hline
            }},
        every row no 5/.append style={
            before row={
                \multicolumn{2}{|c|}{Quill} & \multicolumn{2}{c|}{Bado} & \multicolumn{2}{c|}{Svea} & \multicolumn{2}{c|}{Sora} \\\hline
                \multicolumn{1}{|c|}{train}  & \multicolumn{1}{c|}{test} & \multicolumn{1}{c|}{train}  & \multicolumn{1}{c|}{test} & \multicolumn{1}{c|}{train}  & \multicolumn{1}{c|}{test} & \multicolumn{1}{c|}{train}  & \multicolumn{1}{c|}{test}\\\hline
            }},
        every row no 10/.append style={
            before row={
                \multicolumn{2}{|c|}{ASP} & \multicolumn{2}{c|}{CPMP} & \multicolumn{2}{c|}{GRAPHS} & \multicolumn{2}{c|}{TSP} \\\hline
                \multicolumn{1}{|c|}{train}  & \multicolumn{1}{c|}{test} & \multicolumn{1}{c|}{train}  & \multicolumn{1}{c|}{test} & \multicolumn{1}{c|}{train}  & \multicolumn{1}{c|}{test} & \multicolumn{1}{c|}{train}  & \multicolumn{1}{c|}{test}\\\hline
            }},
        every row 3 column 3/.style={reset styles, string type},
    ]{%
            0.5022,0.2561,0.9429,0.6692,0.8379,0.485,0.8658,0.2182\\
            0.6705,-9.3722,0.9709,0.008,0.827,0.9711,0.7192,0.6078\\
            0.5934,0.931,0.9858,0.4784,0.9723,0.0531,0.7276,0.7111\\
            0.6724,0.8784,0.9754,-30.2757,0.8203,0.7517,0.8031,0.6629\\
            0.7531,-0.0558,0.9878,-0.0556,0.8466,0.9522,0.7752,0.9945\\          
            0.8341,0.7623,0.8852,0.7944,0.6956,0.7455,0.6139,0.2746\\
            0.7644,0.715,0.8865,0.621,0.7495,0.8183,0.5565,0.056\\
            0.8289,0.7596,0.9162,0.8441,0.7591,0.6683,0.5445,0.3103\\
            0.7856,0.595,0.7935,0.7951,0.767,0.6969,0.536,0.3221\\
            0.8963,0.7902,0.8896,0.6961,0.7276,0.627,0.6666,0.089\\
            0.8391,0.6777,0.8079,0.7776,0.8556,0.2083,0.74,-0.0187\\
            0.8708,0.7385,0.845,0.5576,0.815,0.597,0.5863,0.0087\\
            0.8927,0.7366,0.8277,0.8553,0.878,0.7204,0.5933,0.4889\\
            0.8686,0.8114,0.8211,0.769,0.8713,0.9851,0.5854,0.5176\\
            0.9106,0.7165,0.8727,0.7628,0.7927,0.6097,0.6358,-1.8486\\
}
}

\end{table}

If we look at the performance on the single outer folds, we can notice 
that \tool's performance can have a significant variation.
For example, Tab.~\ref{fig:trainValidation}, compares the 
closed gap score of the training set and the 
test
set when running \tool on 
the first repetition of the 5 cross-validation with 
default parameters.

It is quite obvious that the closed gap is higher in the training instances because in 
these cases the instances used for the training are also used for the 
testing. 
It is more interesting to observe that the closed gap of the training 
set is 
sometimes not uniform (e.g., TSP scenario) and that there can be significant differences 
between the closed gap of the train and 
test
set for certain folds (e.g., Caren 
scenario). This could mean that a random fold splitting might have a big 
impact on the learning.

\section{Insights on SUNNY}
\label{sec:sunny-more}
In this section we study the SUNNY algorithm more in depth, by exploring its strengths and 
its weaknesses with the aim of finding meaningful patterns.
We also show the virtual performance that the
version of \tool described in this paper would have achieved in the 2017 OASC, 
and we provide a new empirical comparison including new scenarios and AS approaches.
Before that, let us recall from Sect.~\ref{sec:sunnyas} 
the informal assumptions on which the original SUNNY algorithm relied on, namely:
\begin{enumerate}[(i)]
\item a small portfolio is usually enough to achieve a good performance;
\item solvers either 
solve a problem quite quickly, or cannot solve it in reasonable time;
\item solvers perform similarly on similar instances;
\item a too heavy training phase is often an unnecessary burden.
\end{enumerate}

Points (i) and (ii) motivate the way SUNNY selects
and schedules its solvers, respectively. In fact, if few solvers are enough to solve a 
given set of problems then the solvers selection of SUNNY never falls in its worst-case---exponential w.r.t.~the size of the portfolio. If condition (ii) holds, then the time allocated to each selected solver can be small---provided, of course, that the right solvers are chosen.

Points (iii) and (iv) explain why the $k$-NN algorithm has been chosen for SUNNY.
If assumption (iii) holds, then the solvers' performance over the neighborhood of the new, unforeseen instance to be solved are a good estimation for the solvers' performance on that instance.
Assumption (iv) guided the choice of a \emph{lazy} approach, in fact the $k$-NN 
algorithm does not build explicitly a prediction model. 

As seen in Sect.~\ref{sec:sunny-opt},
\tool partly disagrees with point (iv): 
\toolf, and \toolfk variants of \tool actually mitigate the laziness of SUNNY by 
adding a training phase where $k$-configuration and/or feature selection are 
performed. The experiments of Sect.~\ref{sec:experiments} somehow confirmed 
hypothesis (i) (see, e.g., Sect.~\ref{sec:sched_size} and in particular Tab.~\ref{fig:original_inspection}) and rejected hypothesis (iv): a proper training 
phase, even if computationally expensive, may remarkably boost the performance 
of SUNNY.
Let us now try to empirically understand if conditions (ii) and (iii) are verified on the scenarios evaluated in Sect.~\ref{sec:experiments}. 

\begin{figure}[t]

	\centering
	\begin{subfigure}[b]{0.49\textwidth}
		\includegraphics[width=1\linewidth]{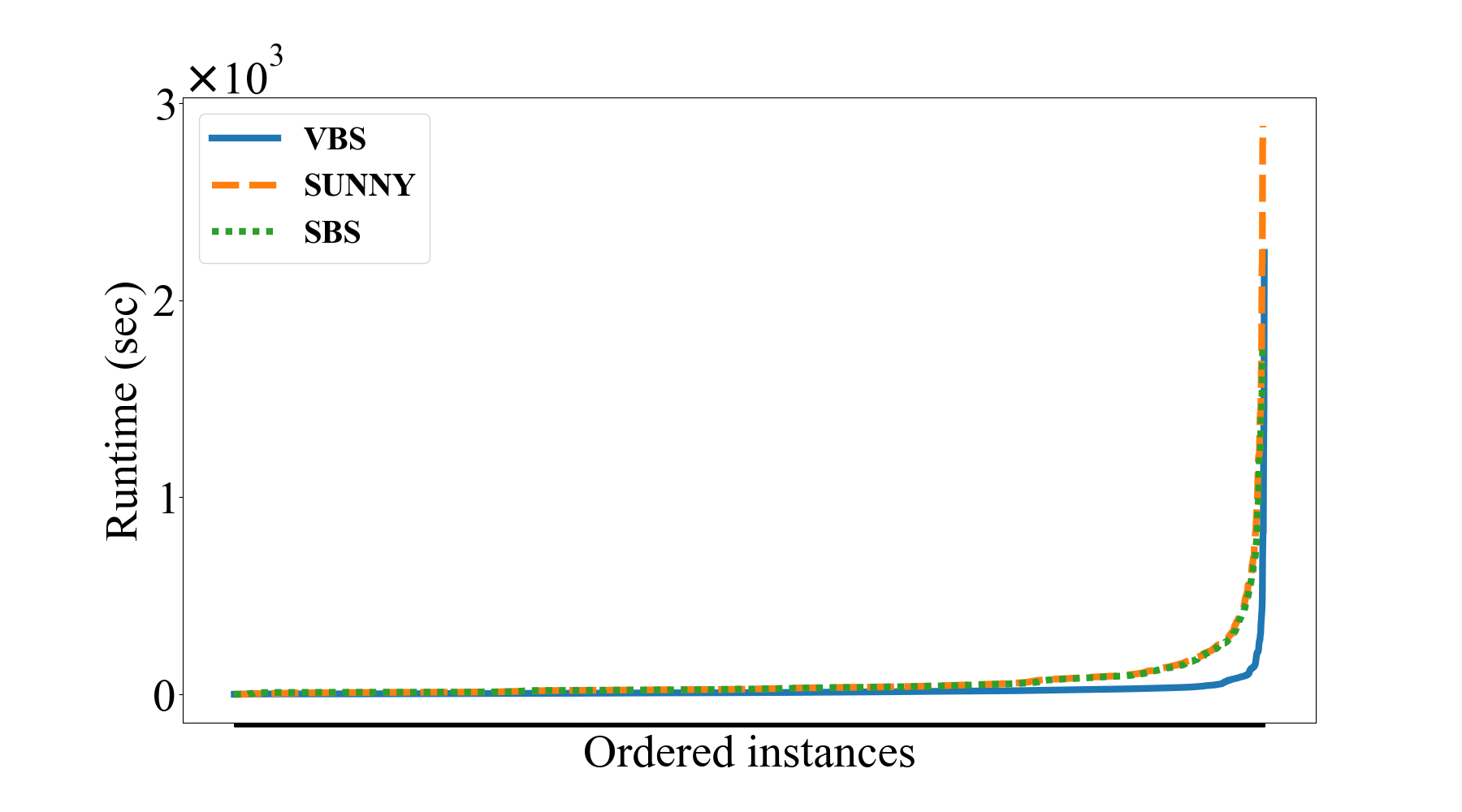}
		\caption{TSP.}
		\label{fig:inst_tsp}
	\end{subfigure}
	\begin{subfigure}[b]{0.49\textwidth}
		\includegraphics[width=1\linewidth]{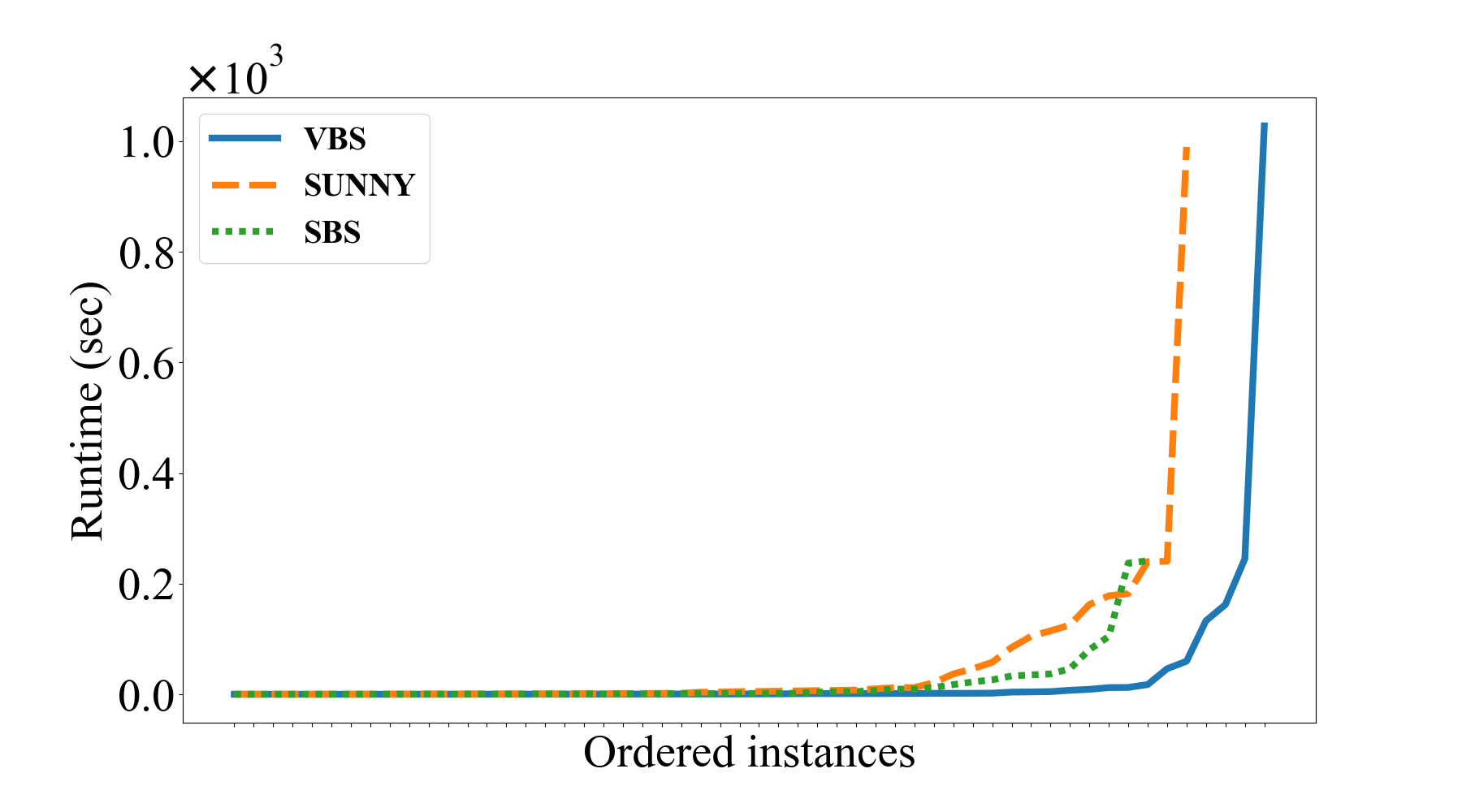}
		\caption{Caren.}
		\label{fig:inst_caren}
	\end{subfigure}

	\begin{subfigure}[b]{0.49\textwidth}
		\includegraphics[width=1\linewidth]{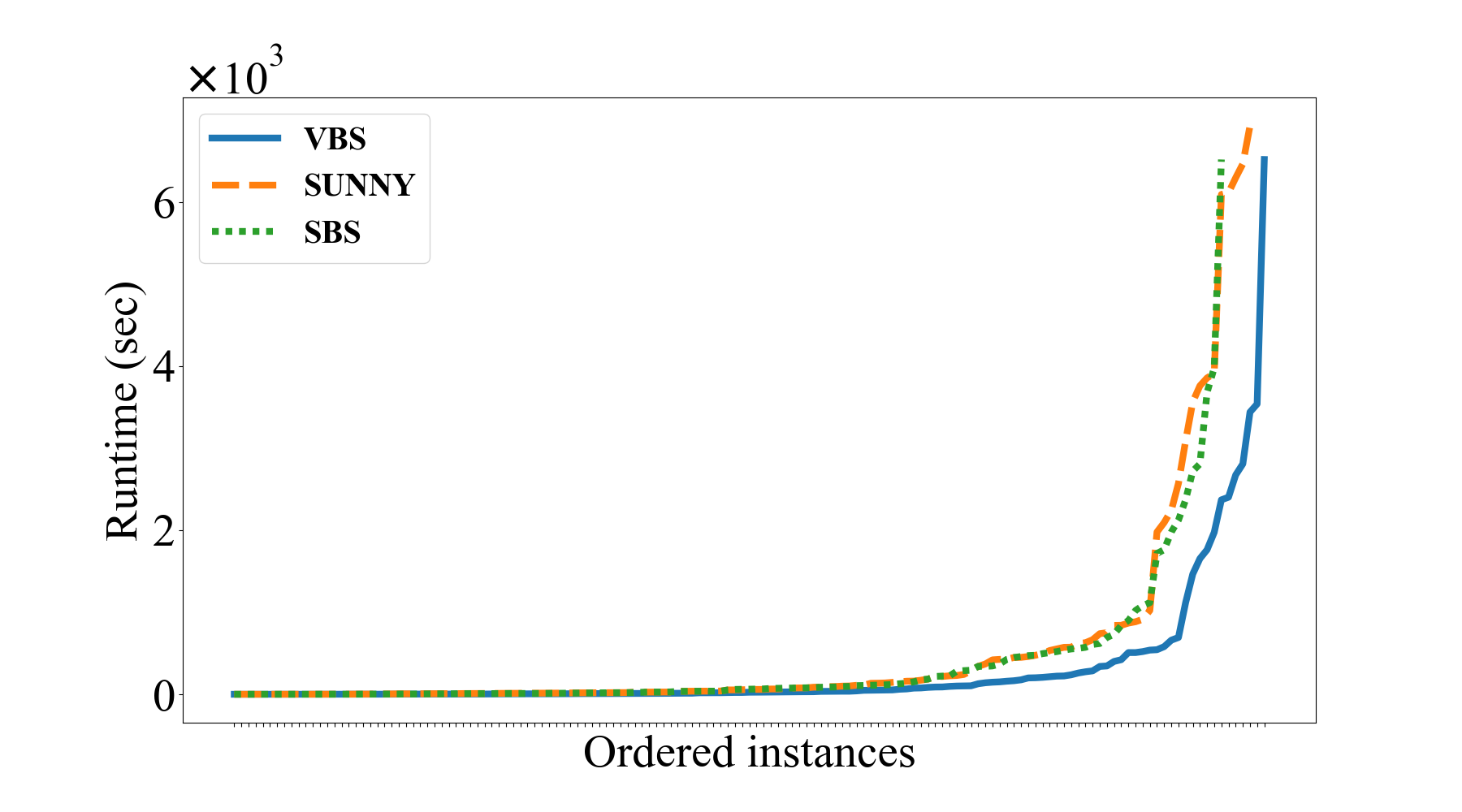}
		\caption{Mira.}
		\label{fig:inst_mira}
	\end{subfigure}
	\begin{subfigure}[b]{0.49\textwidth}
		\includegraphics[width=1\linewidth]{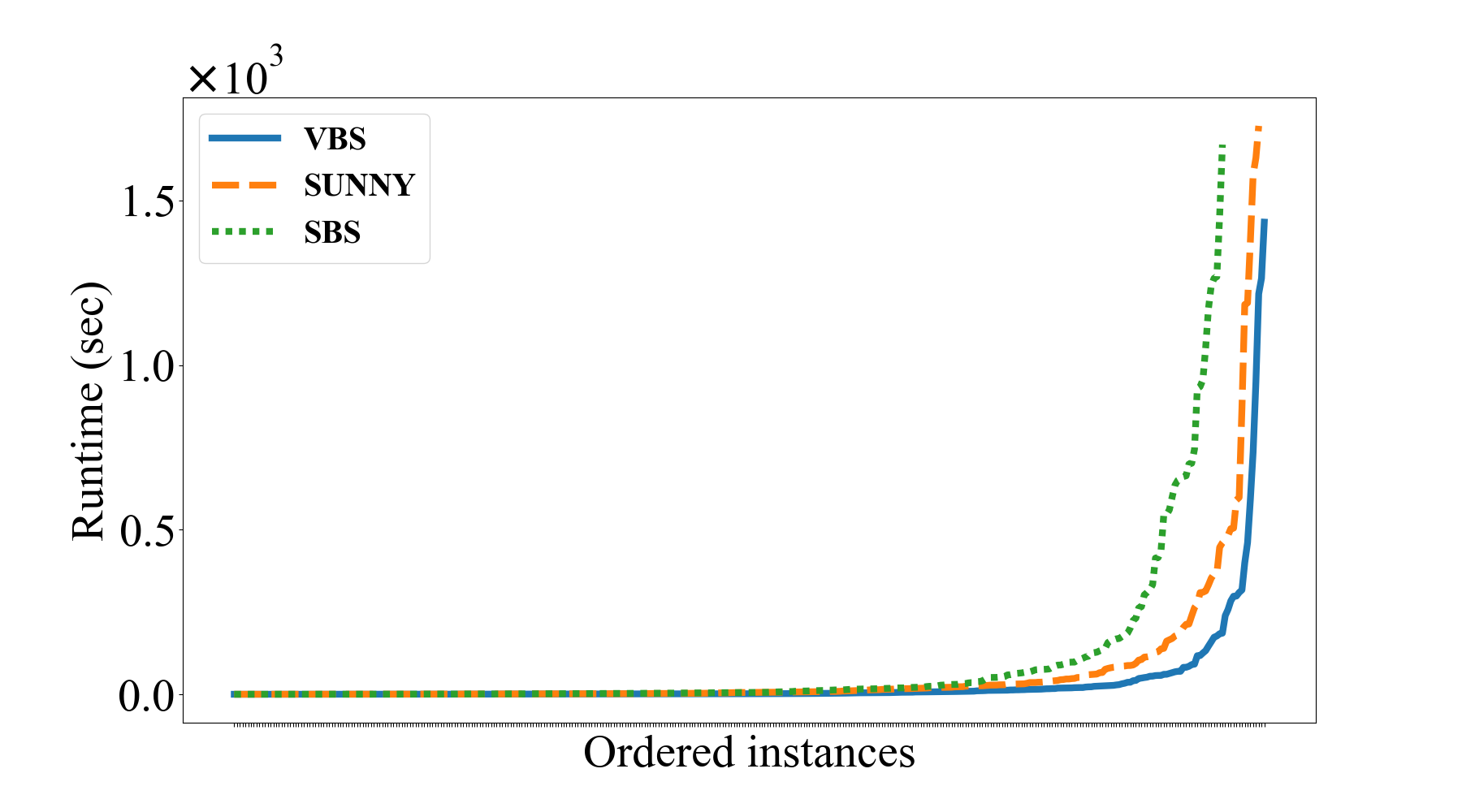}
		\caption{Magnus.}
		\label{fig:inst_magnus}
	\end{subfigure}

	\begin{subfigure}[b]{0.49\textwidth}
		\includegraphics[width=1\linewidth]{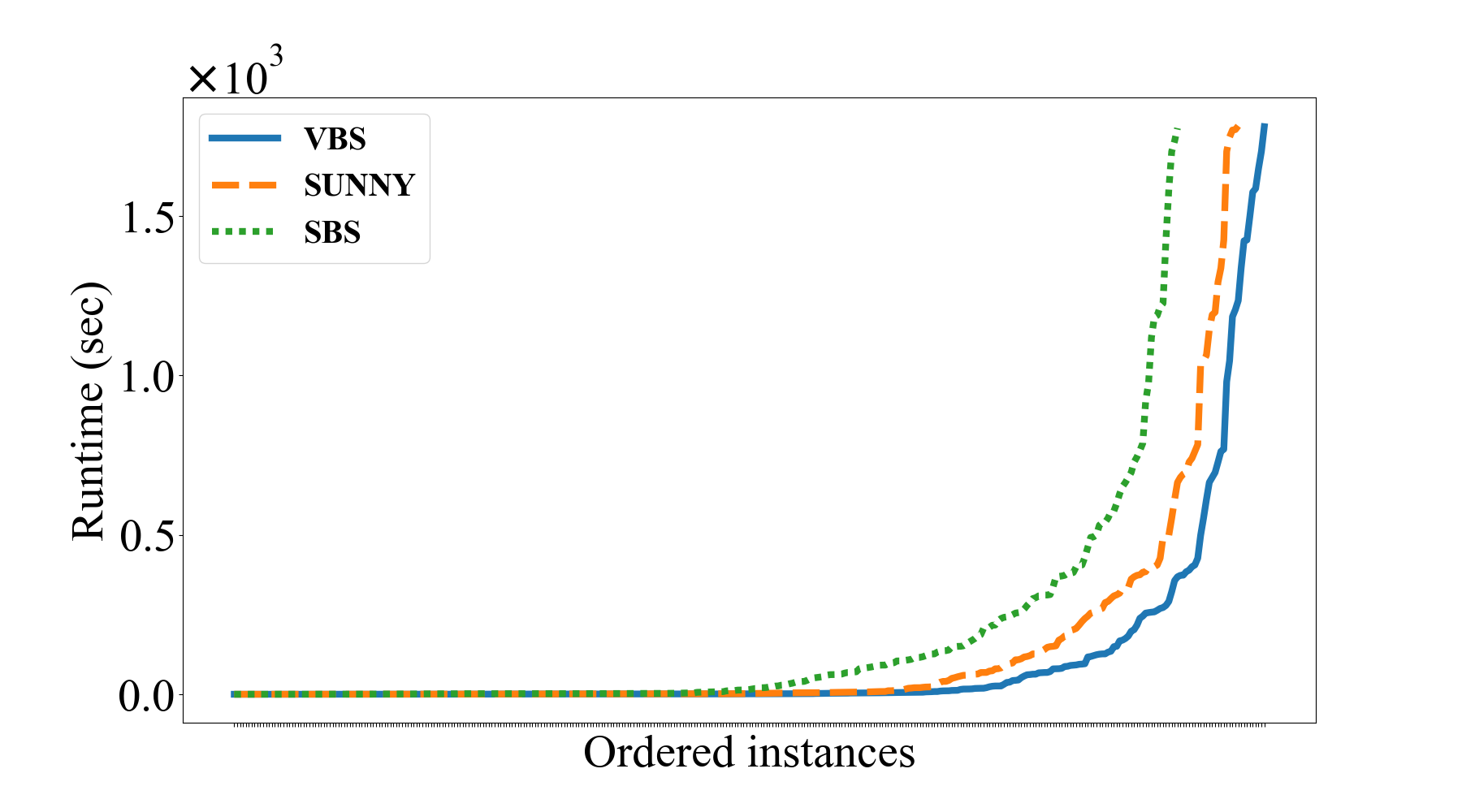}
		\caption{Monty.}
		\label{fig:inst_monty}
	\end{subfigure}
	\begin{subfigure}[b]{0.49\textwidth}
		\includegraphics[width=1\linewidth]{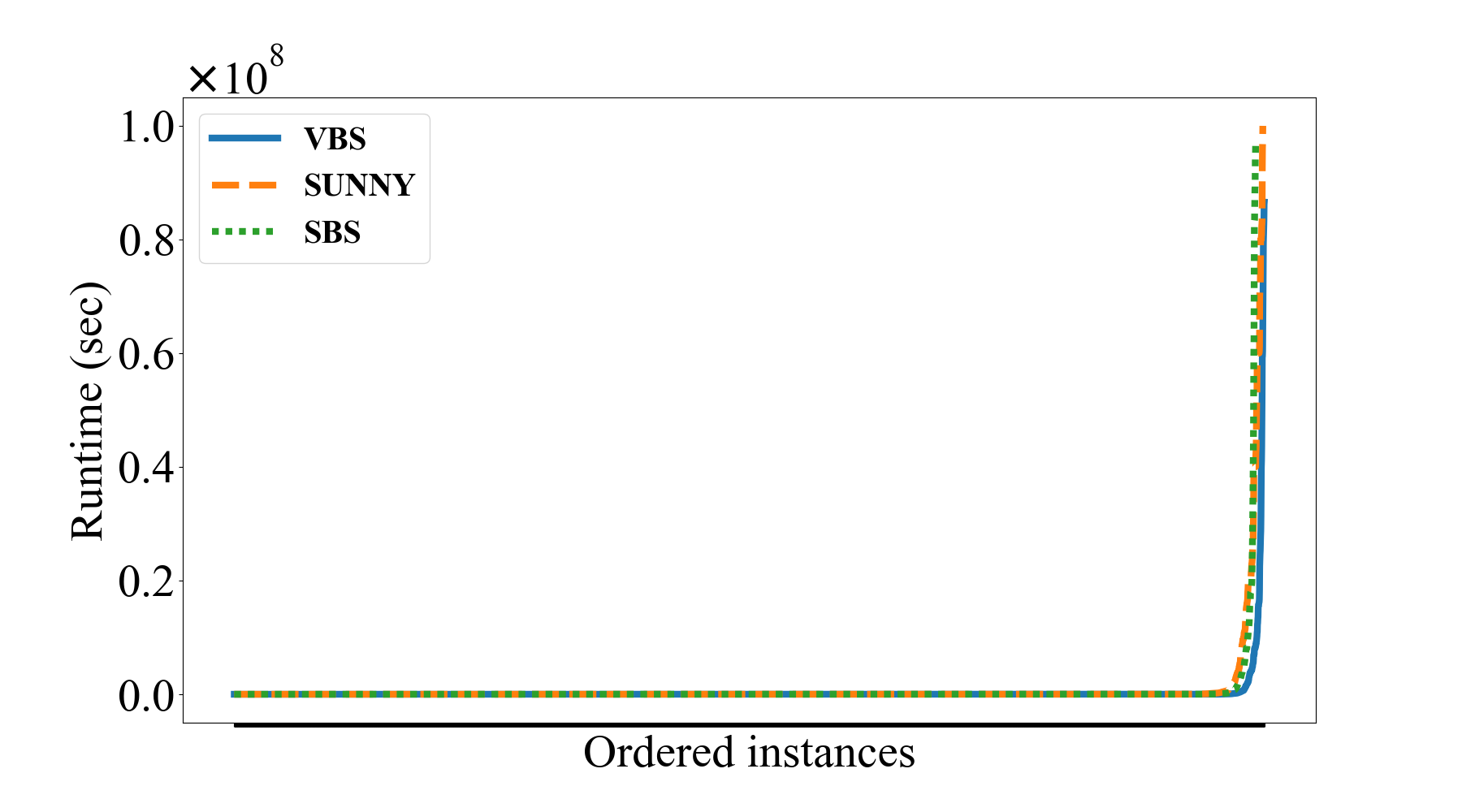}
		\caption{GRAPHS.}
		\label{fig:inst_graphs}
	\end{subfigure}
	\caption{Runtime distribution per scenario.}\label{fig:closedgap1}
\end{figure}

\begin{figure}[t]
	
	\centering
	\begin{subfigure}[b]{0.49\textwidth}
		\includegraphics[width=1\linewidth]{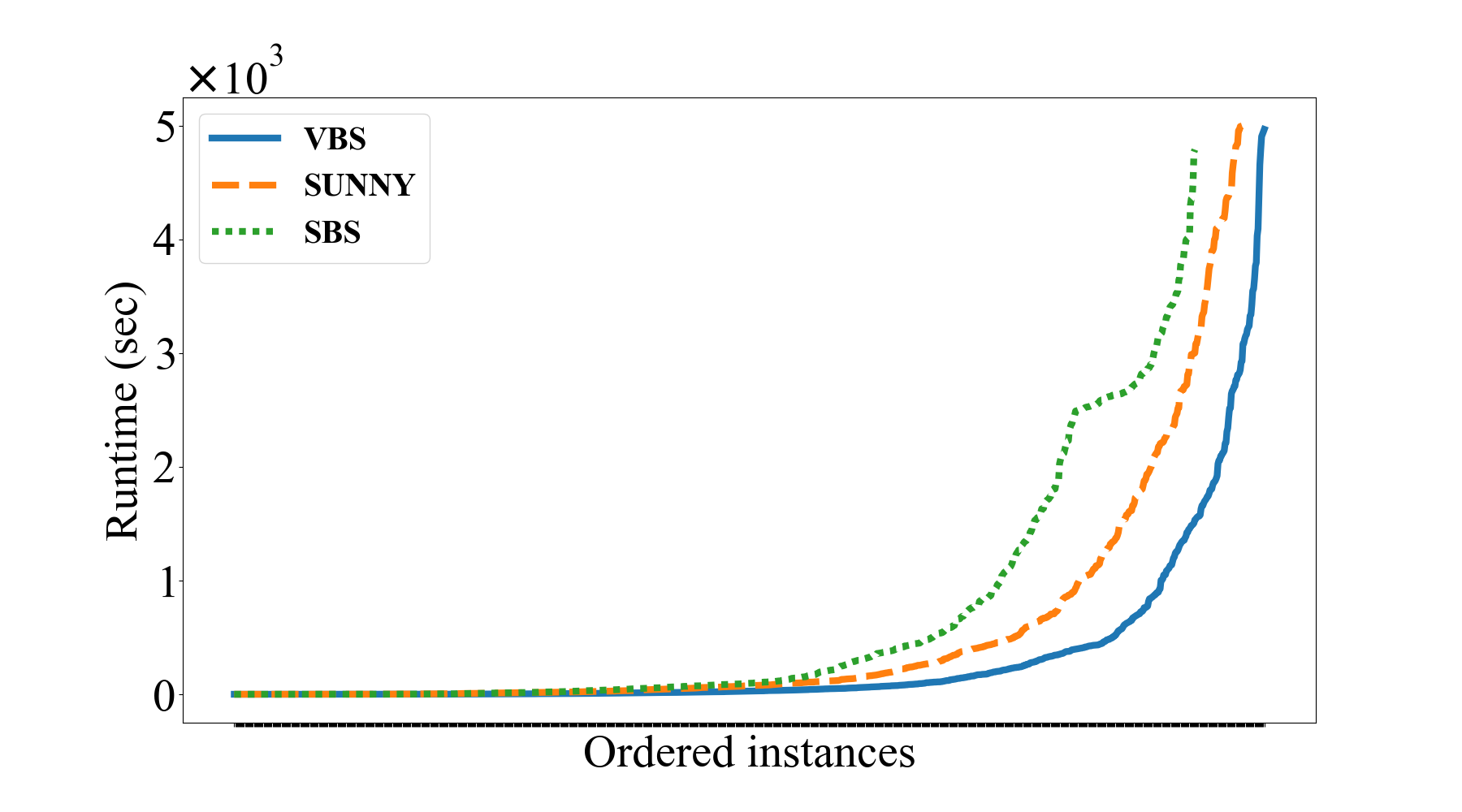}
		\caption{Sora.}
		\label{fig:inst_sora}
	\end{subfigure}
	\begin{subfigure}[b]{0.49\textwidth}
		\includegraphics[width=1\linewidth]{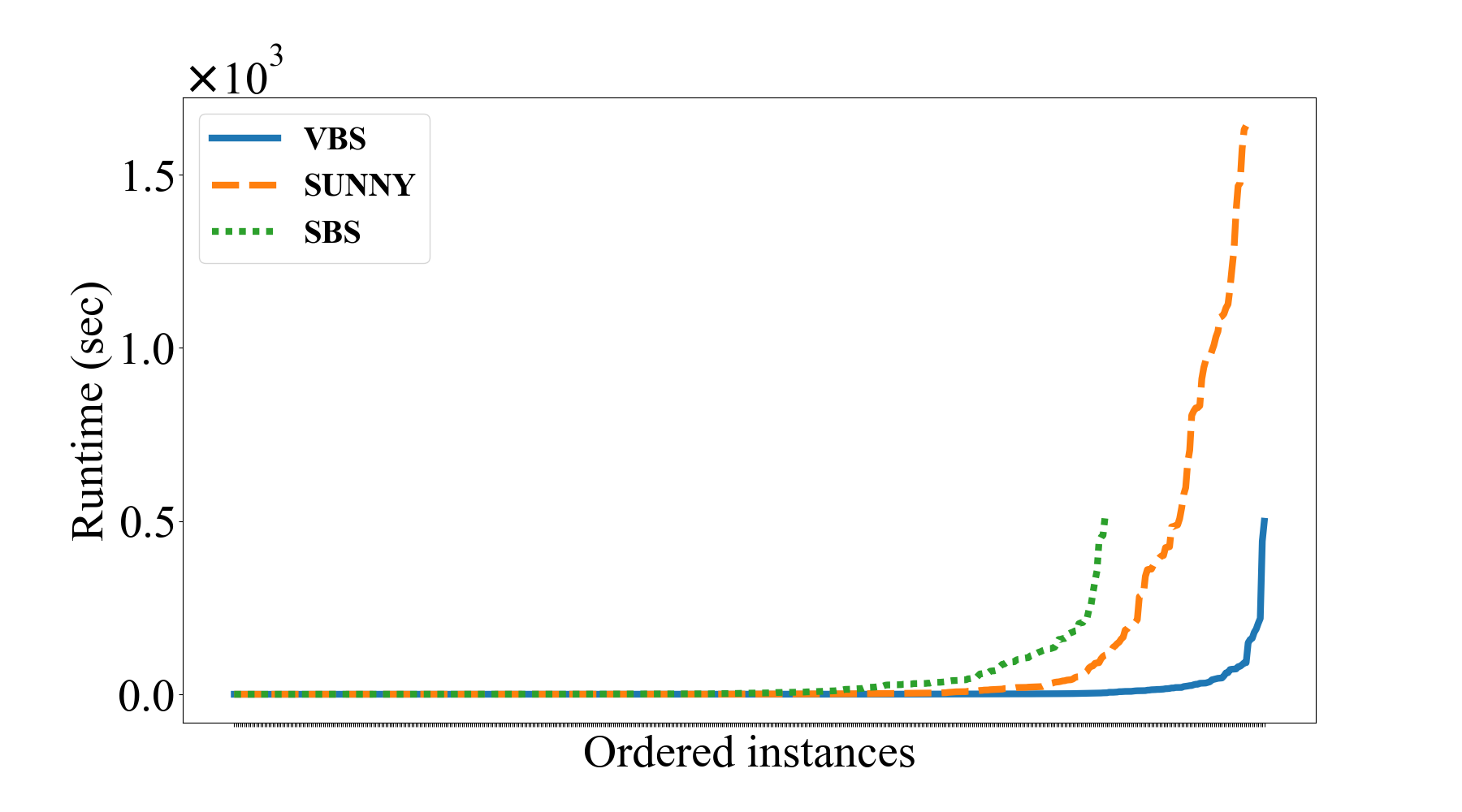}
		\caption{Quill.}
		\label{fig:inst_quill}
	\end{subfigure}

	\begin{subfigure}[b]{0.49\textwidth}
		\includegraphics[width=1\linewidth]{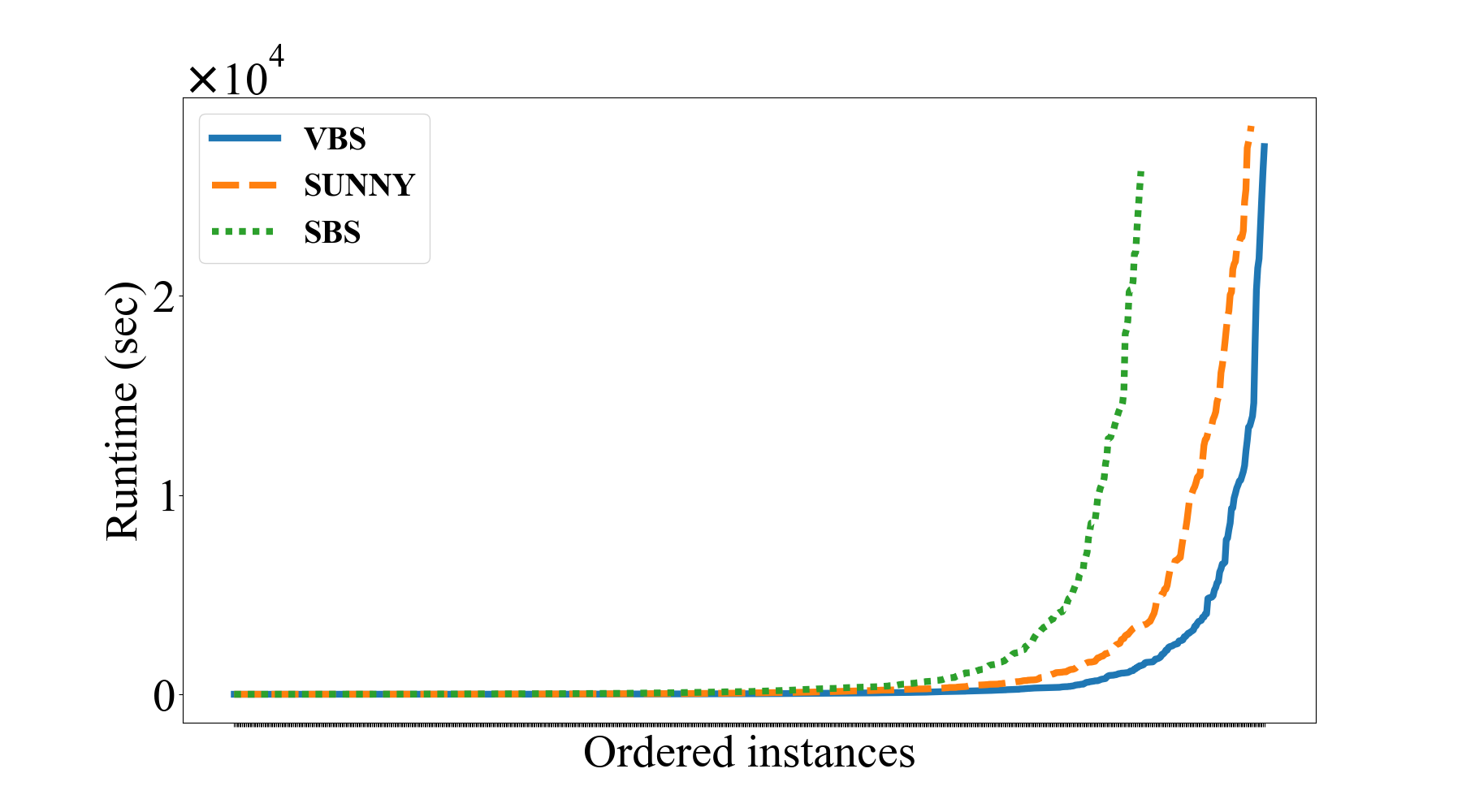}
		\caption{Bado.}
		\label{fig:inst_bado}
	\end{subfigure}
	\begin{subfigure}[b]{0.49\textwidth}
		\includegraphics[width=1\linewidth]{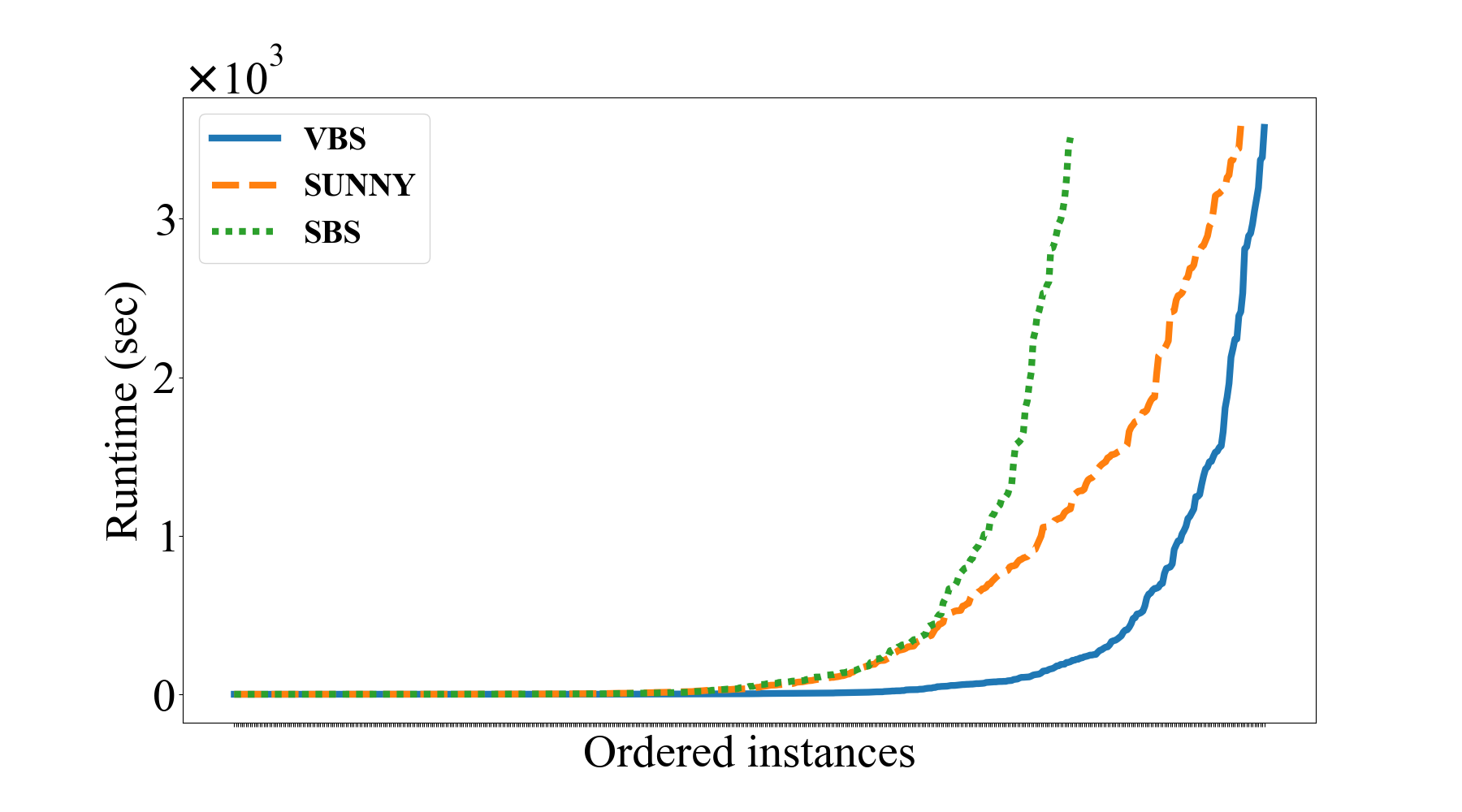}
		\caption{CPMP.}
		\label{fig:inst_cpmp}
	\end{subfigure}

	\begin{subfigure}[b]{0.49\textwidth}
		\includegraphics[width=1\linewidth]{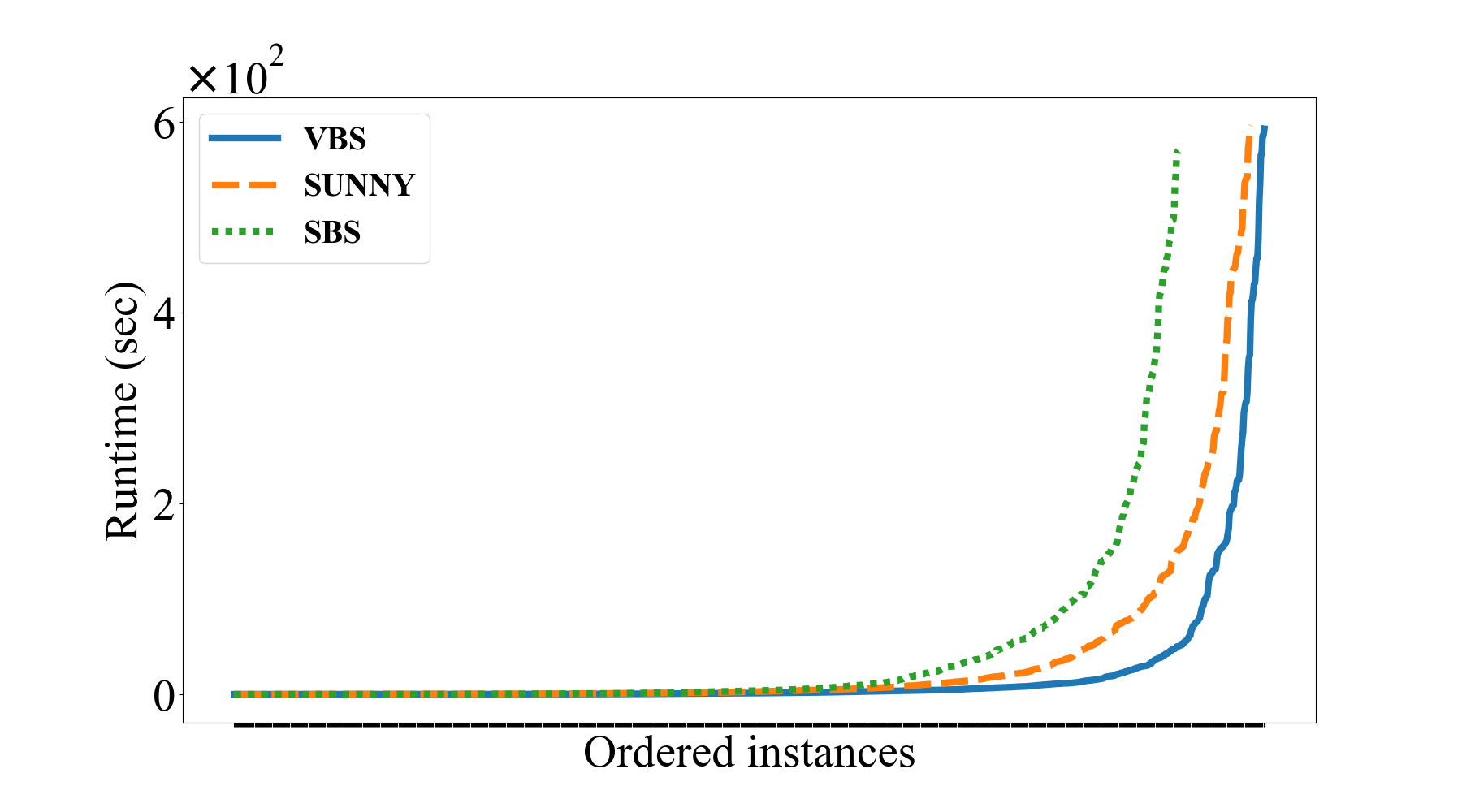}
		\caption{ASP.}
		\label{fig:inst_asp}
	\end{subfigure}
	\begin{subfigure}[b]{0.49\textwidth}
		\includegraphics[width=1\linewidth]{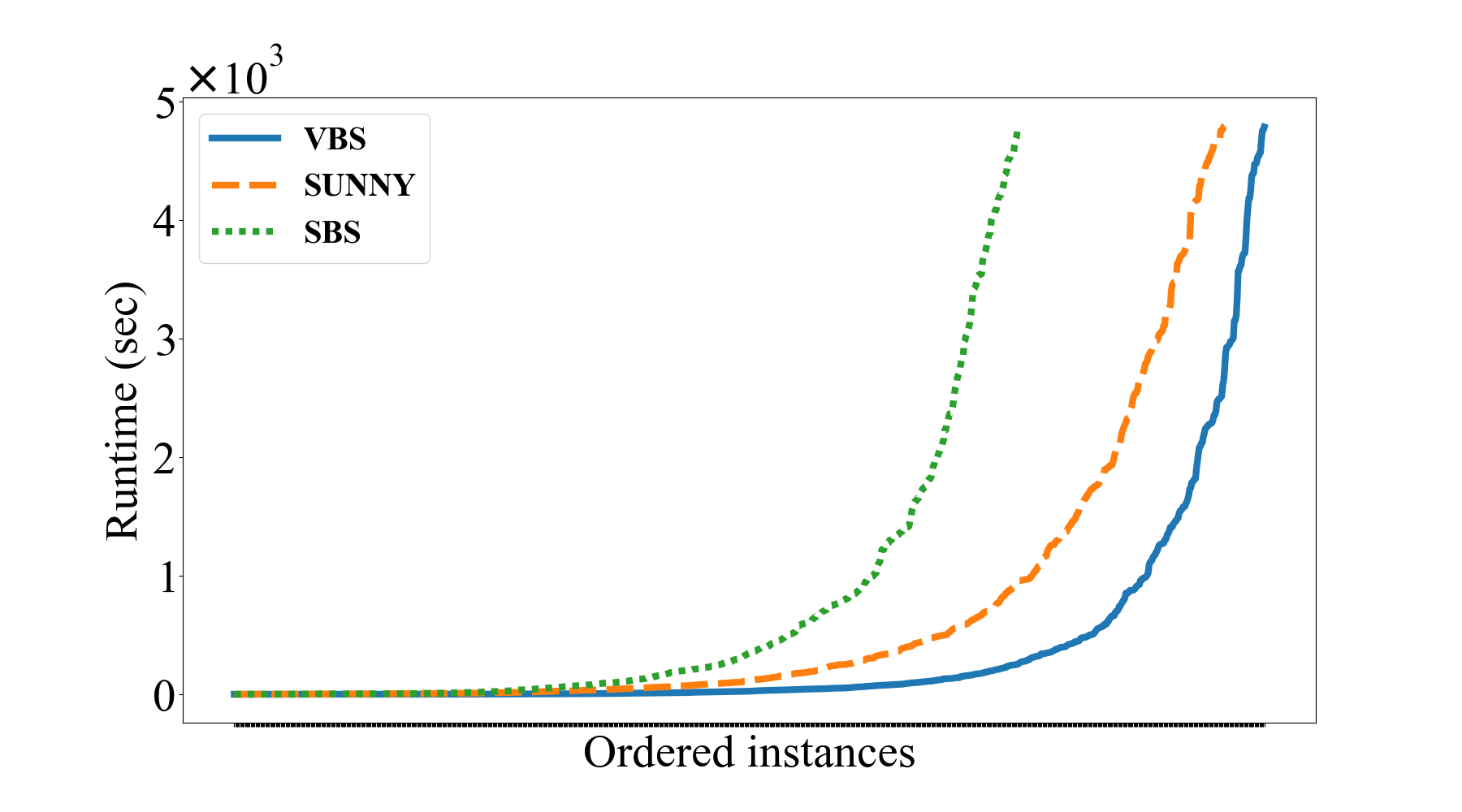}
		\caption{Svea.}
		\label{fig:inst_svea}
	\end{subfigure}
	\caption{Runtime distribution per scenario.}\label{fig:closedgap2}
\end{figure}
Fig.~\ref{fig:closedgap1} and \ref{fig:closedgap2} 
plot the runtime of the SBS (green), the VBS (blue), and \tool (yellow) 
on every instance of each scenario.
The instances are sorted in ascending order by the runtime of the corresponding algorithm selector.
The runtime distributions depicted in the plots provide evidence for hypothesis (ii): the runtime curves are essentially flat until they grow very quickly towards the end.

Hypothesis (iii) informally states that solvers perform similarly on similar 
instances, assuming that the feature vectors are able to describe the nature of 
the instances.
To get an idea of the similarity between instances and solvers' performances we decided to use the \emph{Jaccard index} which, given sets $A$ and $B$, is computed as
$J(A,B)=\dfrac{|A \cap B|}{|A \cup B|}$. This index is a value between 0 (when $A \cap B = \emptyset$) and 1 (when $A=B$) that gives a measure of the \emph{similarity} of sets 
$A$ and $B$: the higher the index, the more similar the sets are.
For each instance $i\in \I$ of a given scenario $(\I, \A, m)$
we compute $J(F_i, P_i)$ where $F_i \subseteq \I$ is 
the ``regular'' $k$-neighborhood computed by \tool according to the feature vectors, 
and $P_i \subseteq \I$ is the ``oracle-like'' $k$-neighborhood computed by 
\tool according to the \emph{performance vectors} defined as 
 $\langle m(i, A_1), \dots, m(i, A_n) \rangle$ where $\A = \{A_1, \dots, A_n\}$.

\begin{figure}[t]
	\centering
		\includegraphics[width=0.8\linewidth]{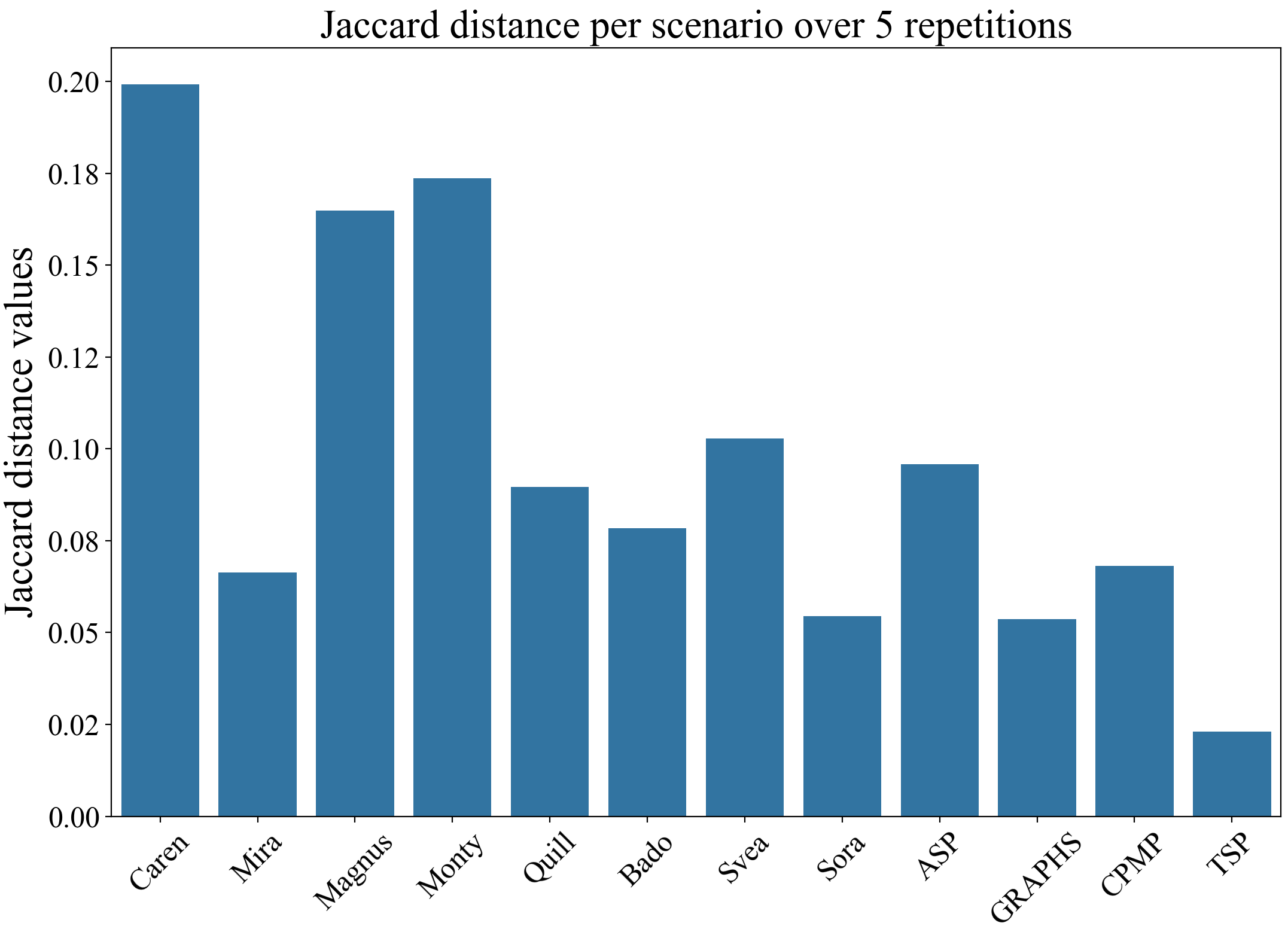}
	\caption{Jaccard index per scenario.\label{fig:jac1}}
\end{figure}

Fig.~\ref{fig:jac1} shows the average Jaccard index computed over all the training instances for each repetition using the runtime as metric for the performance vectors.
As we can see, the average index is usually pretty low: 
it is below 0.1 for the majority of scenarios
and the maximum index is below 0.2. The average value considering all the scenarios is  0.0636, i.e., on average, for every ten instances of $P_i \cup F_i$ less than one belongs to the intersection $P_i \cap F_i$.
The low Jaccard index raises major doubts on the assumption that solvers perform 
similarly on similar instances. We shall talk more in depth about this aspect in 
the next section.


\subsection{Hard Scenarios for SUNNY}
\label{sec:hard}
Let us now closely investigate the scenarios where \tool struggled, trying to 
extract meaningful patterns. 

\begin{figure}[t]

	\begin{subfigure}[b]{1\linewidth}\centering
		\includegraphics[width=0.99\linewidth]{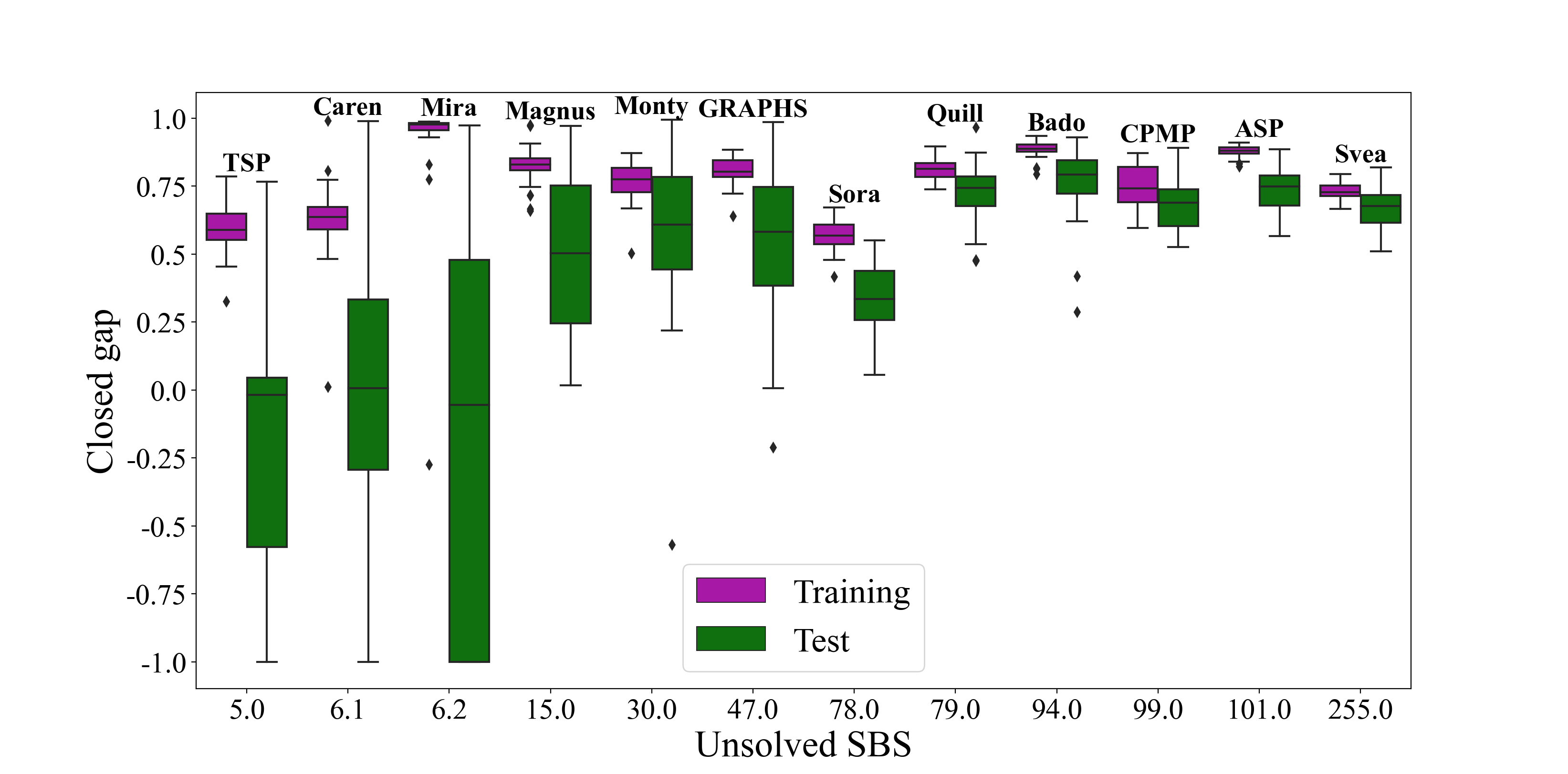}
		\caption{Scenarios ordered by the number of SBS unsolved instances.}
		\label{fig:fig_a}
	\end{subfigure}
	\begin{subfigure}[b]{1\linewidth}\centering
		\includegraphics[width=0.99\linewidth]{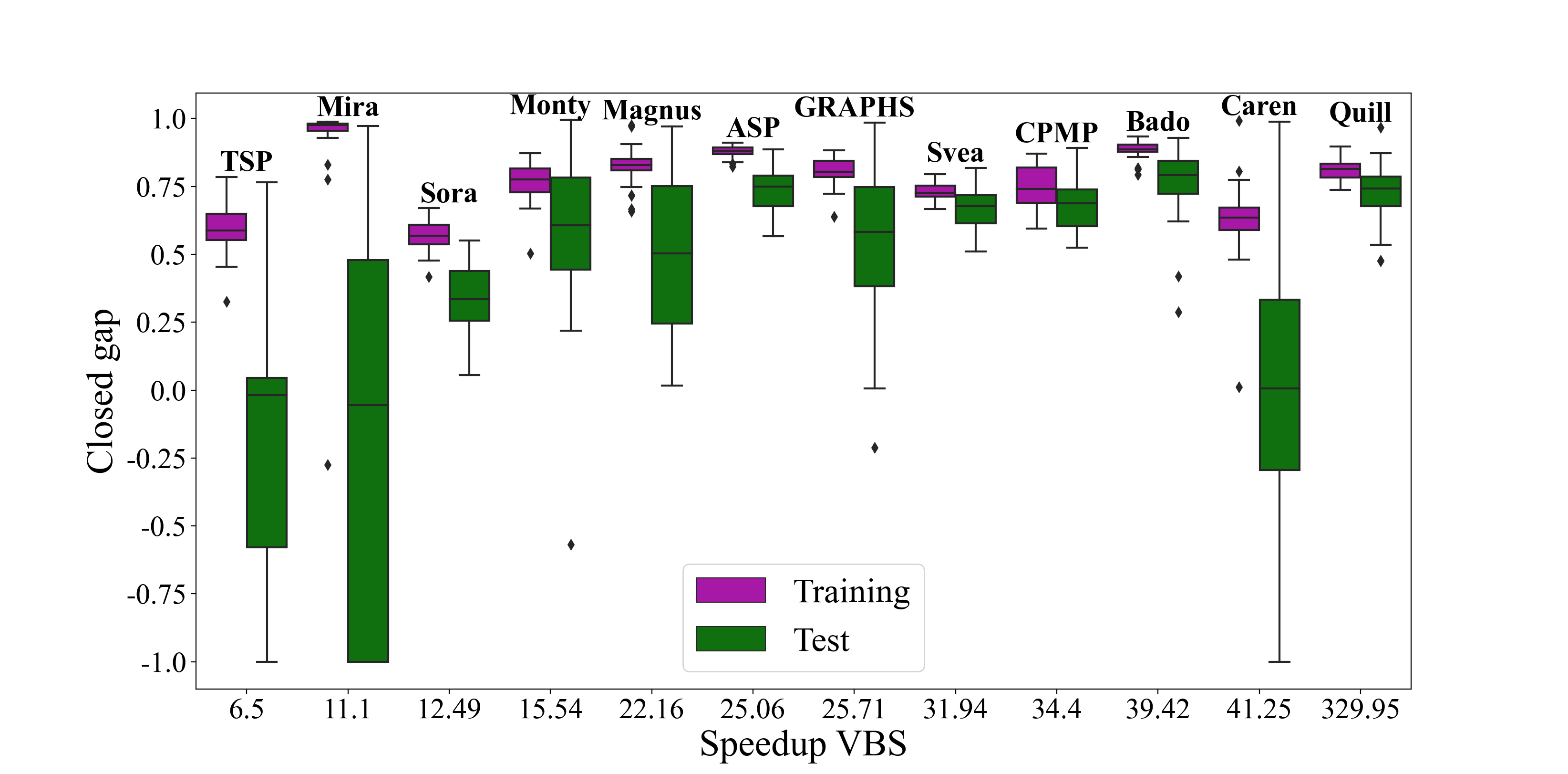}
		\caption{Scenarios ordered by the VBS speedup w.r.t. SBS.}
		\label{fig:fig_b}
	\end{subfigure}
	\caption{Closed gap score distribution in training and test folds.}\label{fig:repeated}
	\centering
\end{figure}

We start by extending the study performed in Sect.~\ref{sec:variab} 
by considering the closed gap distribution over 
the 25 training/test folds of all the repetitions (Tab.~\ref{fig:trainValidation} of Sect.~\ref{sec:experiments} refers to the first repetition only).
Fig.~\ref{fig:repeated} shows the closed gap performance with boxplots.

We found two indicators that seem to well represent the link 
between \tool's performance and the AS scenarios, i.e.,
the number of instances \textit{unsolved} by the SBS
and the \textit{speedup} of the VBS w.r.t. the SBS.
Both metrics somehow measure the distance between SBS and VBS: 
the former only focuses on the problems solved, while the latter also takes  
runtime into account.
Fig.~\ref{fig:fig_a} and \ref{fig:fig_b} show 
the closed gap score distributions for each scenario,
sorted respectively by increasing number of SBS unsolved instances
and by speedup of the VBS w.r.t.~SBS.
For representation purposes, the few closed gap scores having a value below $-1$ 
were replaced with $-1$.

From Fig.~\ref{fig:fig_a} and Fig.~\ref{fig:fig_b} one can clearly see 
that \tool tends to have a more strong and stable performance
in scenarios with higher values of the two indicators (e.g., Bado and CPMP).
Conversely, its performance is poor for scenarios with lower values 
for these indicators (e.g., TSP and Mira).

The plots in Fig.~\ref{fig:fig_a} and Fig.~\ref{fig:fig_b} 
are similar. 
However, the position of Caren scenario in 
Fig.~\ref{fig:fig_b} may suggest that the number of SBS unsolved 
instances is a more reliable indicator 
to analyze the performance of \tool in terms of closed gap.


Overall, SUNNY seems to not work well when the SBS has little room for improvements. 
We argue that the difficulties of SUNNY in scenarios with a low value of the above indicators are quite normal.
Conversely, an algorithm selector performing too well in those scenarios might denote an overfitting w.r.t.~the few instances for which the SBS is not a good choice.
Moreover, there can be other (co-)explanations for the bad performance of SUNNY on TSP, Mira and Caren. In fact, TSP is the scenario with the lowest number of algorithms (only 4) and the performance of the SBS almost overlaps with that of the VBS (see Fig.~\ref{fig:inst_tsp}).
Caren and Mira are instead the scenarios with the fewest number of instances: 
only 66 and 145 respectively.


We further investigated the cases where \tool did not work well by 
focusing on the instances that it could not solve.
We distinguish them in two categories:
\textit{(i)} those unsolved because wrong solvers were scheduled, i.e., no solver in the schedule could actually solve that instance within the timeout; and 
\textit{(ii)} those unsolved because not enough time was allocated, i.e., at least one of the scheduled solvers could actually solve that instance with a time slot larger than the allocated one.

\begin{figure}[t]
	\centering
	
	\includegraphics[width=1\linewidth]{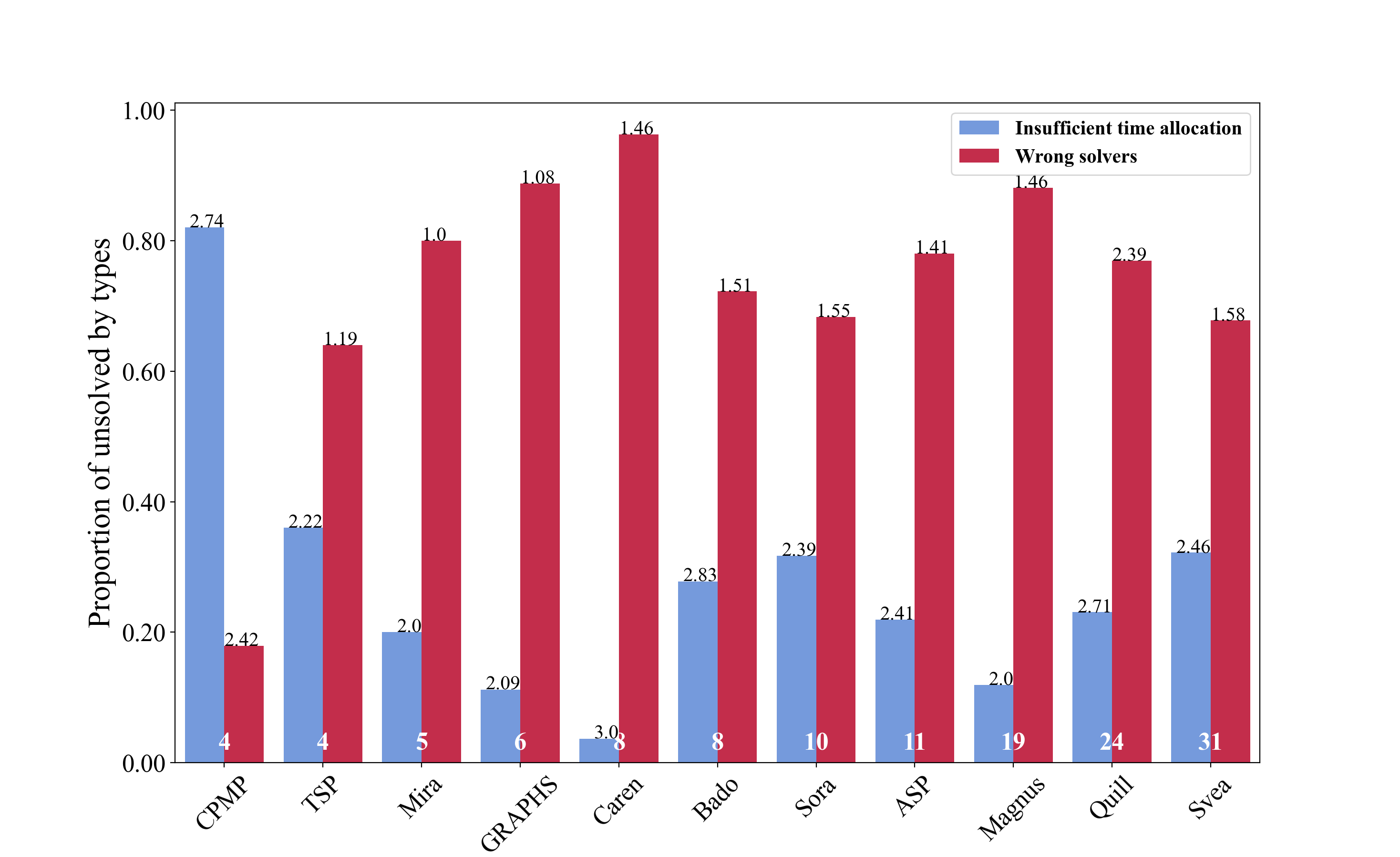}
	
	\caption{Unsolved instances by \tool. White numbers on the bottom of the histograms refer to the portfolio size, while the numbers on the top of the histograms refer to the average schedule size.
	\label{fig:unsolved_types}}

\end{figure}
Fig.~\ref{fig:unsolved_types} shows the instances unsolved by \tool for each scenario, 
grouped by the above categories. The plot also shows the portfolio size of each scenario.
It is quite interesting to see that in all the scenarios except CPMP around  
$70\%$
of \tool's failures are due to a wrong identification of the solvers 
from the neighborhood instances. This means that probably the Achilles' heel of SUNNY is not 
the way the solvers are scheduled, but rather the way they are \textit{predicted}. 
The underlying $k$-NN algorithm might not be the best choice because the assumption that 
similar instances have similar behavior does not always hold.

Despite the good 
closed gap score reached by \tool on the CPMP scenario, this is the only 
scenario where the number of unsolved instances due to the time allocation 
is greater than the number 
of unsolved instances due to wrongly scheduled solvers.
We conjecture that this behavior is motivated by two \textit{co-factors}:
CPMP has the lowest number of
available solvers and the SBS performance is quite far from the VBS performance 
(see Fig.~\ref{fig:inst_cpmp}). For these reasons, SUNNY tends to allocate less 
time to the SBS and more time to the other solvers w.r.t.~to scenarios where the 
speedup of the VBS is low, even when few solvers are available (e.g., the TSP 
scenario). In fact, we note from Tab.~\ref{fig:original_inspection} that the 
average number of scheduled solvers for the CPMP scenario is 2.4, i.e., the 60\% 
of the overall portfolio size. 

Summarizing, according to the experiments we conducted in this work, we can say that the hypothesis (i) and (ii), stating that a small portfolio is usually enough to achieve a good performance and that 
solvers either solve a problem quite quickly, or cannot solve it in reasonable time are mostly true.
Conversely, hypothesis (iii) ``solvers perform similarly on similar instances'' and (iv) ``a too heavy training phase is often an unnecessary burden'' are not empirically confirmed.


\subsection{Comparison with Other Approaches}
\label{sec:comparisons}

In this section we provide a comparison between \tool and other state-of-the-art 
AS approaches.
First of all, we show what would have been the performance of the improved \tool 
in the 2017 OASC.
In fact, the version of \tool submitted to OASC was a preliminary version that, 
among 
other things, used a 10-fold 
cross-validation without implementing the rank split method to create 
the folds.\footnote{Appendix \ref{appendic:OASC_tool} describes in detail the 
technical 
differences between the current version of \tool and the one submitted to OASC.}

\begin{table}[t]
	\caption{Closed gap score results with OASC test instances.}
	\label{tab:oasc}
	\center
	\resizebox{0.9\linewidth}{!}{%
		\pgfplotstabletypeset[
		precision=4,
		fixed,
		fixed zerofill, 
		col sep=comma,
		row sep=crcr,    
		every first column/.style={
			column type={|l|}
		},
		color cells={min=0,max=1},    
		/pgfplots/colormap={whiteblue}{rgb255(0cm)=(255,255,255); rgb255(1cm)=(\tableColorR,\tableColorG,\tableColorB)},
		every head row/.style={%
			before row=\hline,%
			after row=\hline%
		},%
		columns/Approach/.style={reset styles, string type},
		every last column/.style={
			column type/.add={|}{|}
		},
		every last row/.style={
			after row=\hline
		},
		every row 1 column 8/.style={
			postproc cell content/.style={
				string type
			}
		},
		]{%
			Approach,Caren,Mira,Magnus,Monty,Quill,Bado,Svea,Sora,All\\
			\tool,0.7855,0.0291,0.5833,0.8450,0.8414,0.9057,0.6077,0.4059,0.6255 \\
			\toolfk-\textsf{OASC},0.9099,0.4320,0.5723,0.9102,0.5691,0.8444,0.6578,0.0084,0.6130 \\
			ASAP.v2,0.3238,0.5053,0.4979,0.8331,0.6981,0.7573,0.6765,0.2150,0.5634 \\
			AutoFolio,0.5995,0.0846,0.6707,0.6923,0.5165,0.8089,0.6585,0.3479,0.5474 \\
			{}*Zilla,0.6356,0.4761,0.4932,0.4194,0.8001,0.7322,0.5850,0.1754,0.5396 \\
			ASAP.v3,0.3276,0.5091,0.4963,0.7631,0.5797,0.8048,0.6881,0.0639,0.5291 \\
			\toolk-\textsf{OASC},0.6440,-0.0137,0.4924,0.6318,0.8495,0.7441,0.5789,0.0021,0.4911 \\
			SUNNY-original,0.7687,-0.8996,0.5859,0.4025,0.7697,0.7687,0.4866,0.1899,0.3841 \\
			Random Forest,0.1952,0.4892,0.2037,-1.4422,-0.4737,0.7913,0.5966,0.0934,0.0567\\
			AS-RF,-1.0617,0.4947,-1.0521,-6.8992,-0.3280,0.8331,0.5853,-0.3700,-0.9747\\
		}
	}
	
\end{table}

Tab.~\ref{tab:oasc} presents the virtual performance of \tool in the 2017 OASC.
In addition to the original competitors (viz., {}*Zilla, ASAP, AS-RF and the preliminary versions of \tool called \toolfk-\textsf{OASC} and \toolk-\textsf{OASC} in Tab.~\ref{tab:oasc}) we added three more baselines:
AutoFolio~\cite{lindauer2015autofolio},\footnote{The version of AutoFolio we 
	used is AutoFolio 2015 which attended the ICON challenge. Unfortunately, we 
	experienced some issues with the most recent version~\citeA{autofolioGithub} 
	due to the external libraries dependencies used by AutoFolio for parameter 
	tuning. Without parameter tuning, the recent version of AutoFolio has worse 
	results than the 2015 edition and therefore, for fairness reason, we reported 
	only the results of AutoFolio 2015.} the best performing approach 
of the 2015 ICON challenge in terms of $\parten$ score; 
the original SUNNY approach~\cite{sunny}, not performing any training; and an off-the-shelf random forest 
approach trained on the whole training set without additional cross validations. The latter was implemented with Scikit-learn~\cite{pedregosa2011scikit}
by labeling each instance with the fastest solver solving it (i.e., it maps 
the AS problem into a classification problem and uses random forest to tackle 
the classification problem). The number of estimators was set to 200, as done 
by ASAP.v2.

\tool was trained as explained in Sect.~\ref{sec:data_prep}.
For each scenario we picked the configuration that achieved the highest closed gap score among the 5 different 
configurations obtained on the training set (one for each fold of the outer 
cross-validation, we did not perform repetitions here).
For {}*Zilla \cite{cameron17a}, AS-RF \cite{malone2017asl}, and ASAP \cite{gonard17a} approaches, we only present the  
results they obtained in the OASC 2017 because no new 
version of these systems have been released since then.\footnote{The performance 
of these 
	approaches are available at \citeA{cosealhome}. 
	Note that the competition reported in \cite{lindauer2019algorithm} used 
	a different closed gap metric, i.e., $1 - \frac{m_{SBS} - m_s}{m_{SBS} - m_{VBS}}$,
	and the scoring tool was slightly amended.
	This work considers 
    the fixed version of {}*Zilla since the original one submitted to the competition had a critical bug~\cite{lindauer2019algorithm}.
	} 
Note that {}*Zilla, AutoFolio and AS-RF configure
their system hyper-parameters automatically thus they do not require manual tuning.
ASAP instead identified good performing parameters before the competition. 
We tried several other parameters for ASAP, without however outperforming the ones 
used in the challenge (cf. Tab.~\ref{tab:asap_exp} in 
Appendix~\ref{appendic:OASC_meta}).
In these experiments, AutoFolio and Random Forest 
are the only systems predicting a single solver rather than scheduling a number of solvers.



Tab.~\ref{tab:oasc} shows that \tool has the highest average closed gap, and it is the best approach 
in Bado and Sora scenarios. Its performance is quite close to the one of \toolfk-\textsf{OASC}:
the difference is greater than $0.2$ only in two scenarios, i.e., Caren and Mira.
This is not surprising 
since \tool is
quite similar to \toolfk-\textsf{OASC}.
ASAP.v2 does not attain the best score in any scenario, but in general its performance is 
robust and effective---this confirms what reported in~\cite{gonard2019algorithm}. 
AutoFolio is slightly behind ASAP.v2, nevertheless it achieves good results and it is the 
best approach for the Magnus scenario.
As \tool, also AutoFolio suffers in scenarios
like Caren and Mira having a small number of instances. 
{}*Zilla and ASAP.v3 also close more than 
50\% of the gap 
between the SBS and the VBS. \toolk-\textsf{OASC} is instead slightly below this 
threshold: the performance difference  w.r.t.~\toolfk-\textsf{OASC} denotes the 
importance of a proper feature selection.
The original \sunny approach is even worse: this confirms the effectiveness of 
the strategy introduced by \tool. At the bottom of the table we find the AS 
approaches based on random forest. This witnesses that turning an AS 
problem into a classification problem does not seem a good idea in general.


One thing to note is that the results of the OASC 
competition are based on a single training-test split. As discussed also 
by~\citeA{lindauer2019algorithm}, this \textit{``increases the 
risk of a particular submission with
randomized components getting lucky''}. For this reason, we also compared  
the performance of \tool and the other AS approaches
we could reproduce~\footnote{AS-RF has been 
excluded because it has 
library compatibility issues
and it is no longer maintained.} using the default 10 cross-validation splits of the
 ASlib. In addition to the ASlib scenarios considered so far, we included 
 all the runtime scenarios added to the \aslib after the OASC 
 challenge,  viz. GLUHACK-2018, SAT18-EXP and 
 MAXSAT19-UCMS (cf. Tab.~\ref{table:aslibcomp}).
 In the experiments in 
 	Sect. \ref{sec:experiments}, we did not consider these scenarios because 
 	we already had 2 SAT scenarios (i.e., Magnus and Monty) 
 	and 2 Max-Sat scenarios(i.e., Svea and Sora).

\begin{table}[t]
	\caption{Runtime \aslib scenarios added after 2017.}
	\label{table:aslibcomp}
	\centering
	\begin{tabular}{|l|l|l|l|l|}
		\hline
		Scenario   & Algorithms & Problems& Features 
		& Timeout\\ \hline
		GLUHACK-2018  & 8     & 237      & 50     & 5000 s\\ 
		SAT18-EXP           & 37      & 286       & 50      & 5000 s\\ 
		MAXSAT19-UCMS         & 7      &  440    & 54      & 3600  s\\ 
		
		\hline
	\end{tabular}
\end{table}
\begin{table}[t]
	\caption{Closed gap score results using \aslib default splits.}
	\label{tab:oasc_new}
	\center
	\resizebox{0.9\linewidth}{!}{%
		\pgfplotstabletypeset[
		precision=4,
		fixed,
		fixed zerofill, 
		col sep=comma,
		row sep=crcr,    
		every first column/.style={
			column type={|l}
		},
		every row 14 column 4/.style={reset styles, string type},
		every row 14 column 6/.style={reset styles, string type},
		color cells={min=0,max=1},    
		/pgfplots/colormap={whiteblue}{rgb255(0cm)=(255,255,255); rgb255(1cm)=(\tableColorR,\tableColorG,\tableColorB)},
		every head row/.style={%
			before row=\hline,%
			after row=\hline%
		},%
		columns/Scenario/.style={reset styles, string type},
		every row no 14/.style={after row=\hline},
		every last column/.style={
			column type/.add={}{|}
		},
		every last row/.style={
			after row=\hline
		},
		]{
		Scenario,ASAP.v2,sunny-as2,SUNNY-original,AutoFolio,{}*Zilla,Random Forest\\
ASP-POTASSCO,0.7444,0.8258,0.6300,0.8273,0.6786,0.5314\\
BNSL-2016,0.8463,0.8155,0.8372,0.8479,0.7616,0.7451\\
CPMP-2015,0.6323,0.8423,0.7509,0.3046,0.4773,0.1732\\
CSP-Minizinc-Time-2016,0.6251,0.4522,0.3427,0.5365,0.0866,0.2723\\
GLUHACK-2018,0.4663,0.4427,0.2663,0.4426,0.2402,0.4057\\
GRAPHS-2015,0.7580,0.6339,0.6917,0.6597,0.5934,-0.6412\\
MAXSAT-PMS-2016,0.5734,0.5311,0.4951,0.5309,0.5182,0.3263\\
MAXSAT-WPMS-2016,0.7736,0.8144,0.5044,0.6333,0.5909,-1.1826\\
MAXSAT19-UCMS,0.6583,0.6659,0.6728,0.2787,0.4241,-0.2413\\
MIP-2016,0.3500,0.1273,0.2453,0.1116,0.2490,-0.3626\\
QBF-2016,0.7568,0.8434,0.8132,0.6872,0.6260,-0.1366\\
SAT03-16\_INDU,0.3997,0.2494,0.1480,0.3990,0.2888,0.1503\\
SAT12-ALL,0.7617,0.7139,0.6310,0.7516,0.6182,0.6528\\
SAT18-EXP,0.5576,0.5255,0.5726,0.4942,0.3923,0.3202\\
TSP-LION2015,0.4042,-0.7949,-0.2617,-0.4762,-3.2917,-19.1569\\
Average,0.6205,0.5126,0.4893,0.4686,0.2169,-1.2096\\
Average excluding TSP,0.6360,0.6060,0.5429,0.5361,0.4675,0.0724\\
		}
	}
	
\end{table} 
Tab.~\ref{tab:oasc_new} shows the results using the 
default splits of \aslib for each scenario. 
The last two rows of the table denote, respectively, the average closed gap 
across all the scenarios, and the average closed gap across all the scenarios 
excluding the TSP scenario. The latter 
considerably unbalances the results because the performance of all the selectors 
except ASAP is on average worse than the SBS, hence the closed gap score is 
negative. We recall that for the TSP scenario the performance of SBS and VBS 
are very close (cf. Fig.~\ref{fig:inst_tsp}).
%
The best approach in these new experiments is ASAP which looks fairly robust for
all the scenarios. \tool however is not far, especially if we exclude the TSP scenario. 

It is worth noting that the closed gap score can over-penalize an approach performing worse than the SBS. 
Indeed, the average close gap score has upper bound 1 (the VBS cannot be outperformed), 
but not a lower 
bound: one bad result in a fold of a scenario can considerably drop 
the overall average. For 
example, in the TSP scenario the difference in terms of solved instances 
between 
ASAP and \tool is less than 0.3\%. However, the closed gap score is 0.40 versus 
$-0.26$.

We further investigated 
these results, by keeping the very same scenarios and splits while changing the 
evaluation metrics. We considered the Borda count used in the MiniZinc 
Challenge~\cite{mzn-challenge} and defined in Sect.~\ref{sec:eval}. We recall 
that for every instance $i$ of a scenario, a selector $s\in\mathcal{S}$ gets a 
score $\borda(i,s) \in [0,|\mathcal{S}|-1]$ proportional to how many other 
selectors in $\mathcal{S}-\{s\}$ it beats.

\begin{table}[t]
\caption{Borda count results.}
    \label{tab:borda}
    \center
    \resizebox{0.9\linewidth}{!}{%
    \pgfplotstabletypeset[
        precision=4,
        fixed zerofill, 
        col sep=comma,
        row sep=crcr,    
         color cells={min=0,max=3.5},    
        /pgfplots/colormap={whiteblue}{rgb255(0cm)=(255,255,255); rgb255(1cm)=(\tableColorR,\tableColorG,\tableColorB)},
        every head row/.style={%
            before row=\hline,%
            after row=\hline%
        },%
        columns/Scenario/.style={reset styles, string type},
        every last column/.style={
                column type/.add={}{|},
                column type/.add={|}{}
        },
        every first column/.style={
                column type/.add={}{|},
                column type/.add={|}{}
            },
        every first column/.style={
        		column type={|l|}
            },
        every last row/.style={
        	before row=\hline,%
            after row=\hline
        },
        every row 15 column 1/.style={reset styles, string type},
        every row 15 column 2/.style={reset styles, string type},
        every row 15 column 3/.style={reset styles, string type},
        every row 15 column 4/.style={reset styles, string type},
        every row 15 column 5/.style={reset styles, string type},
        every row 15 column 6/.style={reset styles, string type},
    ]{%
    Scenario,Sunny-as2 , ASAP.v2               , AutoFolio , {}*Zilla   , SUNNY-original , Random Forest      \\
ASP-POTASSCO           , 2.5373             , 2.2235             , 2.5465    , 2.4129  , 2.5033          , 2.6163  \\
BNSL-2016              , 2.5368             , 1.283              , 2.7249    , 2.6694  , 2.615           , 3.025   \\
CPMP-2015              , 2.7489             , 2.0501             , 2.407     , 2.2744  , 2.5957          , 2.366   \\
CSP-Minizinc-Time-2016 , 2.5902             , 2.1552             , 2.8031    , 2.2682  , 1.8956          , 2.7214  \\
GLUHACK-2018           , 2.4031             , 1.904              , 2.4275    , 2.0343  , 2.1623          , 2.4528  \\
GRAPHS-2015            , 2.5709             , 2.3045             , 2.6037    , 1.3239  , 2.8054          , 3.3731  \\
MAXSAT-PMS-2016        , 2.5857             , 1.4747             , 2.8863    , 2.3444  , 2.5938          , 2.8616  \\
MAXSAT-WPMS-2016       , 2.9194             , 1.5168             , 2.8361    , 2.428   , 2.6662          , 2.4043  \\
MAXSAT19-UCMS          , 2.5782             , 2.0893             , 2.3597    , 2.5424  , 2.4888          , 2.5189  \\
MIP-2016               , 2.4419             , 2.4239             , 2.4605    , 2.5087  , 2.5368          , 2.4035  \\
QBF-2016               , 2.3819             , 1.8642             , 3.0209    , 2.1644  , 2.5688          , 2.7154  \\
SAT03-16\_INDU         , 2.4301             , 2.1508             , 2.6197    , 2.4159  , 2.3414          , 2.5812  \\
SAT12-ALL              , 2.656              , 1.6785             , 2.7395    , 2.2674  , 2.3141          , 2.825   \\
SAT18-EXP              , 2.4686             , 1.9239             , 2.3395    , 2.0738  , 2.3971          , 2.4998  \\
TSP-LION2015           , 2.4215             , 2.4352             , 2.238     , 2.7398  , 2.4463          , 2.6979  \\
Tot.                   , 38.2705 , 29.4776 , 39.0129   , 34.4679 , 36.9306         , 40.0622\\
}
}

\end{table}

Tab.~\ref{tab:borda} reports the normalized Borda scores for each scenario: if 
$\I$ is the dataset of the scenario, the normalized Borda score of selector $s$ 
is $\dfrac{1}{|\I|} \sum_{i \in \I} \borda(i,s)$.

Results are interesting: by using this metric, adopted by the MiniZinc 
Challenge since 2008, the ranking of Tab.~\ref{tab:oasc_new} is turned upside 
down. Random Forest, which was the approach with the worst closed gap score, is 
now the best approach in terms of Borda count. AutoFolio and \tool are not far 
from it, whereas, surprisingly, ASAP is the one with the worst Borda count.

There can be different reasons for this overturning.
As also discussed in \cite{sunnycp}, the Borda count of the 
MiniZinc Challenge can excessively penalize minimal time differences. 
For instance, if $s$ and $s'$ solve a problem in 1 and 2 seconds
respectively, $s$ scores 2/3 = 0.667 while $s'$ scores 1/3 = 0.333. However, 
if they solve another problem in 500 and 1000 seconds respectively 
the score would remain invariant
even if the absolute time difference in the latter case is 500 seconds.

To investigate whether the difference between closed gap and Borda count scores 
is due to the amplification of small time differences, we define a 
parametric variant of Borda score that considers equivalent 
selectors having runtime difference below a given threshold.
Formally, given a threshold $\delta \geq 0$, we define 
$\borda_\delta(i,s) = \sum_{s' \in \mathcal{S} - \{s\}} \cmp_\delta(m(i,s), 
m(i,s'))$ where:
\[
\cmp_\delta(t, t') = \begin{cases}
0   & \text{ if }t = \tau \\
1   & \text{ if }t < \tau \wedge t' = \tau\\
0.5 & \text{ if }|t - t'| \leq \delta\\
\dfrac{t'}{t+t'} & \text{ otherwise.}
\end{cases} 
\]
If $\delta=0$, $\borda_\delta$ is exactly the $\borda$ score defined in 
Sect.~\ref{sec:eval}, and $\cmp_0$ actually corresponds to the $\cmp$ function.
A score of $0.5$  is instead given if the difference between the runtime 
is less than the time threshold $\delta$.

\begin{figure}[t]
	\includegraphics[width=1\linewidth]{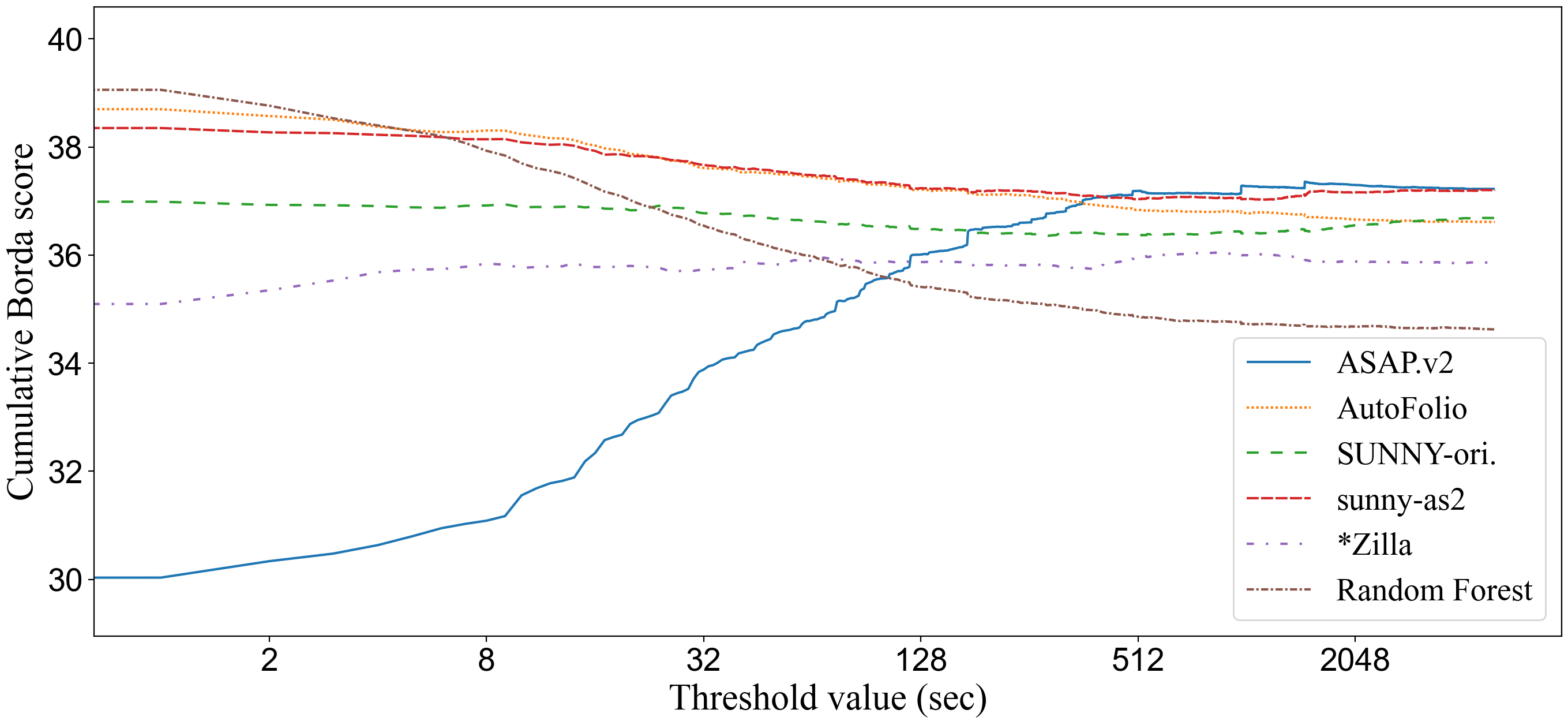}
	\caption{Cumulative Borda count by varying the $\delta$ 
threshold.}\label{fig:threshold_log}
\end{figure}

Fig.~\ref{fig:threshold_log} shows how the cumulative Borda score of 
each approach 
varies when increasing $\delta$ (note the logscale on the x-axis). We can clearly see a reversal 
of performance between ASAP and Random Forest as $\delta$ increases.
This means that Random Forest solve faster more easy instances while ASAP 
and \tool approaches are better in dealing with harder instances.
The reason for this behavior might be that ASAP uses a 
pre-scheduling that could tamper its performance for easy instances that cannot be solved 
in short time by the solver(s) in the pre-schedule. 
\tool seems less susceptible to this problem because no pre-solving is performed and its  scheduling heuristics prioritize the solvers having lower runtime in the neighborhood.

\begin{table}[t]
\caption{Average normalized scores and percentage of solved instances.}
\label{tab:other_stats}	
    \center
    \resizebox{0.9\linewidth}{!}{%
    \pgfplotstabletypeset[
        precision=4,
        fixed zerofill, 
        col sep=comma,
        row sep=crcr,    
         color cells={min=0,max=0.15},    
        /pgfplots/colormap={whiteblue}{rgb255(0cm)=(255,255,255); rgb255(1cm)=(\tableColorR,\tableColorG,\tableColorB)},
        every head row/.style={%
            before row=\hline,%
            after row=\hline%
        },%
        columns/Scenario/.style={reset styles, string type},
        every last column/.style={
                column type/.add={}{|}
        },
        every first column/.style={
                column type/.add={}{|},
                column type/.add={|}{}
            },
        every first column/.style={
        		column type={|l|}
            },
        every last row/.style={
        	before row=\hline,%
            after row=\hline
        },
        every row 2 column 1/.style={reset styles, string type},
        every row 2 column 2/.style={reset styles, string type},
        every row 2 column 3/.style={reset styles, string type},
        every row 2 column 4/.style={reset styles, string type},
        every row 2 column 5/.style={reset styles, string type},
        every row 2 column 6/.style={reset styles, string type},
    ]{%
Scenario, Sunny-as2          , ASAP.v2               , 
AutoFolio , {}*Zilla   , SUNNY-original , Random Forest      \\
Avg. normalized $\parx{1}$  , 0.1062    , 0.1103  , 0.1083    , 
0.1187  , 0.1142          , 0.1449  \\
Avg. normalized $\parten$ , 0.0508    , 0.0497  , 0.0613    , 
0.073   , 0.0597          , 0.1027  \\
Avg. \% solved instances  , 95.5387   , 95.7039, 94.3874   , 
93.2127 , 94.6362         , 90.1971 \\
}
}
\end{table}

			

When small values of $\delta$ are considered, the best approaches is the one 
based on a simple Random Forest classification, followed by AutoFolio for 
values of $\delta$ between 5 and 24 seconds.
\tool becomes the best approach from $\delta=25$ to $\delta=405$.
Then, ASAP takes over. This behavior reflects the 
fact that ASAP solves slightly more instances than \tool, while on average \tool 
is slightly faster doing good choices for easy instances.
This is also corroborated by the numbers in Tab.~\ref{tab:other_stats} 
reporting the average $\parx{1}$ and $\parx{10}$ scores normalized w.r.t.~the 
timeouts, and the 
average percentage of solved instances.
Compared to 
ASAP, \tool has a lower average $\parx{1}$ but higher $\parx{10}$ due to the 
fact that $\parx{10}$ penalizes more the timeouts and \tool solved in average 
0.16\% fewer instances.

The performance of \tool and ASAP asymptotically coincides, and interestingly 
also AutoFolio and SUNNY-original seem to converge.

Overall, \tool achieves a good and robust performance with different evaluation metrics, even if it 
is not always the best approach. Importantly, it consistently outperforms 
its original version on which \sunnyas was based.


\section{Conclusions and Future Work}
\label{sec:conclusions}


In this work we described \tool,
an algorithm selector that---by applying techniques like 
wrapper-based feature selection and configuration of the neighborhood size---significantly 
outperforms its early version \sunnyas and improves on its preliminary version 
submitted in the OASC 2017, when it reached the first 
position in the runtime minimization track.

We conducted an extensive study by varying different 
parameters of \tool, showing how its performance can fluctuate across different 
scenarios of the ASlib.
We also performed an original and in-depth study of the SUNNY algorithm, 
including insights on the instances unsolved by \tool and the use of a 
greedy approach as an effective surrogate of the original SUNNY approach.
We compared \tool against other state-of-the-art AS approaches, and observed how results can change when different evaluation metrics are adopted.

What we experimentally learned from the evaluations performed is that 
feature selection and $k$-configuration are quite effective for SUNNY, and perform 
better when integrated. Moreover, the greedy approach we introduced enables a faster 
and more effective training w.r.t.~the schedule generation procedure of the original SUNNY approach.
Concerning the SUNNY algorithm itself, we exposed the weakness of the similarity assumption on 
which the $k$-NN algorithm used by SUNNY relies. The empirical evaluations we performed confirm both 
the effectiveness of \tool on several AS scenarios, and its robustness under different performance metrics.

A natural future direction for SUNNY that emerges from our experiments is the study of alternative sub-portfolio selection mechanisms not relying on $k$-NN. 
Moreover, we are planning to improve \tool by targeting the solution quality in the optimization scenarios of the OASC competition. 
In these scenarios \tool is strongly penalized because the scheduling of solvers is not allowed. 
We would also like to consider different strategies for scenarios having a low speedup and a limited number of unsolved instances by the best solver of the portfolio.

Another direction for future works is to further study the correlation 
between simple, easy-to-get properties of the scenario (e.g., 
skewness, distribution of labels, distribution of hard instances, number of 
instances, solver marginal contribution) and the best parameters for \tool, 
hoping to find good values for its parameters depending on these simple scenario properties.
Our initial findings have already excluded, e.g., the use of  
mutual information between features in order to limit the number of features. However, additional investigations are needed.



\section*{Acknowledgement}
We thank 
Marius Lindauer 
for the valuable feedback and 
constructive suggestions.

\newpage

\appendix

\section{Composition of OASC Scenarios}
\label{appendix:oasc_analysis}
In this section we provide more details about the composition of
OASC scenarios, by focusing on the performance of the best solvers on the training/test set of every scenario.

In particular, Tab.~\ref{fig:scenario_skewed} shows the three 
fastest algorithms
for each scenario (by merging training and test set). 
For each scenario, the first column indicates the algorithm ID; 
the second and the third column show the
fraction of solved instances in training and test set respectively.
In case of skewed scenarios, e.g. Caren and Monty, the values
in training and test set are significantly different.

\begin{table}[ht]
\caption{The best three overall algorithms in each scenario and the fraction of solved instances in training fold and test fold. }
\label{fig:scenario_skewed}
\resizebox{\columnwidth}{!}{%
\begin{tabular}{|l|l|l|l|l|l|l|l|l|l|l|l|}
\hline
\multicolumn{3}{|c|}{Caren}  & \multicolumn{3}{c|}{Mira}   & \multicolumn{3}{c|}{Magnus}  & \multicolumn{3}{c|}{Monty}   \\ \hline
Best          & Train & Test & Best         & Train & Test & Best          & Train & Test & Best          & Train & Test \\ \hline
algorithm\_8  & 0.89  & 0.83 & algorithm\_4 & 0.94  & 0.96 & algorithm\_17 & 0.95  & 0.94 & algorithm\_14 & 0.75  & 0.83 \\ 
algorithm\_13 & 0.78  & 0.86 & algorithm\_2 & 0.9   & 0.9  & algorithm\_5  & 0.95  & 0.93 & algorithm\_17 & 0.82  & 0.73 \\ 
algorithm\_6  & 0.76  & 0.86 & algorithm\_1 & 0.66  & 0.6  & algorithm\_2  & 0.95  & 0.92 & algorithm\_5  & 0.79  & 0.74 \\ \hline
\multicolumn{3}{|c|}{Quill}  & \multicolumn{3}{c|}{Bado}   & \multicolumn{3}{c|}{Svea}    & \multicolumn{3}{c|}{Sora}    \\ \hline
Best          & Train & Test & Best         & Train & Test & Best          & Train & Test & Best          & Train & Test \\ \hline
algorithm\_15 & 0.84  & 0.81 & algorithm\_3 & 0.88  & 0.88 & algorithm\_27 & 0.73  & 0.75 & algorithm\_8  & 0.93  & 0.93 \\ 
algorithm\_18 & 0.8   & 0.79 & algorithm\_8 & 0.87  & 0.88 & algorithm\_30 & 0.73  & 0.73 & algorithm\_3  & 0.9   & 0.91 \\ 
algorithm\_10 & 0.78  & 0.78 & algorithm\_1 & 0.87  & 0.87 & algorithm\_10 & 0.55  & 0.54 & algorithm\_7  & 0.88  & 0.89 \\ \hline
\end{tabular}
}

\end{table}

\section{Notes on the Preliminary Version of \tool Submitted to the OASC}
\label{appendic:OASC_tool}

In the OASC, \tool did not use the nested CV but a simple 10-fold CV thus 
making it more prone to overfitting.
Moreover, it used a different ranking method for sampling and to generate the 
CV folds.
The instances were clustered in sets based on their best solver and then sorted 
by hardness. The clusters formed a circular list $[c_1, 
\dots, c_n]$. At this point, by iterating over the list of clusters, the 
instances have been added to the folds adding the first 
instance of $c_1$ to the first fold, the first instance of $c_2$ to the 
second fold, and so forth until all the instances in all the clusters were 
added into folds. When the last instance of the cluster was picked, the cluster 
was removed from the list. When the instance was added to the 10th fold, the 
assignment restarted with the first fold.

The instance distribution in each fold generated by this particular systematic 
sampling
is different from all other sampling methods: some class labels are shared 
equally in each fold (as stratified),
and some class labels may have a periodic appearance (half of the 
folds have more instances of a certain class and other folds have few).
We overlooked this behavior at the time of the  
submission.

However, 
this systematic split may have a drawback when the dataset is ordered
against the sampling, i.e. the sampling always picks the instance of the same 
class. This happened for the Sora scenario which has exactly 10 class labels.
The instances were ordered periodically with these classes and each of the 
resulting folds contained more than $90\%$
of the instances of the same class, thus making each fold not 
representative.
For this reason, the submitted version to the OASC had a bad 
performance in Sora.
Due to these limitations, in this work we decided not to use the systematic sampling used in the competition.

\section{Experiments with Relative Values}
\label{appendic:OASC_meta}
In this Section we show the relative values w.r.t. the total number of instances, features, and solvers of a given scenario. In particular:
\begin{itemize}
\item Tab. \ref{fig:training_varying_inst2} shows the fraction between the number of instances on that column and the total number of instances of the scenario on that row
\item Tab. \ref{fig:nfeats2} shows the fraction between the number of features on that column and the total number of features of the scenario on that row
\item Tab. \ref{fig:schedule2} shows the fraction between the number of solvers on that row and the total number of solvers of the scenario on that column
\end{itemize}
In each table, we mark in bold font the cell corresponding to the best closed-gap performance for the given scenario.

\begin{table}[t]
\caption{Relative values of instance limits considering the total number of instances per scenario, in reference to Tab.~\ref{fig:training_varying_inst}. Values in bold font correspond to the best closed-gap performance.}
    \label{fig:training_varying_inst2}
\resizebox{\columnwidth}{!}{%
\begin{tabular}{|l|llllllllllll|l|}
\hline
 & 100          & 150    & 200    & 300    & 400    & 500    & 600    & 700    & 800    & 900    & 1000   & All    & Problems       \\
 \hline
Caren        &        &        &        &        &        &        &        &        &        &        &        & 1        & 66   \\
Mira         & 0.6897 &        &        &        &        &        &        &        &        &        &        & 1        & 145  \\
Magnus       & 0.25   & 0.375  & \textbf{0.5}    & 0.75   &        &        &        &        &        &        &        & 1        & 400  \\
Monty        & 0.2381 & 0.3571 & 0.4762 & \textbf{0.7143} &        &        &        &        &        &        &        & 1        & 420  \\
Quill        & 0.1818 & 0.2727 & \textbf{0.3636} & 0.5455 & 0.7273 & 0.9091 &        &        &        &        &        & 1        & 550  \\
Bado         & 0.1272 & 0.1908 & \textbf{0.2545} & 0.3817 & 0.5089 & 0.6361 & 0.7634 & 0.8906 &        &        &        & 1        & 786  \\
Svea         & 0.0929 & 0.1394 & 0.1859 & 0.2788 & 0.3717 & 0.4647 & 0.5576 & 0.6506 & \textbf{0.7435} & 0.8364 &        & 1        & 1076 \\
Sora         & 0.075  & 0.1125 & 0.15   & 0.2251 & 0.3001 & 0.3751 & 0.4501 & \textbf{0.5251} & 0.6002 & 0.6752 & 0.7502 & 1        & 1333 \\
ASP-POTASSCO & 0.0825 & 0.1238 & 0.165  & 0.2475 & 0.33   & 0.4125 & 0.495  & 0.5776 & 0.6601 & \textbf{0.7426} &        & 1        & 1212 \\
CPMP-2015    & 0.1802 & 0.2703 & 0.3604 & \textbf{0.5405} & 0.7207 & 0.9009 &        &        &        &        &        & 1        & 555  \\
GRAPHS-2015  & 0.0175 & 0.0262 & 0.0349 & 0.0524 & 0.0699 & 0.0873 & 0.1048 & 0.1223 & \textbf{0.1397} & 0.1572 & 0.1747 & 1        & 5725 \\
TSP-LION2015 & 0.0322 & 0.0483 & 0.0644 & 0.0966 & 0.1288 & \textbf{0.161}  & 0.1932 & 0.2254 & 0.2576 & 0.2898 & 0.322  & 1        & 3106 \\
\hline
\end{tabular}
}

\end{table}

\begin{table}[htpb]
\caption{Relative values of features considering the total number of features per scenario, in reference to Tab.~\ref{fig:nfeats}. Values in bold font correspond to the best closed-gap performance.}
    \label{fig:nfeats2}
\resizebox{\columnwidth}{!}{%
\begin{tabular}{|l|llllllllll|l|}
\hline
 &  1            & 2      & 3      & 4      & 5      & 6      & 7      & 8      & 9      & 10     & Features    \\
 \hline
Caren        & \textbf{0.0105} & 0.0211 & 0.0316 & 0.0421 & 0.0526 & 0.0632 & 0.0737 & 0.0842 & 0.0947 & 0.1053   & 95  \\
Mira         & 0.007  & 0.014  & 0.021  & 0.028  & \textbf{0.035}  & 0.042  & 0.049  & 0.0559 & 0.0629 & 0.0699   & 143 \\
Magnus       & 0.027  & 0.0541 & \textbf{0.0811} & 0.1081 & 0.1351 & 0.1622 & 0.1892 & 0.2162 & 0.2432 & 0.2703   & 37  \\
Monty        & 0.027  & 0.0541 & 0.0811 & 0.1081 & \textbf{0.1351} & 0.1622 & 0.1892 & 0.2162 & 0.2432 & 0.2703   & 37  \\
Quill        & 0.0217 & 0.0435 & 0.0652 & \textbf{0.087}  & 0.1087 & 0.1304 & 0.1522 & 0.1739 & 0.1957 & 0.2174   & 46  \\
Bado         & 0.0116 & 0.0233 & \textbf{0.0349} & 0.0465 & 0.0581 & 0.0698 & 0.0814 & 0.093  & 0.1047 & 0.1163   & 86  \\
Svea         & 0.0087 & 0.0174 & 0.0261 & 0.0348 & 0.0435 & 0.0522 & 0.0609 & \textbf{0.0696} & 0.0783 & 0.087    & 115 \\
Sora         & 0.0021 & 0.0041 & 0.0062 & 0.0083 & 0.0104 & \textbf{0.0124} & 0.0145 & 0.0166 & 0.0186 & 0.0207   & 483 \\
ASP-POTASSCO & 0.0072 & 0.0145 & 0.0217 & \textbf{0.029}  & 0.0362 & 0.0435 & 0.0507 & 0.058  & 0.0652 & 0.0725   & 138 \\
CPMP-2015    & \textbf{0.0455} & 0.0909 & 0.1364 & 0.1818 & 0.2273 & 0.2727 & 0.3182 & 0.3636 & 0.4091 & 0.4545   & 22  \\
GRAPHS-2015  & 0.0286 & 0.0571 & 0.0857 & 0.1143 & \textbf{0.1429} & 0.1714 & 0.2    & 0.2286 & 0.2571 & 0.2857   & 35  \\
TSP-LION2015 & \textbf{0.0082} & 0.0164 & 0.0246 & 0.0328 & 0.041  & 0.0492 & 0.0574 & 0.0656 & 0.0738 & 0.082    & 122 \\
\hline
\end{tabular}
}

\end{table}

\begin{table}[htpb]
\caption{Relative values of schedule sizes considering the total number of solvers per scenario, in reference to Tab.~\ref{fig:training_schedule_size}. Values in bold font correspond to the best closed-gap performance.}
    \label{fig:schedule2}
\resizebox{\columnwidth}{!}{%
\begin{tabular}{|l|llllllllllll|}
\hline
   & Caren & Mira & Magnus & Monty  & Quill  & Bado  & Svea   & Sora & ASP-POTASSCO & CPMP-2015 & GRAPHS-2015 & TSP-LION2015 \\
\hline
1       & 0.125 & 0.2  & \textbf{0.0526} & 0.0556 & 0.0417 & 0.125 & 0.0323 & 0.1  & \textbf{0.0909}       & 0.25      & 0.1429      & 0.25         \\
2       & 0.25  & \textbf{0.4}  & 0.1053 & 0.1111 & 0.0833 & 0.25  & 0.0645 & 0.2  & 0.1818       & 0.5       & 0.2857      & \textbf{0.5}          \\
3       & \textbf{0.375} & 0.6  & 0.1579 & \textbf{0.1667} & \textbf{0.125}  & \textbf{0.375} & \textbf{0.0968} & 0.3  & \textbf{0.2727}       & \textbf{0.75}      & 0.4286      & 0.75         \\
4       & 0.5   & 0.8  & 0.2105 & 0.2222 & 0.1667 & 0.5   & 0.129  & 0.4  & 0.3636       & 1         & 0.5714      & 1            \\
5       & 0.625 &   1  & 0.2632 & 0.2778 & 0.2083 & 0.625 & 0.1613 & 0.5  & 0.4545       &           & 0.7143      &              \\
6       & 0.75  &      & 0.3158 & 0.3333 & 0.25   & 0.75  & 0.1935 & 0.6  & 0.5455       &           & 0.8571      &              \\
\hline
\hline
Solvers & 8     & 5    & 19     & 18     & 24     & 8     & 31     & 10   & 11           & 4         & 7           & 4           \\
\hline
\end{tabular}
}

\end{table}

\section{Experiments Related to ASAP-v2 Parameter Tuning}

As described by~\citeA{gonard2019algorithm,gonard17a}, 
the relevant parameters for ASAP-v2  
are the number of estimators (decision trees) and the weight for regularization.
In Tab.~\ref{tab:asap_exp}, we present the results obtained by performing 
various experiments with ASAP-v2 choosing different value combinations for these 
two parameters.
For an entry \emph{Asap-t-$i$-w-$j$}, $i$ means the number of estimators
and $j$ refers to the weight.

Overall, the results show that ASAP-v2 is quite stable but that we did not find 
a combination of hyper-parameters that dominates all the other values for all 
the scenarios.

\begin{table}[ht]
	\caption{ASAP-v2 results with different parameter values.}
	\label{tab:asap_exp}
	\center
	\resizebox{0.9\linewidth}{!}{%
		\pgfplotstabletypeset[
		precision=4,
		fixed,
		fixed zerofill, 
		col sep=comma,
		row sep=crcr,    
		every first column/.style={
			column type={|l}
		},
		color cells={min=0,max=1},    
		/pgfplots/colormap={whiteblue}{rgb255(0cm)=(255,255,255); rgb255(1cm)=(\tableColorR,\tableColorG,\tableColorB)},
		every head row/.style={%
			before row=\hline,%
			after row=\hline%
		},%
		columns/Approach/.style={reset styles, string type},
		every last column/.style={
			column type/.add={}{|}
		},
		every column/.style={
			column type/.add={|}{}
		},
		every last row/.style={
			after row=\hline
		},
		]{%
Approach,Bado,Caren,GLUHACK-2018,MAXSAT19-UCMS,Magnus,Mira,Monty,Quill,SAT18-EXP,Sora,Svea,Average\\
Asap-t-10-w-0.001,0.8414,0.7810,0.3856,0.5590,0.4977,0.0258,0.6965,0.5474,0.5281,0.2716,0.6543,0.5262\\
Asap-t-10-w-0.01,0.8414,0.7810,0.3856,0.5590,0.4977,0.0258,0.6965,0.5474,0.5281,0.2716,0.6543,0.5262\\
Asap-t-10-w-0.05,0.8414,0.7810,0.3856,0.5590,0.4977,0.0258,0.6965,0.5474,0.5281,0.2716,0.6543,0.5262\\
Asap-t-10-w-0.1,0.8414,0.7810,0.3856,0.5590,0.4977,0.0258,0.6965,0.5474,0.5281,0.2716,0.6543,0.5262\\
Asap-t-10-w-0.25,0.8414,0.7810,0.3856,0.5590,0.4977,0.0258,0.6965,0.5474,0.5281,0.2716,0.6543,0.5262\\
Asap-t-10-w-0.5,0.8414,0.7810,0.3856,0.5590,0.4977,0.0258,0.6965,0.5474,0.5281,0.2716,0.6543,0.5262\\
Asap-t-10-w-1.0,0.8414,0.7810,0.3856,0.5590,0.4977,0.0258,0.6965,0.5474,0.5281,0.2716,0.6543,0.5262\\
Asap-t-10-w-2.0,0.8414,0.7810,0.3856,0.5590,0.4977,0.0258,0.6965,0.5474,0.5281,0.2716,0.6543,0.5262\\
Asap-t-20-w-0.001,0.7829,0.5907,0.3883,0.5580,0.4941,0.0546,0.7606,0.6686,0.5975,0.3536,0.6283,0.5343\\
Asap-t-20-w-0.01,0.7829,0.5907,0.3883,0.5580,0.4941,0.0546,0.7606,0.6686,0.5975,0.3536,0.6283,0.5343\\
Asap-t-20-w-0.05,0.7829,0.5907,0.3883,0.5580,0.4941,0.0546,0.7606,0.6686,0.5975,0.3536,0.6283,0.5343\\
Asap-t-20-w-0.1,0.7829,0.5907,0.3883,0.5580,0.4941,0.0546,0.7606,0.6686,0.5975,0.3536,0.6283,0.5343\\
Asap-t-20-w-0.25,0.7829,0.5907,0.3883,0.5580,0.4941,0.0546,0.7606,0.6686,0.5975,0.3536,0.6283,0.5343\\
Asap-t-20-w-0.5,0.7829,0.5907,0.3883,0.5580,0.4941,0.0546,0.7606,0.6686,0.5975,0.3536,0.6283,0.5343\\
Asap-t-20-w-1.0,0.7829,0.5907,0.3883,0.5580,0.4941,0.0546,0.7606,0.6686,0.5975,0.3536,0.6283,0.5343\\
Asap-t-20-w-2.0,0.7829,0.5907,0.3883,0.5580,0.4941,0.0546,0.7606,0.6686,0.5975,0.3536,0.6283,0.5343\\
Asap-t-40-w-0.001,0.8617,0.5906,0.3535,0.5992,0.6399,0.0553,0.6863,0.6677,0.5272,0.3103,0.6475,0.5399\\
Asap-t-40-w-0.01,0.8617,0.5906,0.3535,0.5992,0.6399,0.0553,0.6863,0.6677,0.5272,0.3103,0.6475,0.5399\\
Asap-t-40-w-0.05,0.8617,0.5906,0.3535,0.5992,0.6399,0.0553,0.6863,0.6677,0.5272,0.3103,0.6475,0.5399\\
Asap-t-40-w-0.1,0.8617,0.5906,0.3535,0.5992,0.6399,0.0553,0.6863,0.6677,0.5272,0.3103,0.6475,0.5399\\
Asap-t-40-w-0.25,0.8617,0.5906,0.3535,0.5992,0.6399,0.0553,0.6863,0.6677,0.5272,0.3103,0.6475,0.5399\\
Asap-t-40-w-0.5,0.8617,0.5906,0.3535,0.5992,0.6399,0.0553,0.6863,0.6677,0.5272,0.3103,0.6475,0.5399\\
Asap-t-40-w-1.0,0.8617,0.5906,0.3535,0.5992,0.6399,0.0553,0.6863,0.6677,0.5272,0.3103,0.6475,0.5399\\
Asap-t-40-w-2.0,0.8617,0.5906,0.3535,0.5992,0.6399,0.0553,0.6863,0.6677,0.5272,0.3103,0.6475,0.5399\\
Asap-t-80-w-0.001,0.8814,0.5837,0.3883,0.5992,0.6391,0.0550,0.6860,0.6368,0.5645,0.3126,0.6872,0.5485\\
Asap-t-80-w-0.01,0.8814,0.5837,0.3883,0.5992,0.6391,0.0550,0.6860,0.6368,0.5645,0.3126,0.6872,0.5485\\
Asap-t-80-w-0.05,0.8814,0.5837,0.3883,0.5992,0.6391,0.0550,0.6860,0.6368,0.5645,0.3126,0.6872,0.5485\\
Asap-t-80-w-0.1,0.8814,0.5837,0.3883,0.5992,0.6391,0.0550,0.6860,0.6368,0.5645,0.3126,0.6872,0.5485\\
Asap-t-80-w-0.25,0.8814,0.5837,0.3883,0.5992,0.6391,0.0550,0.6860,0.6368,0.5645,0.3126,0.6872,0.5485\\
Asap-t-80-w-0.5,0.8814,0.5837,0.3883,0.5992,0.6391,0.0550,0.6860,0.6368,0.5645,0.3126,0.6872,0.5485\\
Asap-t-80-w-1.0,0.8814,0.5837,0.3883,0.5992,0.6391,0.0550,0.6860,0.6368,0.5645,0.3126,0.6872,0.5485\\
Asap-t-80-w-2.0,0.8814,0.5837,0.3883,0.5992,0.6391,0.0550,0.6860,0.6368,0.5645,0.3126,0.6872,0.5485\\
Asap-t-160-w-0.001,0.8630,0.7833,0.3525,0.5992,0.6417,0.0556,0.6918,0.6365,0.5956,0.2690,0.6813,0.5609\\
Asap-t-160-w-0.01,0.8630,0.7833,0.3525,0.5992,0.6417,0.0556,0.6918,0.6365,0.5956,0.2690,0.6813,0.5609\\
Asap-t-160-w-0.05,0.8630,0.7833,0.3525,0.5992,0.6417,0.0556,0.6918,0.6365,0.5956,0.2690,0.6813,0.5609\\
Asap-t-160-w-0.1,0.8630,0.7833,0.3525,0.5992,0.6417,0.0556,0.6918,0.6365,0.5956,0.2690,0.6813,0.5609\\
Asap-t-160-w-0.25,0.8630,0.7833,0.3525,0.5992,0.6417,0.0556,0.6918,0.6365,0.5956,0.2690,0.6813,0.5609\\
Asap-t-160-w-0.5,0.8630,0.7833,0.3525,0.5992,0.6417,0.0556,0.6918,0.6365,0.5956,0.2690,0.6813,0.5609\\
Asap-t-160-w-1.0,0.8630,0.7833,0.3525,0.5992,0.6417,0.0556,0.6918,0.6365,0.5956,0.2690,0.6813,0.5609\\
Asap-t-160-w-2.0,0.8630,0.7833,0.3525,0.5992,0.6417,0.0556,0.6918,0.6365,0.5956,0.2690,0.6813,0.5609\\
Asap-t-320-w-0.001,0.8831,0.5899,0.3872,0.5992,0.7146,0.0542,0.7572,0.6370,0.5957,0.2693,0.6942,0.5620\\
Asap-t-320-w-0.01,0.8831,0.5899,0.3872,0.5992,0.7146,0.0542,0.7572,0.6370,0.5957,0.2693,0.6942,0.5620\\
Asap-t-320-w-0.05,0.8831,0.5899,0.3872,0.5992,0.7146,0.0542,0.7572,0.6370,0.5957,0.2693,0.6942,0.5620\\
Asap-t-320-w-0.1,0.8831,0.5899,0.3872,0.5992,0.7146,0.0542,0.7572,0.6370,0.5957,0.2693,0.6942,0.5620\\
Asap-t-320-w-0.25,0.8831,0.5899,0.3872,0.5992,0.7146,0.0542,0.7572,0.6370,0.5957,0.2693,0.6942,0.5620\\
Asap-t-320-w-0.5,0.8831,0.5899,0.3872,0.5992,0.7146,0.0542,0.7572,0.6370,0.5957,0.2693,0.6942,0.5620\\
Asap-t-320-w-1.0,0.8831,0.5899,0.3872,0.5992,0.7146,0.0542,0.7572,0.6370,0.5957,0.2693,0.6942,0.5620\\
Asap-t-320-w-2.0,0.8831,0.5899,0.3872,0.5992,0.7146,0.0542,0.7572,0.6370,0.5957,0.2693,0.6942,0.5620\\
Asap-t-640-w-0.001,0.8634,0.3924,0.4233,0.5992,0.7161,0.0542,0.6180,0.6666,0.5991,0.2693,0.6961,0.5362\\
Asap-t-640-w-0.01,0.8634,0.3924,0.4233,0.5992,0.7161,0.0542,0.6180,0.6666,0.5991,0.2693,0.6961,0.5362\\
Asap-t-640-w-0.05,0.8634,0.3924,0.4233,0.5992,0.7161,0.0542,0.6180,0.6666,0.5991,0.2693,0.6961,0.5362\\
Asap-t-640-w-0.1,0.8634,0.3924,0.4233,0.5992,0.7161,0.0542,0.6180,0.6666,0.5991,0.2693,0.6961,0.5362\\
Asap-t-640-w-0.25,0.8634,0.3924,0.4233,0.5992,0.7161,0.0542,0.6180,0.6666,0.5991,0.2693,0.6961,0.5362\\
Asap-t-640-w-0.5,0.8634,0.3924,0.4233,0.5992,0.7161,0.0542,0.6180,0.6666,0.5991,0.2693,0.6961,0.5362\\
Asap-t-640-w-1.0,0.8634,0.3924,0.4233,0.5992,0.7161,0.0542,0.6180,0.6666,0.5991,0.2693,0.6961,0.5362\\
Asap-t-640-w-2.0,0.8634,0.3924,0.4233,0.5992,0.7161,0.0542,0.6180,0.6666,0.5991,0.2693,0.6961,0.5362\\
Asap-t-1280-w-0.001,0.8629,0.3961,0.3884,0.5992,0.6425,0.0550,0.6866,0.6665,0.5633,0.2457,0.7043,0.5282\\
Asap-t-1280-w-0.01,0.8629,0.3961,0.3884,0.5992,0.6425,0.0550,0.6866,0.6665,0.5633,0.2457,0.7043,0.5282\\
Asap-t-1280-w-0.05,0.8629,0.3961,0.3884,0.5992,0.6425,0.0550,0.6866,0.6665,0.5633,0.2457,0.7043,0.5282\\
Asap-t-1280-w-0.1,0.8629,0.3961,0.3884,0.5992,0.6425,0.0550,0.6866,0.6665,0.5633,0.2457,0.7043,0.5282\\
Asap-t-1280-w-0.25,0.8629,0.3961,0.3884,0.5992,0.6425,0.0550,0.6866,0.6665,0.5633,0.2457,0.7043,0.5282\\
Asap-t-1280-w-0.5,0.8629,0.3961,0.3884,0.5992,0.6425,0.0550,0.6866,0.6665,0.5633,0.2457,0.7043,0.5282\\
Asap-t-1280-w-1.0,0.8629,0.3961,0.3884,0.5992,0.6425,0.0550,0.6866,0.6665,0.5633,0.2457,0.7043,0.5282\\
Asap-t-1280-w-2.0,0.8629,0.3961,0.3884,0.5992,0.6425,0.0550,0.6866,0.6665,0.5633,0.2457,0.7043,0.5282\\
		}
	}
	
\end{table}

\newpage


\bibliography{biblio}
\bibliographystyle{apacite}


\end{document}